%% file: main.tex
\documentclass[11pt]{article}
\usepackage[preprint]{acl}

\usepackage{times}
\usepackage{latexsym}
\usepackage[T1]{fontenc}
\usepackage[utf8]{inputenc}
\usepackage{microtype}
\usepackage{inconsolata}

\usepackage{booktabs}
\usepackage{multirow}
\usepackage{makecell}
\usepackage{tabularx}
\usepackage{amsmath}
\usepackage{amsfonts}
\usepackage{graphicx}
\usepackage{subcaption}
\usepackage{xcolor}
\usepackage{listings}
\usepackage{float}
\usepackage{enumitem}
\usepackage{pifont}
\usepackage{placeins}
\usepackage[table]{xcolor}
\newcommand{\cmark}{\textcolor{green!55!black}{\ding{51}}}
\newcommand{\xmark}{\textcolor{red!75!black}{\ding{55}}}
\newcommand{\pmark}{\textcolor{orange!85!black}{$\triangle$}}
\definecolor{rowhl}{HTML}{EAF2FB}  %
\newcommand{\colorrule}{\specialrule{\lightrulewidth}{\aboverulesep}{0pt}}
\newcommand{\colorbottomrule}{\specialrule{\heavyrulewidth}{0pt}{\belowbottomsep}}
\newcommand{\colortop}{\rule{0pt}{\dimexpr\arraystretch\ht\strutbox+\belowrulesep\relax}}
\newcommand{\colorbot}{\rule[-\dimexpr\arraystretch\dp\strutbox+\aboverulesep\relax]{0pt}{0pt}}
\newcolumntype{Y}{>{\raggedright\arraybackslash}X}

\newcommand{\editppt}{\textsc{EditPPT}}
\newcommand{\bench}{\textsc{DeckEdit-Bench}}
\newcommand{\todo}[1]{\textcolor{red}{\textbf{[TODO]}}}

\title{EditPPT: Structured Execution and Evaluation for Reliable Slide Editing}

\author{
Jiheon Kim \quad Kyudan Jung \quad Jaegul Choo \\
KAIST AI \\
\texttt{\{ben8169, kyudan, jchoo\}@kaist.ac.kr}
}

\begin{document}
\maketitle

\begin{abstract}
Reliable slide editing requires precise modifications to intended objects while preserving the rest of the presentation. We argue that slide editing is a problem of structured, constrained modification rather than open-ended code generation. Despite the complexity of real presentations, editing requests can largely be decomposed into the objects to change, the operations to apply, and the state to preserve. This perspective guides both evaluation and execution. For evaluation, we introduce \bench{}, 183 editing tasks on 28 human-authored decks whose target and preserved objects are annotated at the shape level, so that intended and unintended changes are scored mostly by deterministic rules. For execution, we present \editppt{}, a multi-agent framework that resolves targets before editing, restricts each specialist to a subset of 37 predefined tools, and validates each edit for instruction fulfillment and preservation, correcting or rolling back failed ones. Across five backbone models, \editppt{} reaches 66.7--74.9\% Task Success while keeping Preservation above 97\%. Even with an open 27B model and no task-specific fine-tuning, it approaches Claude Code with Sonnet~5 in Task Success while better preserving the original deck. These results suggest that constraining how edits are executed, not only model scale, makes slide editing more reliable.
Our code is available at \href{https://anonymous.4open.science/r/EditPPT-2027/}{\textbf{https://anonymous.4open.science/r/EditPPT-2027/}}, and our benchmark is available at \href{https://huggingface.co/datasets/EditPPT/DECKEDIT-BENCH}{\textbf{https://huggingface.co/datasets/EditPPT/\\DECKEDIT-BENCH}}.
\end{abstract}
\section{Introduction}
\label{sec:intro}
Editing an existing slide deck is fundamentally different from generating a new one. A presentation already contains its author's layout, formatting, and structure, which an editing system must preserve while applying the requested modifications. This requires identifying the intended targets, modifying them correctly, and leaving unrelated content unchanged. Even a seemingly successful edit can fail if it silently alters formatting or objects outside the requested scope. Reliable slide editing therefore requires not only instruction fidelity but also precise control over what is allowed to change. For example, changing body-text fonts across a deck requires distinguishing body text from titles, even when both are stored as ordinary text boxes. A simple formatting operation can therefore require reasoning about each object's semantic role.

Evaluating this reliability is equally challenging, as success depends not only on what changes but also on what remains unchanged. While existing benchmarks employ diverse evaluation methods, few provide explicit object-level editing scopes for systematically assessing preservation across entire decks~\citep{jung2026talkslideshighefficiencyslide,ofengenden2025pptarenabenchmarkagenticpowerpoint}. This motivates object-grounded evaluation of slide-editing tasks on realistic presentations.

Our central observation is that, despite the structural complexity of human-authored presentations, practical slide-editing requests can largely be characterized by their target objects, required modifications, and preservation constraints. This perspective motivates two complementary approaches, one that evaluates editing outcomes against explicit object-level criteria and one that constrains execution to a predefined set of object-level operations.

For evaluation, we introduce \bench{} (Section~\ref{sec:bench-eval}), comprising 183 editing tasks across 28 human-authored decks. The tasks cover dense, multi-slide editing scenarios, including semantic targeting, cross-slide contextual understanding, and higher-level reasoning. We manually annotate editing scopes at the slide and shape levels, enabling predominantly deterministic evaluation of instruction fidelity and preservation, with model judgment restricted to open-ended outcomes.

This perspective also motivates a different approach to editing. Existing systems often rely on model-generated code or direct OOXML manipulation, exposing the model to low-level document representations and a broad space of possible modifications~\citep{jung2026talkslideshighefficiencyslide,ofengenden2025pptarenabenchmarkagenticpowerpoint}. Although flexible, such approaches can introduce unnecessary complexity, formatting corruption, and unintended side effects, particularly when edits span multiple slides. We argue that editing existing decks at scale requires pairing a restricted action space with explicit target resolution and checks on what must not change.

Building on this principle, we introduce \editppt{} (Section~\ref{sec:method}), a multi-agent framework that separates target resolution from execution. A planner and a dispatcher reason over a compact parse of the deck to resolve the target slides and shapes. Five domain-specific specialists then edit them, each restricted to its own subset of 37 predefined tools, while a validation loop checks for missed requests and unintended changes, guiding correction.

Experiments (Section~\ref{sec:results}) show that \editppt{} maintains strong Task Success and Preservation across five backbone models. With Qwen3.8-27B as the shared backbone, our framework improves Task Success over PPTPilot by 21.9 percentage points. With this open backbone and no task-specific fine-tuning, it also approaches Claude Code with Sonnet~5 in Task Success while achieving higher Preservation and lower Over-edit. These results are consistent with the view that constraining how edits are executed, and not only model scale, helps keep edits within the requested scope.

Our contributions are:
\begin{itemize}[leftmargin=1.8em,itemsep=4pt,parsep=0pt,topsep=6pt]
    \item \textbf{\bench{}: Structured and reproducible evaluation.} We introduce a benchmark of dense, multi-slide, and reasoning-intensive editing tasks on human-authored decks. Manually annotated editing scopes enable predominantly deterministic, object-level evaluation of both instruction fidelity and preservation across entire decks.
    \item \textbf{\editppt{}: Scope-controlled agentic editing.} We propose a multi-agent framework that resolves targets before execution, restricts each specialist to a subset of 37 predefined tools, and validates edits for missed requests and unintended changes.
\end{itemize}

\section{Related Work}
\label{sec:related}

\paragraph{Presentation editing benchmarks.}
PPTC evaluates API-based PowerPoint task completion~\citep{guo-etal-2024-pptc}, while PPTBench includes slide-level modification tasks over structured JSON~\citep{pptbench2025}. TSBench evaluates modular editing instructions using model-based visual assessment~\citep{jung2026talkslideshighefficiencyslide}, and PPTArena assesses in-place edits through structural-diff and visual judges~\citep{ofengenden2025pptarenabenchmarkagenticpowerpoint}. PPT-Eval combines executable rubric checks with model judgments and explicitly penalizes unintended changes~\citep{ppteval2026}. DECKBench studies multi-turn generation and editing with HTML-based slide agents~\citep{jang2026deckbenchbenchmarkingmultiagentframeworks}. These settings offer complementary coverage. Slide-level or HTML-based tasks do not directly evaluate preservation across native PPTX decks, and holistic or task-specific scoring does not by itself establish uniform object-level preservation checks. \bench{} focuses on broad-scope editing of human-authored decks, with explicit target, acceptable, and forbidden shapes and largely deterministic checks of requested changes and preservation (Table~\ref{tab:bench-compare}).

\paragraph{Presentation generation and editing agents.}
Presentation-generation systems explore program synthesis, reference-guided editing, and feedback-driven refinement, as exemplified by AutoPresent, PPTAgent, and DeepPresenter~\citep{ge2025autopresent,zheng-etal-2025-pptagent,zheng2026deeppresenterenvironmentgroundedreflectionagentic}. In-place editors provide closer comparisons. Talk to Your Slides (TTYS) generates Python code against PowerPoint's COM object model~\citep{jung2026talkslideshighefficiencyslide}, PPTPilot combines programmatic and XML edits~\citep{ofengenden2025pptarenabenchmarkagenticpowerpoint}, and Claude Code with the PPTX skill edits the unpacked OOXML package~\citep{anthropic2026claudecode,anthropic2026pptxskill}. Building on structured planning and API-based execution~\citep{yao2023react,schick2023toolformer,wu2023autogen}, \editppt{} restricts model actions to predefined, object-level tools assigned to domain-specific specialists, with explicit editing scopes and post-edit validation. The distinction is therefore not tool use alone, but its integration with fine-grained target resolution and preservation-oriented execution on existing decks.

\section{\editppt{}: Constrained Tool-Based Editing}
\label{sec:method}

\begin{figure*}[t]
\centering
\includegraphics[width=\textwidth]{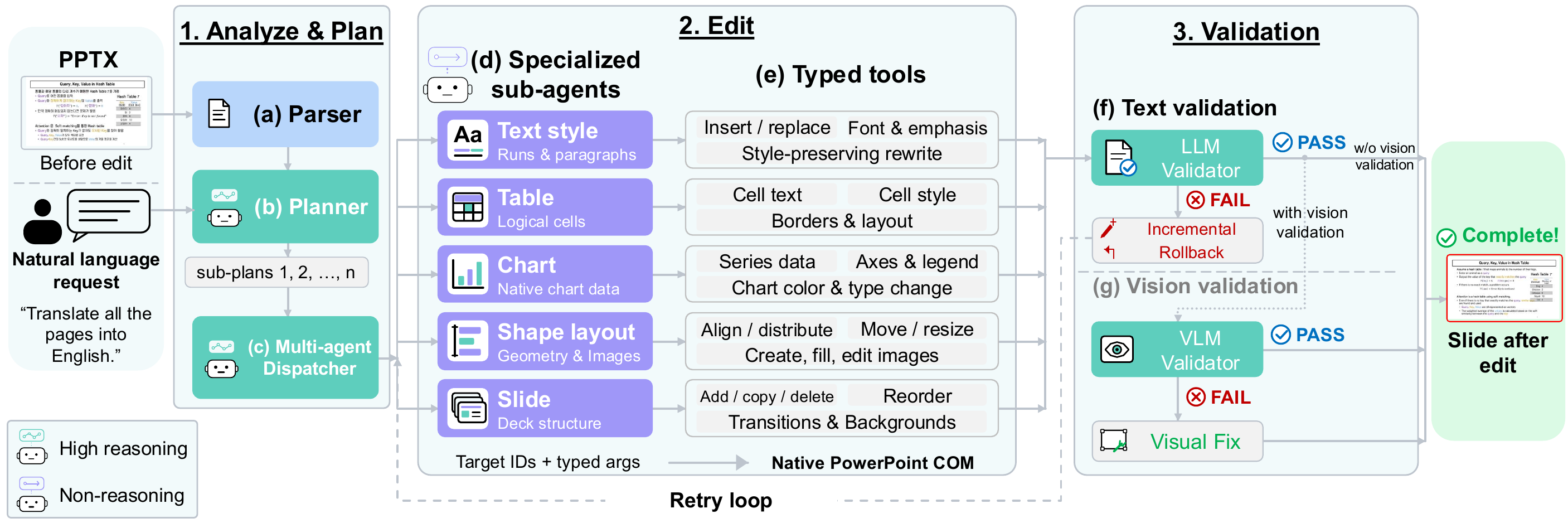}
\caption{Overview of \editppt{}. (a) The parser converts each slide into compact JSON through COM. (b) The planner reasons over the deck outline and splits the request into slide-level tasks. (c) The dispatcher resolves the target shapes and routes subtasks to specialists. (d) Specialists edit only through (e) the tools of their own domain. (f) The text validator checks object-level changes and triggers an incremental fix or a rollback. (g) The vision validator runs only when enabled and checks the rendered slide for visual defects.}
\label{fig:framework}
\end{figure*}

We formulate slide editing as a constrained tool-selection problem rather than open-ended code generation. As shown in Figure~\ref{fig:framework}, the framework combines hierarchical target resolution (\textit{Analyze \& Plan}), domain-specific tool use (\textit{Edit}), and checks of instruction fulfillment and preservation (\textit{Validation}). This split concentrates the reasoning on target resolution and constrains the following edits.

\subsection{Efficient COM-Based Parsing}
\label{sec:method-parser}
Native PowerPoint files contain complex object hierarchies, inherited formatting, and fragmented OOXML structures that are difficult to interpret directly. Our parser reads each slide through the Component Object Model (COM) interface, with inherited formatting already resolved, and converts it into compact JSON that lists text runs, paragraph structure, table cells, chart properties, and grouped shapes, the same objects the tools later edit.

COM, however, leaves several gaps that the parser fills. It rebuilds merged table cells and fine-grained text formatting that COM reports unreliably, and it reads equations as LaTeX from a saved copy of the deck. Since COM cannot describe pictures, the parser captions each picture once with a VLM, so that agents can tell content-bearing pictures from decorative ones without parsing slides from screenshots, which remains unreliable for fine-grained style~\citep{kang2025vlmslideevalevaluatingvlmsstructured}. With these additions, the parser output carries the text, formatting, and structure the agents edit while using 72.4\% fewer input tokens than raw OOXML across 570 slides of \bench{} (Appendix~\ref{app:framework}).

\subsection{Hierarchical Reasoning and Target Resolution}
\label{sec:method-planner-and-dispatcher}
We assign high reasoning effort to the planner and dispatcher, as target resolution requires understanding deck-wide context and fine-grained object information. The planner examines the user instruction alongside a deck-wide text outline to identify content-based targets (e.g., ``every slide mentioning Attention'') and interpret cross-slide references (e.g., ``use the color of slide 3's header''), producing a coordinated set of slide-level tasks. For each task, the dispatcher examines detailed parsed objects to identify target shape IDs, decomposes complex requests into domain-specific subtasks, and assigns them to appropriate specialists. By resolving deck-level intent and shape-level targets upfront, these agents reduce the reasoning required during subsequent editing and avoid repeated object searches.

\subsection{Domain-Constrained Edit Execution} 
\label{sec:method-edit} 
The dispatched subtasks are executed by five domain-specific specialists responsible for text and styling, tables, charts, shape layout, and slide-level operations. With target selection and task decomposition already completed, each specialist runs at the lowest reasoning setting its backbone allows (Appendix~\ref{app:implementation}), focusing on selecting and executing appropriate editing tools. Rather than having access to all 37 tools, each specialist can invoke only tools within its domain. This restriction simplifies tool selection while limiting the risk of inappropriate operations. The selected tools are executed through COM with validated arguments and explicit editing scopes (Appendix~\ref{app:algorithms}).

\subsection{Validation and Recovery}
\label{sec:method-validation}
After each attempt at a subtask, the framework re-parses the slide and computes object-level differences against its state before the subtask. An LLM-based text validator examines these differences alongside the instruction, reporting missed targets and unintended modifications. A failed attempt is retried in one of two ways. When the edits made so far are correct and only some targets remain, the next attempt keeps them and makes only the missing changes. When an edited shape is itself wrong, the framework restores the pre-subtask state and starts the subtask over, giving the next attempt the feedback from all failed ones. If validation still fails after three attempts, the framework keeps the attempt in which the validator found the fewest missed targets and no wrong edits, or otherwise restores the previous state (Appendix~\ref{app:algorithms}). An optional vision-based validator further detects new visual defects and suggests localized corrections.

\section{\bench{}: Structured and Reproducible Evaluation}
\label{sec:bench-eval}
Unlike open-ended slide generation, slide editing allows the intended targets, required modifications, and preservation constraints of a request to be specified explicitly at the object level. \bench{} builds on this property, combining broad-scope editing tasks on human-authored decks with exhaustive target annotations and predominantly deterministic evaluation at the shape level. Each task states what must change, what may change, and what must stay as it was, so that both the requested edits and any unintended side effects can be scored without a reference deck.

\subsection{Benchmark Composition and Difficulty}
\label{sec:benchmark}

\bench{} comprises 183 editing tasks across 28 human-authored decks (582 slides), including research talks, lectures, proposals, and reports in Korean and English. We wrote the tasks for each deck, favoring edits that repeat across slides, combine several changes, or leave room for interpretation, and each annotation specifies the required end state without assuming any tool or execution path (Appendix~\ref{app:task-construction}). As shown in Table~\ref{tab:bench-compare}, \bench{} contains 15.7 shapes per slide on average, compared with 9.3 in PPTArena. These decks also contain irregular structures, such as titles stored in ordinary text boxes, grouped objects, and tables embedded as images, making object identification difficult from file-level types alone.

More importantly, DeckEdit-Bench emphasizes editing scope over deck length. Tasks target 9.4 slides on average, with 30.6\% spanning at least 10 slides and 19.7\% at least 20. On average, target slides contain over three times as many candidate shapes as in PPTArena, with 32\% of tasks involving at least 100 candidates versus 18\% in PPTArena.

The main challenge is semantic targeting at scale. Instructions often identify objects by content or function rather than explicit IDs, while some require reasoning across slides, interpreting embedded images, or selecting key information. Even a simple request such as ``change every slide's body font'' requires identifying body text across varied layouts without modifying unrelated elements. Further task statistics are provided in Appendix~\ref{app:benchmark}.

\begin{table*}[t]
\centering
\small
\setlength{\tabcolsep}{3.2pt}
\renewcommand{\arraystretch}{1.12}
\resizebox{\textwidth}{!}{%
\begin{tabular}{l ccccccc cccc ccc}
\toprule
& \multicolumn{7}{c}{\textbf{Source decks}} & \multicolumn{4}{c}{\textbf{Per task}} & \multicolumn{3}{c}{\textbf{Ground truth \& scoring}} \\
\cmidrule(lr){2-8}\cmidrule(lr){9-12}\cmidrule(lr){13-15}
\textbf{Benchmark} & \textbf{PPTX} & \makecell{\textbf{Decks /}\\\textbf{slides}} & \makecell{\textbf{Sl./deck}\\(med.)} & \makecell{\textbf{Sh./sl.}\\(mean/med.)} & \makecell{\textbf{Pic./sl.}\\(\% sl.)} & \makecell{\textbf{Eq.}\\\textbf{decks}} & \makecell{\textbf{Multiling.}\\\textbf{slides}}
 & \makecell{\textbf{Tgt sl.}\\(mean/med.)} & \makecell{\textbf{$\ge$10 / $\ge$20}\\\textbf{sl.}} & \makecell{\textbf{Cand.}\\(mean/med.)} & \makecell{\textbf{$\ge$100}\\\textbf{cand.}}
 & \textbf{Obj.} & \textbf{Pres.} & \textbf{Scoring} \\
\midrule
PPTBench  & \xmark & --        & --   & --        & --          & --    & --    & 1       & 0 / 0\%  & --      & --   & \pmark & \pmark & R \\
DECKBench & \xmark & --        & --   & --        & --          & --    & --    & --      & --       & --      & --   & \xmark & \xmark & S, L \\
\midrule
TSBench   & \cmark & 56 / 72   & 1    & 4.3 / 3.5 & 0.72 (54\%) & 0\%   & 1.4\% & $\le$4 & 0 / 0\%  & $\le$17 & 0\%  & \xmark & \xmark & V \\
PPTArena  & \cmark & 100 / 999 & 5    & 9.3 / 4   & 1.10 (46\%) & 5\%   & 1.2\% & 6.6 / 4 & 22 / 9\% & 55 / 17 & 18\% & \xmark & \pmark & L, V \\
PPT-Eval  & \cmark & 12 / 404  & 30.5 & 13.1 / 4  & 0.70 (30\%) & 0\%   & 0\%   & 2.9 / 1$^\dagger$ & 4 / 4\% & 34 / 4$^\dagger$ & 5\% & \pmark & \pmark & R, V \\
\colorrule
\rowcolor{rowhl}\colortop\colorbot\textbf{\bench{}} & \cmark & 28 / 582 & 18 & \textbf{15.7} / \textbf{7} & \textbf{2.34} (\textbf{73\%}) & \textbf{36\%} & \textbf{21\%} & \textbf{9.4} / 3 & \textbf{31} / \textbf{20\%} & \textbf{179} / \textbf{39} & \textbf{32\%} & \cmark & \cmark & R (+L, V) \\
\colorbottomrule
\end{tabular}}
\caption{Comparison of slide-editing benchmarks. Corpus statistics are recomputed from the released source decks with one counting rule (Appendix~\ref{app:benchmark}). \textbf{Sh./sl.} and \textbf{Pic./sl.} count leaf shapes and pictures per slide, with the share of slides containing a picture. \textbf{Multiling.} counts slides that mix English with a non-Latin script. \textbf{Tgt sl.} counts target slides per task and \textbf{Cand.} all shapes on them, a proxy for the target-search space, and columns with $\ge$ give the share of tasks at or above each threshold. \textbf{Obj.} and \textbf{Pres.} mark object-level target annotation and preservation scoring. \textbf{Scoring} uses \textbf{R}ules, \textbf{L}LM, \textbf{V}LM, or \textbf{S}imilarity to a reference, and \bench{} uses model judgments, in parentheses, for only 26 of its 183 tasks. \pmark\ denotes partial support. $^\dagger$Estimated from prompt wording.}
\label{tab:bench-compare}
\end{table*}

\subsection{All Slides and Shapes Are Annotated}
\label{sec:eval-annotation}

A single reference output is insufficient for evaluating slide editing, as multiple outcomes may satisfy the same instruction. For instance, making two charts ``the same size'' with ``aligned tops'' can be satisfied at any common size and height, and ``a different color for each box'' admits countless palettes. More importantly, verifying the requested changes alone does not reveal unintended modifications. We therefore manually annotate every target slide and shape for all 183 tasks, specifying the required postconditions and accounting for all remaining objects through preservation constraints.

Within each target slide, every shape is \emph{target}, \emph{acceptable}, or \emph{forbidden}. \emph{Target} shapes must satisfy the requested changes, \emph{acceptable} shapes may remain unchanged or change only as annotated, and \emph{forbidden} shapes must remain untouched.

\begin{figure*}[t]
\centering
\includegraphics[width=\textwidth]{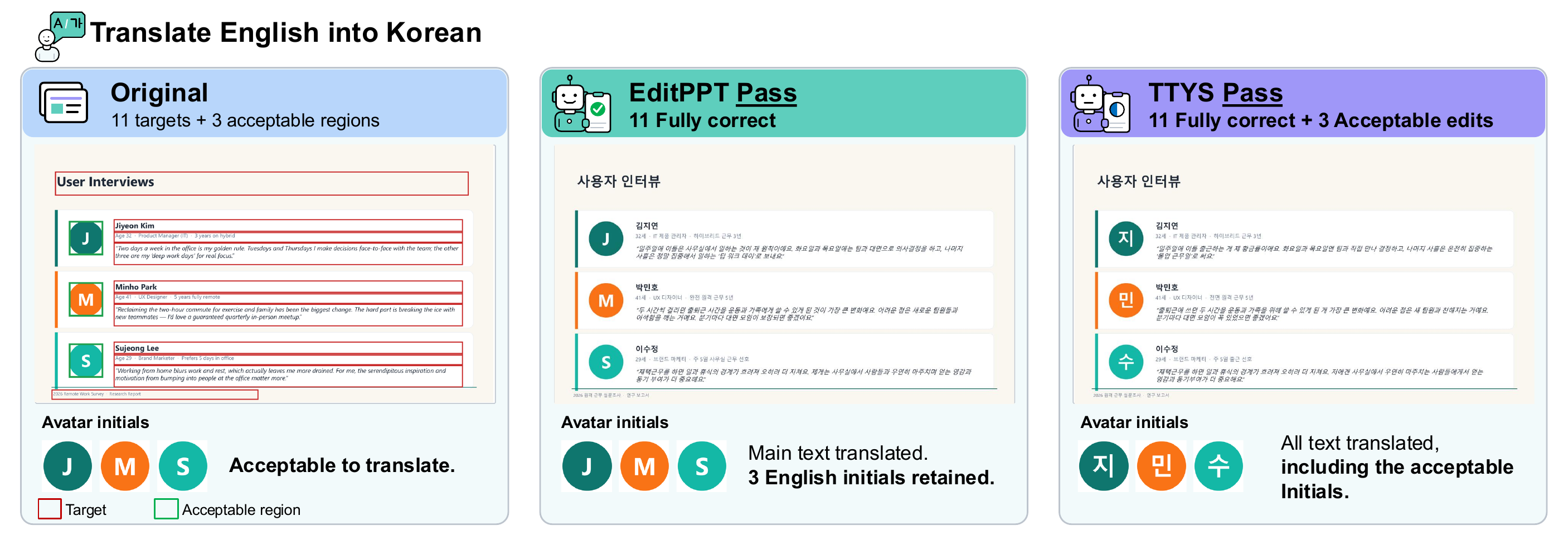}
\caption{An acceptable annotation in practice, for ``Translate every slide's text into Korean'' on one slide of a survey deck. Left: the original slide with its 11 target shapes and three avatar initials, which are design badges that may be translated or kept. Middle and right: \editppt{} keeps the initials while TTYS translates them, and both translate every target and succeed on this slide. Without the acceptable category, translating the initials would count as damage.}
\label{fig:acceptable}
\end{figure*}

The \emph{acceptable} category handles instructions that admit multiple reasonable interpretations. Rather than enforcing a single interpretation, we explicitly allow alternative edits without penalizing unchanged objects. For each acceptable shape, the annotation also specifies which changes are allowed, such as the font only or the text only, and can restrict the allowance to particular text spans or terms. For example, when changing the font of ``all body text,'' whether footnotes or source references belong to the body may be ambiguous. Changing their font is therefore permitted but not required, while modifications to their size, color, or content are penalized. Similarly, a proper noun such as ``Transformers'' may be transliterated or left unchanged during translation, and Figure~\ref{fig:acceptable} shows a similar case with two real outputs. This treatment applies to 32 tasks, and all annotations were reviewed in detail by the authors.

To check whether these allowances favor particular systems, we also rescore all systems with acceptable shapes treated either as forbidden or as required targets. Our framework loses up to 5.5 points in Task Success but stays ahead of PPTPilot and TTYS, and its difference from Claude Code with Sonnet~5 remains non-significant under both alternative scoring interpretations (Appendix~\ref{app:sensitivity}).

\subsection{Deterministic and Hybrid Evaluation}
\label{sec:eval-scoring}

Our evaluator compares the original and edited PPTX files against the annotations defined for each benchmark task, independently of the editing system. Requested modifications are verified using 69 types of checks covering text formatting, content, geometry, and slide-level properties. Preservation is assessed by comparing key object properties before and after editing to verify that \emph{forbidden} shapes remain unchanged and to detect unintended shape additions or deletions across all slides.

Of the 183 tasks, 157 are evaluated entirely through deterministic rules. The remaining 26 require additional LLM-based judgment for open-ended outcomes, such as translation fidelity, summarization, and identifying important content. For three of them, the judge model also compares pictures before and after the edit. For these tasks, deterministic rules first verify target selection and preservation, while the LLM judges only specific outcomes on annotated objects that cannot be evaluated through rules. These judgments use gpt-6-luna at high reasoning effort, with the same prompts for every system (Appendix~\ref{app:check-types}).

Each target slide is graded on its target shapes alone. It is \emph{fail} if none of its required postconditions is met. Otherwise, it is \emph{pass} only if all of them are met and the targets change no property beyond what the request allows, and \emph{partial} in all remaining cases, such as a title that moves when only its color was to change. Separately, the slide is damaged by any unauthorized change to an acceptable or forbidden shape and by any unrequested addition or deletion of a shape, even when the requested edit succeeds. A slide succeeds only if it passes without damage, and unauthorized changes to non-target slides are counted as over-editing.

We report five complementary metrics to distinguish successful edits from unintended modifications. \textbf{Task Success} requires every target slide to pass without damage, with no unauthorized changes to non-target slides. \textbf{Slide Success} is the fraction of target slides that pass without damage, and \textbf{Instruction Following} (IF) is the fraction graded pass or partial regardless of damage. \textbf{Over-edit} is the fraction of over-edited slides among the over-edited and the correctly targeted slides, where a target slide is correctly targeted if the system changed at least one of its target shapes or it already met the request. \textbf{Preservation} is the proportion of undamaged slides among the correctly targeted slides, so it concerns the non-target objects on those slides rather than the whole deck. Tasks without a usable result count as failures but do not enter Over-edit or Preservation. Formal definitions are provided in Appendix~\ref{app:golden}.

\section{Experimental Setup}
\label{sec:experiments}

\paragraph{Systems and backbones.}
We compare our framework with Claude Code + PPTX skill, PPTPilot, and Talk to Your Slides (TTYS). Our five backbones are gpt-6-sol, gpt-6-luna, Qwen3.8-27B, Sonnet~5, and Haiku~4.5~\citep{openai2026gpt6,qwen2026qwen38,anthropic2026sonnet5,anthropic2025haiku45}. Table~\ref{tab:main} lists the system--backbone combinations used for same-backbone comparisons, and Appendix~\ref{app:baselines} details the baseline configurations.

\paragraph{Evaluation setup.}
Each task is run from a fresh, unmodified copy of the source deck. The optional vision validator of our framework is disabled in all runs. Missing, unchanged, or unreadable outputs are treated as unsuccessful executions.

Cost is computed at API list prices, while no comparable API cost is reported for locally served Qwen. We report 95\% confidence intervals using 10,000 bootstrap resamples over the 28 decks, accounting for correlations among tasks within the same deck~\citep{efron1993bootstrap}. These intervals and additional statistics are in Appendix~\ref{app:stats}.

\section{Results}
\label{sec:results}

\begin{table*}[t]
\centering
\small
\setlength{\tabcolsep}{4pt}
\resizebox{\textwidth}{!}{%
\begin{tabular}{ll c rrrrr rrr}
\toprule
System & Model & Modified & \textbf{TS}\,$\uparrow$ & Slide\,$\uparrow$ & IF\,$\uparrow$ & Over-edit\,$\downarrow$ & Pres\,$\uparrow$ & Cost/task\,$\downarrow$ & Time/task & In / Out tok \\
\midrule
\multirow{3}{*}{\textbf{PPTPilot}}
 & gpt-6-sol     & 147/183 & 51.4 & 69.1 & 83.3 & 0.68 & \underline{98.3} & \$0.298 & 1.7 min & 148k / 4.3k \\
 & gpt-6-luna    & 137/183 & 43.2 & 59.0 & 73.6 & 0.68 & 96.3 & \$0.017 & 1.6 min & 153k / 5.0k \\
 & Qwen3.8-27B   & 150/183 & 50.3 & 47.3 & 53.2 & 0.42 & 93.1 & -- & 10.2 min & 122k / 17.8k \\
\midrule
\multirow{2}{*}{\textbf{Talk to Your Slides}}
 & gpt-6-sol     & 176/183 & 35.5 & 64.8 & 76.6 & 6.23 & 93.6 & \$0.625 & 7.4 min & 105k / 41.7k \\
 & gpt-6-luna    & 169/183 & 29.0 & 41.3 & 49.3 & 5.63 & 95.5 & \$0.027 & 6.7 min & 84k / 36.5k \\
\midrule
\multirow{2}{*}{\textbf{Claude Code + PPTX skill}}
 & Sonnet 5      & 183/183 & 73.8 & 81.5 & \textbf{97.8} & 0.88 & 93.9 & \$0.387 & 1.6 min & 577k / 11.0k \\
 & Haiku 4.5     & 179/183 & 53.6 & 54.4 & 81.0 & 1.55 & 84.7 & \$0.182 & 1.6 min & 677k / 8.1k \\
\colorrule
\rowcolor{rowhl}\colortop& gpt-6-sol & 183/183 & \textbf{74.9} & \textbf{89.0} & 95.7 & 0.42 & 98.1 & \$0.429 & 6.1 min & 225k / 8.0k \\
\rowcolor{rowhl}& gpt-6-luna & 183/183 & 73.8 & 88.5 & 95.5 & 0.54 & \textbf{98.9} & \$0.036 & 5.1 min & 237k / 8.4k \\
\rowcolor{rowhl}& Qwen3.8-27B & 183/183 & 72.1 & 87.4 & 95.8 & \underline{0.36} & 97.4 & -- & 7.3 min & 255k / 14.6k \\
\rowcolor{rowhl}& Sonnet 5 & 183/183 & \underline{74.3} & \underline{88.8} & \underline{97.0} & \textbf{0.30} & 97.6 & \$0.919 & 4.9 min & 484k / 15.1k \\
\rowcolor{rowhl}\multirow{-5}{*}{\textbf{\editppt{}}}\colorbot & Haiku 4.5 & 183/183 & 66.7 & 83.8 & 93.3 & 0.55 & 98.0 & \$1.031 & 12.2 min & 408k / 107.6k \\
\colorbottomrule
\end{tabular}}
\caption{Results on all 183 tasks of \bench{}. \emph{Modified} counts tasks for which the system returned a deck that differs from the input, and all other tasks count as failures. TS, Slide, IF, and Pres denote Task Success, Slide Success, Instruction Following, and Preservation (Section~\ref{sec:eval-scoring}). Cost is the mean list price per task, unavailable for the locally served Qwen (--). Input tokens include cached prompt tokens. Best value per quality metric in bold, second underlined.}
\label{tab:main}
\end{table*}

\subsection{Main Results}
\label{sec:res-main}
Table~\ref{tab:main} summarizes the results across all 183 tasks. Across all five backbones, our framework keeps both Task Success and Preservation high with little over-editing, whereas every baseline falls behind on at least one of them. These trade-offs between instruction following, task completion, and preservation are not captured by a single success metric.

\paragraph{Same backbone, different system.}
With Sonnet~5, our framework and Claude Code achieve comparable Task Success, but differ in balancing instruction following and preservation. Claude Code achieves slightly higher Instruction Following, whereas our framework improves Slide Success by 7.3 points and Preservation by 3.7 points, while reducing Over-edit from 0.88\% to 0.30\%.

These differences are consistent with the systems' editing approaches. Claude Code permits broader code-based modifications, whereas we restrict execution to predefined tools on selected objects. Although this comparison does not isolate the effects of individual components, the results support the value of constrained execution to preserve unrelated content. The same pattern holds with Haiku~4.5, where our framework achieves higher Task Success and Preservation than Claude Code, although at substantially greater runtime and cost.

With Qwen3.8-27B, our framework improves Task Success over PPTPilot by 21.9 points and at least partly follows the instruction on nearly all target slides, against about half for PPTPilot. Part of this gap arises from execution failures, as 19 PPTPilot runs exceed the model's 262k-token context window. On the 152 tasks its runtime allowed, TTYS reaches 16.4 Task Success, against 75.0 for our framework (Appendix~\ref{app:baselines}).

\paragraph{Open-model comparison.}
With Qwen3.8-27B and no task-specific fine-tuning, our framework comes within 1.6 points of Claude Code with Sonnet~5 in Task Success. Claude Code leads slightly in Instruction Following, but ours leads in Slide Success and Preservation with under half the Over-edit. Thus, a constrained framework achieves competitive task completion using an open 27B model while better preserving the original deck.

\paragraph{Preservation and completion.}
High preservation alone does not imply successful editing. PPTPilot with gpt-6-sol matches our Preservation but returns no modified deck for one in five tasks and succeeds on only about half of all tasks. TTYS shows a different failure pattern, with over ten times the Over-edit of any of our configurations. In contrast, our framework returns a modified deck for every task across five backbones while keeping preservation high. Its main advantage is not preservation alone, but combining it with consistently high task completion and low over-editing.

\subsection{Robustness to Deck Length}
\label{sec:res-length}

\begin{table}[t]
\centering
\small
\setlength{\tabcolsep}{4pt}
\begin{tabular}{ll ccc}
\toprule
System & Backbone & Short & Medium & Long \\
\midrule
\multirow{3}{*}{\textbf{PPTPilot}}
 & gpt-6-sol   & 55.6 & 58.0 & 35.4 \\
 & gpt-6-luna  & 40.7 & 50.6 & 33.3 \\
 & Qwen3.8-27B & 61.1 & 59.3 & 22.9 \\
\midrule
\multirow{2}{*}{\textbf{TTYS}}
 & gpt-6-sol   & 50.0 & 33.3 & 22.9 \\
 & gpt-6-luna  & 44.4 & 27.2 & 14.6 \\
\midrule
\multirow{2}{*}{\textbf{Claude Code}}
 & Sonnet 5    & 74.1 & \textbf{84.0} & 56.2 \\
 & Haiku 4.5   & 59.3 & 58.0 & 39.6 \\
\colorrule
\rowcolor{rowhl}\colortop& gpt-6-sol   & 77.8 & \underline{80.2} & \underline{62.5} \\
\rowcolor{rowhl}& gpt-6-luna  & \textbf{90.7} & 71.6 & 58.3 \\
\rowcolor{rowhl}& Qwen3.8-27B & 75.9 & 77.8 & 58.3 \\
\rowcolor{rowhl}& Sonnet 5    & \underline{83.3} & 72.8 & \textbf{66.7} \\
\rowcolor{rowhl}\multirow{-5}{*}{\textbf{\editppt{}}}\colorbot& Haiku 4.5   & 75.9 & 66.7 & 56.2 \\
\colorbottomrule
\end{tabular}
\caption{Task Success by deck length: short ($\le$10 slides, 54 tasks), medium (11--30, 81), long ($>$30, 48). Best per column in bold, second underlined.}
\label{tab:length}
\end{table}

Table~\ref{tab:length} breaks Task Success down by deck length. All systems perform worse on long decks, where missing a single target slide fails the whole task. On long decks, each of our configurations stays ahead of the baselines that use the same model, and our Sonnet~5 configuration is the strongest of all, whereas Claude Code with Sonnet~5 leads on medium decks. Our framework also declines on long decks. Grouping tasks by their number of target slides shows the same decline, which also appears within medium and long decks (Appendix~\ref{app:scope}). Even on tasks with 20 or more target slides, our framework reaches about 87--90\% Slide Success with the gpt-6, Qwen, and Sonnet backbones, which illustrates how strict the requirement that every target slide succeed becomes as the scope grows. Differences in task composition across these groups may also contribute.

\begin{table}[t]
\centering
\small
\setlength{\tabcolsep}{3.5pt}
\resizebox{\columnwidth}{!}{%
\begin{tabular}{l rrrrr r}
\toprule
Variant & \textbf{TS}\,$\uparrow$ & Slide\,$\uparrow$ & IF\,$\uparrow$ & Over-edit\,$\downarrow$ & Pres\,$\uparrow$ & Cost/task \\
\midrule
\multicolumn{7}{l}{\emph{All 183 tasks}} \\
Full system   & \textbf{73.8} & \textbf{88.5} & \textbf{95.5} & \underline{0.54} & \textbf{98.9} & \$0.036 \\
Single agent  & 61.7 & 82.4 & 93.3 & 5.59 & \underline{98.7} & \$0.048 \\
No validation & \underline{70.5} & \underline{86.3} & \underline{94.8} & \textbf{0.48} & 96.8 & -- \\
\midrule
\multicolumn{7}{l}{\emph{57 tasks in which the loop triggered a retry}} \\
Full system   & \textbf{38.6} & \textbf{79.3} & \textbf{94.7} & 0.81 & \textbf{97.7} & \$0.092 \\
No validation & 28.1 & 74.4 & 93.2 & \textbf{0.67} & 93.0 & \$0.053 \\
\bottomrule
\end{tabular}}
\caption{Ablations of \editppt{} with gpt-6-luna. \emph{Single agent} replaces the dispatcher and specialists with one agent that can use all 37 tools. \emph{No validation} re-runs only the 57 tasks in which validation triggered a retry and reuses the other 126 results, so its cost over all 183 tasks is not reported. Best value per block in bold, second underlined.}
\label{tab:ablation}
\end{table}

\subsection{Ablations}
\label{sec:res-ablation}
We examine the contributions of the dispatcher--specialist architecture and the validation loop using gpt-6-luna (Table~\ref{tab:ablation}). Both variants are compared against the full system on the same tasks.

\paragraph{Single agent.}
We replace the dispatcher and five specialists with a single agent that accesses all 37 tools, while keeping the planner and validation loop unchanged. Task Success drops by 12.0 points. Although Preservation remains nearly unchanged, Over-edit increases more than tenfold. The single agent modifies 98 non-target slides, compared with only 9 for the full system. These errors occur primarily in conditional, deck-wide requests, where the agent modifies slides that do not satisfy the specified conditions. The effect is particularly pronounced on long decks, where Task Success falls sharply from 58.3\% to 35.4\%.

The single-agent variant also uses about one and a half times as many input tokens per task and costs more, which reflects the added overhead of exposing the full tool set to a single agent.

\paragraph{Validation loop.}
To assess feedback-guided recovery, we disable post-edit validation while retaining retries for tool errors. Since validation triggers retries on only 57 tasks, we re-run those tasks without the loop and reuse the first-attempt results for the remaining 126. Across all 183 tasks, removing validation reduces Task Success by 3.3 points and Preservation by 2.1 points.

On these 57 tasks, validation raises Preservation by 4.7 points and Task Success by 10.5 points. The preservation gain is supported by the paired bootstrap analysis, while the improvement in Task Success is less conclusive. These results suggest that the loop is particularly useful for recovering from damaging edits, rather than merely increasing instruction completion. This benefit comes at about 1.7 times the cost per task in the retry subset.

\subsection{Efficiency Trade-offs}
\label{sec:res-cost}
Our accuracy and preservation gains come with additional runtime. With the gpt-6 and Sonnet~5 backbones, our framework is roughly three to four times slower per task than Claude Code and PPTPilot, which edit the files directly. This overhead is consistent with repeated COM interactions, re-parsing, and validation, although their individual costs are not isolated. TTYS, which also drives PowerPoint, is slower still, while with Qwen3.8-27B our framework is faster than PPTPilot.

Input tokens alone do not determine cost. With Sonnet~5, we read fewer input tokens than Claude Code but pay more than twice as much per task, and Haiku~4.5 is an outlier that writes far more output and runs longest of all configurations (Table~\ref{tab:main}). Constrained tool-based editing can therefore improve reliability without necessarily reducing end-to-end latency or cost. Detailed cost distributions are provided in Appendix~\ref{app:cost}.

\subsection{Visual Defect Audit}
\label{sec:vision-validation}
Our object-level evaluator does not see rendering defects such as overflowing or overlapping text. However, in a separate audit of the 3{,}149 edited slides whose rendering changed in two runs of our framework, with gpt-6-luna and gpt-6-sol, only 116 (3.7\%) show a newly introduced defect. Two survey raters reviewed 88 of these 116, and at least one of them confirmed a defect in 72. Of the 135 and 137 tasks that the two runs succeed on (Task Success, Table~\ref{tab:main}), only six each have such a slide. As edits rarely break rendering and a visual check costs a render and a model call per slide, vision validation is optional and off in all runs.

Still, a single defect can spoil an otherwise successful edit, so we fine-tune Qwen3.5-9B with LoRA as an open vision validator on person-confirmed edits, with and without defects, synthesized on public decks. It judges whether an edit introduced a critical rendering defect and proposes a repair as tool calls. On held-out synthetic edits, its yes-or-no judgment reaches 0.932 accuracy, against 0.60 before fine-tuning (Appendix~\ref{app:vision}).

\section{Conclusion}
\label{sec:conclusion}

We presented \bench{} and \editppt{}, grounded in a structured view of slide editing that distinguishes target objects, intended modifications, and preservation constraints. \bench{} enables predominantly deterministic evaluation through object-level annotations, while \editppt{} combines target resolution, constrained tool execution, and validation. Across five backbone models, our framework maintains strong Task Success and Preservation, with an open 27B backbone approaching Claude Code with Sonnet~5 in Task Success. Ablations indicate that the dispatcher--specialist architecture improves scope control, while validation supports recovery from damaging edits. Overall, our findings suggest that reliable slide editing depends not only on model capability, but also on how modifications are structured and constrained.

\section*{Limitations}
Every task in \bench{} requires manual annotation of its target slides and shapes, which limits how quickly the benchmark can grow. The evaluation checks object properties, so it cannot see rendering defects such as overflowing text, which we audit separately (Section~\ref{sec:vision-validation}).

Our current implementation drives PowerPoint through COM and edits decks sequentially, so running experiments takes considerable time. The framework also depends on PowerPoint, and its ports to other platforms cannot represent some operations, such as text measurement and, on Google Slides, chart editing (Appendix~\ref{app:portability}).

Each system was run once per task, so our deck-level bootstrap intervals do not capture run-to-run variation (Appendix~\ref{app:stats}).

Our single-agent ablation keeps the tools fixed, so it isolates the dispatcher--specialist structure but not the effect of predefined tools relative to code generation. The comparison with Claude Code also differs in execution backend and prompting.

\section*{Ethical Considerations}
All decks in \bench{} were contributed by the authors or by colleagues who were informed of the research purpose and consented to their release in a public benchmark. We removed or anonymized personal and sensitive content before release. The rendering audit and the survey in Appendix~\ref{app:vision-audit} involved one undergraduate annotator and three non-author raters, who took part voluntarily after giving verbal consent and judged only rendered slides. To tell survey submissions apart, we recorded only a self-chosen nickname, a hashed network address, and the browser type. The vision validator is trained only on public decks released under CC BY or CC0 (Appendix~\ref{app:vision-data}). Because editing agents modify user content, deployments should keep the source file, log every edit, and avoid overwriting user content without confirmation. We used AI assistants for parts of the implementation and for language editing, and the authors checked all claims and results.

\bibliography{custom}

\appendix
\input{appendix}

\end{document}

%% file: appendix.tex
\section*{Appendix}
\raggedbottom
\renewcommand{\dbltopfraction}{0.85}
\renewcommand{\dblfloatpagefraction}{0.8}

\section{Framework Details}
\label{app:framework}
Through COM, the parser reads the text, formatting, and structure that an editor needs, already resolved the way PowerPoint displays them. COM does not expose all of this information, however, and some of what it reports is unreliable. It does not expose the structure of equations or the content of pictures, it cannot serve several slides in parallel, and its text ranges can report formatting incorrectly or absorb inserted text into a neighboring equation. The parser and the tools therefore add procedures that fill these gaps. The subsections below describe the slide representation and its size, the input-size limits of a baseline that reads raw slide XML, the tool set, and these procedures.

\subsection{Parser Output}
\label{app:parser-extraction}
The parser turns each slide into one record that lists every shape with the properties an editor needs to find and change it. The record begins with slide-level properties such as layout, background, transition, and notes. Each shape then carries its ID, name, type, position, size, stacking order, fill, and line, and a text shape also carries its full text with its paragraphs and runs and their resolved formatting. Tables, charts, and groups are expanded into their cells, chart properties, and child shapes. Two indexes summarize the slide for the agents. One maps layout placeholders to shapes, and the other assigns shapes to visual roles such as title, body, footer, and page number, which helps resolve a request such as ``every slide's title'' when the title is an ordinary text box.

\paragraph{Pictures.}
COM does not report what a picture shows, so the parser captions every picture once with a VLM, including pictures inside groups and picture placeholders. A caption gives a one-sentence description and states whether the picture contains text, in which languages, and with a short sample of that text, which lets the agents decide whether a request such as translation applies to the picture. Each picture is captioned once per session, and the caption is reused across re-parses and rollbacks.

\begin{figure}[t]
\centering
\begin{subfigure}{\columnwidth}
\centering
\includegraphics[width=\columnwidth]{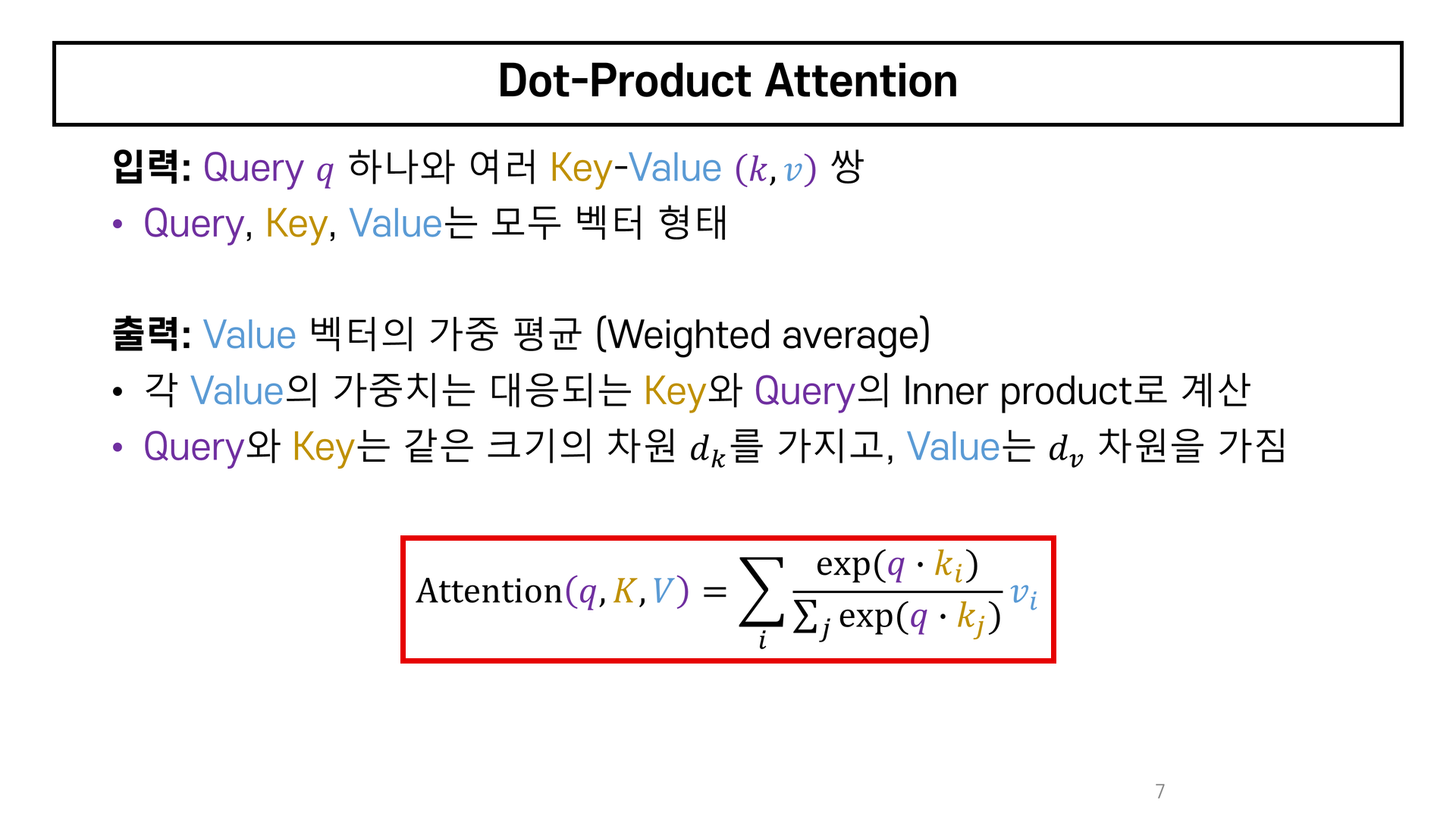}
\caption{Original slide.}
\end{subfigure}

\vspace{0.4em}
\begin{subfigure}{\columnwidth}
\centering
\includegraphics[width=\columnwidth]{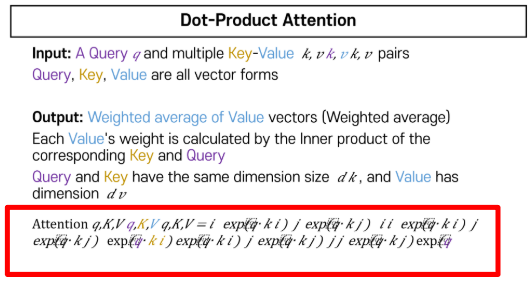}
\caption{Earlier version without equation-aware editing.}
\end{subfigure}

\vspace{0.4em}
\begin{subfigure}{\columnwidth}
\centering
\includegraphics[width=\columnwidth]{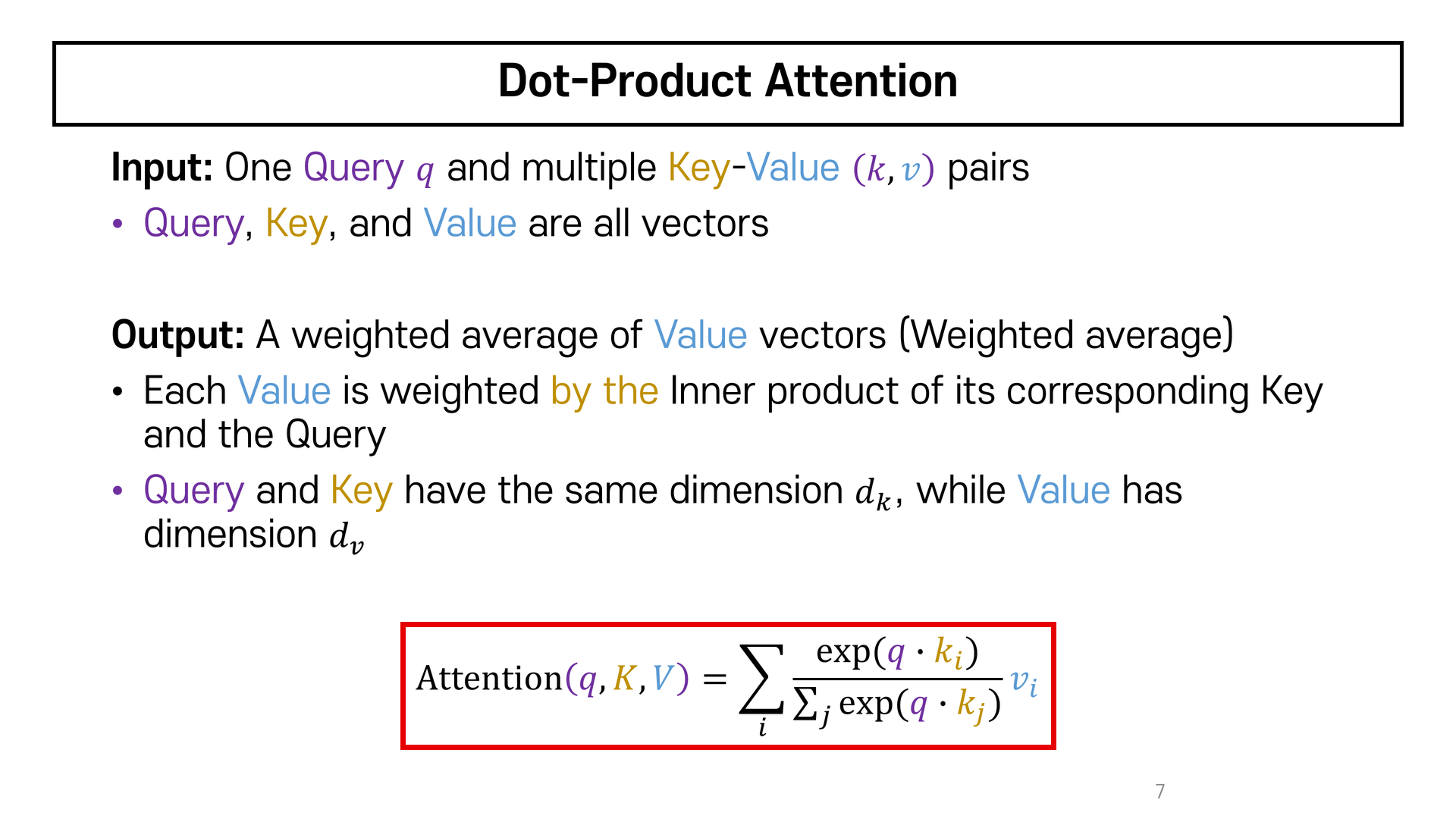}
\caption{Current framework with gpt-6-sol.}
\end{subfigure}
\caption{The equation slide of a translation task. The earlier version turns the equation into plain characters and repeats the inline symbols, while the current framework keeps both intact.}
\label{fig:equation-broken}
\end{figure}

\begin{figure}[t]
\centering
\includegraphics[width=\columnwidth]{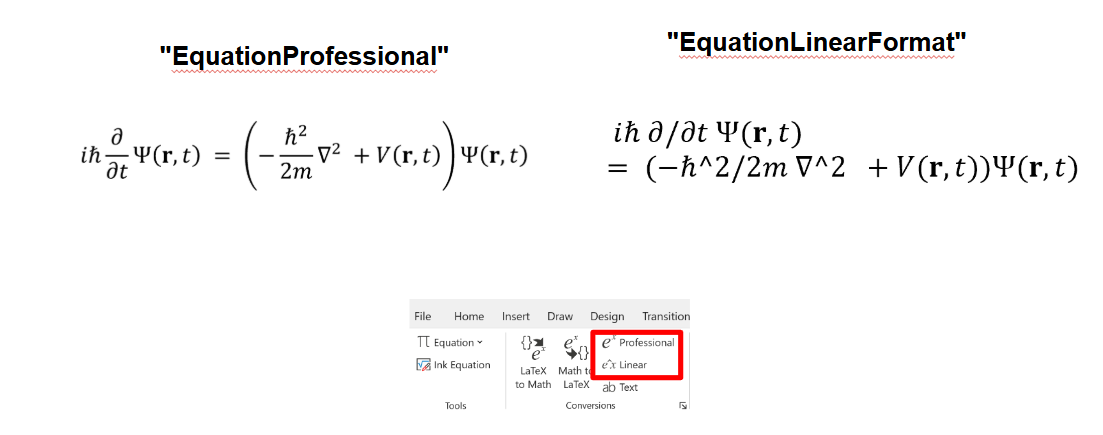}
\caption{The two forms in which PowerPoint shows an equation. New equations are inserted in the linear form and then converted to the built-up professional form.}
\label{fig:equation-forms}
\end{figure}

\paragraph{Equations.}
COM shows the text of an equation but not its structure, and ordinary text operations on a text box that contains an equation can break the equation, as in Figure~\ref{fig:equation-broken}(b). The parser therefore reads equations in text boxes and table cells from the file. It saves a copy of the open deck, reads each equation from the copy, converts it to LaTeX, and places it in the shape's text between dollar signs, so the model reads and writes equations in a familiar notation. The copy is saved again whenever an edit or a rollback may have changed the deck. Edits still go through COM. The text tools map each position in this notation back to the live text and change only the parts that differ, so equations the edit does not concern are left untouched. Text is inserted next to an equation so that it stays outside the equation, and an equation is created or rebuilt only when the edit changes it, by converting its LaTeX to PowerPoint's linear equation format and inserting it with PowerPoint's own equation commands, which convert it to the displayed professional format (Figure~\ref{fig:equation-forms}). With these procedures, the same translation keeps the equation and its inline symbols intact (Figure~\ref{fig:equation-broken}(c)).

\subsection{Token Footprint}
\label{app:parser-tokens}
Table~\ref{tab:token-footprint} compares the size of the parser output with the raw OOXML of the same slides. We measure the 570 of the 582 slides in \bench{} that have a cached parse from the final parser version, counting tokens with the o200k\_base tokenizer. Raw XML is measured either as the slide part alone or in full scope, where the slide's relationship file and the chart and diagram parts it references are added; media files are excluded. In total, the parser output is 3.6 times smaller than the full-scope XML, a 72.4\% reduction. Per slide the reduction is smaller and varies: the median slide shrinks by 56.3\% (10th--90th percentile 26.3--75.3\%), and 11 slides (1.9\%) grow. The total exceeds the median because the slides with the most XML shrink the most. The agents receive a further filtered form of the parser output, which we do not measure here.

\begin{table}[t]
\centering
\small
\setlength{\tabcolsep}{4pt}
\begin{tabular}{l rrr}
\toprule
Input representation & Tokens & Est.\ cost & Efficiency \\
\midrule
\editppt{} parser & \textbf{2.14M} & \textbf{\$4.27} & \textbf{3.63$\times$} \\
Raw XML, slide part only & \underline{7.61M} & \underline{\$15.22} & \underline{1.02$\times$} \\
Raw XML, full scope & 7.74M & \$15.49 & 1.00$\times$ \\
\bottomrule
\end{tabular}
\caption{Token and cost comparison between the \editppt{} parser output and raw OOXML over 570 parsed slides. Full scope adds the slide's relationship file and the chart and diagram parts it references to the slide XML. Estimated cost assumes gpt-6-sol input pricing (\$2.00 per million tokens) without caching. Efficiency is relative to full-scope raw XML; higher is better.}
\label{tab:token-footprint}
\end{table}

\subsection{Low-Level Inputs in Practice: PPTPilot}
\label{app:xml-context}
Real slides often carry more XML than PPTPilot passes to its model, and many of its tasks that end without a usable deck involve such slides. PPTPilot routes each task either to python-pptx code generation or to direct editing of slide XML. For direct editing it cuts any slide longer than 30{,}000 characters down to its first and last 15{,}000 characters, and 132 of the 582 slides in \bench{}, spread over 19 of the 28 decks, exceed this length. As Table~\ref{tab:pptpilot} shows, the XML route completes about as many tasks as the code route when the target slides fit within the limit, but rarely succeeds when a target slide exceeds it. With gpt-6-sol, 25 of the 26 XML-route tasks that ended without a usable deck had such a slide. When the model writes a cut slide back, content can also be lost. With Qwen3.8-27B, one slide with 159 shapes came back with a single shape.

The code route avoids this cut but puts a summary of the whole deck into the prompt. On four long decks with dense content, the summary exceeds the 262k-token context window with which we served Qwen3.8-27B, so 19 tasks fail before any edit is attempted, whereas the gpt-6 models accept the same prompts. Malformed XML plays a smaller role and leaves four tasks without a deck with gpt-6-luna and one with Qwen3.8-27B. These observations suggest that input size is a practical limit for low-level editing on real decks, which motivates the compact representation of Section~\ref{sec:method-parser}. Because PPTPilot's router chooses the route for each task, the routes are not compared on the same tasks.

\begin{table}[t]
\centering
\small
\setlength{\tabcolsep}{4pt}
\begin{tabular}{l cc cc}
\toprule
 & \multicolumn{2}{c}{Code route} & \multicolumn{2}{c}{XML route} \\
\cmidrule(lr){2-3}\cmidrule(lr){4-5}
Backbone & $\le$30k & $>$30k & $\le$30k & $>$30k \\
\midrule
gpt-6-sol & 31/45 & 31/63 & 28/41 & 4/34 \\
gpt-6-luna & 23/44 & 32/64 & 22/42 & 2/33 \\
Qwen3.8-27B & 27/45 & 28/65 & 31/41 & 6/30 \\
\bottomrule
\end{tabular}
\caption{PPTPilot tasks completed out of tasks routed, split by route and by whether any target slide exceeds 30{,}000 characters of XML. Two Qwen3.8-27B tasks ended before routing.}
\label{tab:pptpilot}
\end{table}

\subsection{Tool Set}
\label{app:tools}
The 37 tools are divided among the five specialists by the kind of object they edit, as summarized in Table~\ref{tab:tools}. Each tool performs one PowerPoint operation through COM on slides and shapes identified by number or ID, with arguments of fixed types, and no tool executes code written by a model. A few tools call a model inside the operation. The text replacement and rewriting tools use a style mapper to carry the formatting of the original runs over to the new text, and the image tools can search for, generate, or edit a picture with an external service. In the single runs with Qwen3.8-27B, Sonnet~5, and Haiku~4.5, 32 of the 37 tools appear in at least one retained edit. The five unused tools all belong to the shape layout specialist and create placeholders, duplicate or copy shapes, apply gradient fills, or insert pictures.

\begin{table}[t]
\centering
\small
\setlength{\tabcolsep}{4pt}
\begin{tabular}{l r p{0.57\columnwidth}}
\toprule
Specialist & Tools & Operations \\
\midrule
Text and style & 7 & Character formatting, inserting, deleting, replacing, and rewriting text, paragraph alignment, bullets \\
Table & 3 & Cell text style, cell text replacement, table layout and borders \\
Chart & 6 & Categories, series, chart structure, axes, colors, a new chart from data \\
Shape layout & 15 & Moving and resizing, aligning and distributing, reordering items into slots, creating text boxes, placeholders, and shapes, deleting, duplicating, and copying shapes, visual styles and gradient fills, inserting, replacing, and editing pictures \\
Slide & 6 & Adding, deleting, duplicating, and moving slides, transitions, backgrounds \\
\bottomrule
\end{tabular}
\caption{Tools available to each specialist.}
\label{tab:tools}
\end{table}

\subsection{Supporting Algorithms}
\label{app:algorithms}
Structured editing in \editppt{} rests not only on the components of Section~\ref{sec:method-parser}--\ref{sec:method-validation} but also on many smaller procedures that keep the model's view of a slide faithful and its actions predictable. Each procedure moves a decision that a model would otherwise make freely, such as how text is split into runs, where inserted text lands, or which shapes an edit may touch, into code that behaves the same way every time. The model decides what should change, while deterministic procedures decide how the slide is represented to it, how the change is executed, and how the result is checked. Tables~\ref{tab:algorithms} and~\ref{tab:algorithms-2} list the procedures of the final framework, and the paragraphs below describe the main ones by stage.

\begin{table*}[tp]
\centering
\small
\setlength{\tabcolsep}{4pt}
\renewcommand{\arraystretch}{1.05}
\begin{tabular}{p{0.24\textwidth} p{0.72\textwidth}}
\toprule
Procedure & What it does \\
\midrule
\multicolumn{2}{l}{\emph{Parsing}} \\
Run reconstruction & Compares formatting character by character and starts a run wherever it changes, avoiding mixed-state and empty values that COM returns over ranges \\
Paragraph records & Records alignment, indent, spacing, and bullet labels per paragraph and assigns each paragraph the runs inside it \\
Merged cells & Groups table cells with the same position and size into one cell with row and column spans \\
Group expansion & Parses group children recursively with the same fields as top-level shapes \\
Placeholder index & Maps placeholder types to shapes, and treats a picture placeholder as a picture \\
Visual-role index & Marks titles, subtitles, body text, footers, and page numbers from placeholder types, names, position, and font size \\
Value normalization & Converts COM colors to RGB, filters invalid values, and reads attributes defensively so that one failure does not stop a slide \\
Picture captions & Describes each picture once with a VLM, including whether it contains text and in which languages \\
Equation notation & Reads equations from a saved copy of the deck and shows them inline as LaTeX \\
Parse cache & Reuses slide parses, forces a fresh parse on retries, and discards parses of slides that shift \\
\midrule
\multicolumn{2}{l}{\emph{Target resolution}} \\
Planner modes & Writes per-slide tasks or templates expanded with slide-number arithmetic, and checks slide numbers against the current deck \\
Deck outline & Gives the planner one line per shape for every slide so that targets can be found by content \\
Plan recall & Adds slides the plan missed that contain a quoted term of the request or source-language text in a translation \\
Slide identity & Binds each task to a stable slide identifier and rebinds later tasks to slides that earlier tasks created \\
Dispatch view & Shows the dispatcher each shape's full text, geometry, role, and children, and splits a task into specialist subtasks \\
Dispatch checks & Re-asks once when the dispatcher returns no shapes or unreadable output, with fallbacks for routing \\
Scope rules & Adds shapes of a named role or body text that the dispatcher left out, and replaces its choice when the role is certain \\
Scoped view & Shows a specialist only its assigned shapes, with previews of other text shapes as read-only context \\
Scope widening & Adds shapes that the validator says should have been edited, if the specialist can act on them \\
Prefetching & Parses upcoming slides and runs their dispatch ahead of time, overlapping model calls with COM work \\
\bottomrule
\end{tabular}
\caption{Supporting procedures for parsing and target resolution.}
\label{tab:algorithms}
\end{table*}

\paragraph{Parsing.}
The parser rebuilds what COM reports incompletely. Formatting read over a range of characters is unreliable, since a bold or italic flag over mixed text returns a ``mixed'' value that a direct read takes as set, and the font name over a range can come back empty even when every character in it reports the same font. The parser therefore reads the formatting of each character and starts a new run wherever it changes, keeping positions both in the UTF-16 units that COM expects and in the code points that the model sees. Table cells carry no merge span, so cells that share the same position and size are grouped into one logical cell with row and column spans. Groups are expanded recursively, so their children carry the same fields as top-level shapes. A visual-role index marks titles, subtitles, body text, footers, and page numbers, first from placeholder types and shape names and then from position and font size when a title or a footer is an ordinary text box. After an edit, cached parses of the affected slides are discarded, and slides that move are tracked by their stable identifiers.

\paragraph{Target resolution.}
Several procedures keep the chosen targets complete and stable. The planner reads a text outline of the whole deck, so it can find slides by their content, and pattern tasks are expanded with slide-number arithmetic and checked against the current slide count. A recall step then adds slides that the plan missed but that contain a quoted term from the request or, for a translation, text in the source language. Each task is bound to its slide's stable identifier, so later tasks still find their slide after slides are added, deleted, or moved. The dispatcher sees every shape with its full text, position, size, and role, and deterministic scope rules check its choice. When a style request names the body text or a role such as the title, the rules add shapes of that role that the model left out, and they replace the model's choice when the role is certain from the placeholder type. A specialist then sees only its assigned shapes, with short previews of the other text shapes as read-only context.

\begin{table*}[tp]
\centering
\small
\setlength{\tabcolsep}{4pt}
\renewcommand{\arraystretch}{1.05}
\begin{tabular}{p{0.24\textwidth} p{0.72\textwidth}}
\toprule
Procedure & What it does \\
\midrule
\multicolumn{2}{l}{\emph{Execution}} \\
Style mapper & Carries the formatting of the original runs over to rewritten or translated text \\
Paragraph reconciliation & Rebuilds the mapper's output when it returns a different number of paragraphs \\
Paragraph assignment & Writes each paragraph in one assignment so that its paragraph format survives \\
Conditional writes & Writes a formatting value only when it differs, keeping inherited formatting inherited \\
Unrequested arguments & Drops style arguments that the request does not mention \\
Range resolution & Moves model-given character ranges to the nearest exact match and converts them to UTF-16 \\
Replacement by assignment & Replaces text without deleting first, which in COM can also remove a neighboring space \\
Color-conditional scope & Restricts a change to text of a named color and fails when nothing matches \\
Text fitting & Shrinks text that outgrows its box within a lower bound and keeps targeted shapes inside the slide \\
Write-scope guard & Restores out-of-scope shapes that moved and rejects calls that change their text or delete them \\
Argument validation & Removes unknown arguments and rejects invalid slide numbers, shape IDs, and shape types before execution \\
Picture guard & Blocks picture tools unless the request mentions pictures and, if named, the slides \\
Coverage continuation & Asks a text specialist to continue when assigned shapes or untranslated paragraphs remain, and removes repeated calls \\
Platform fixes & Handles COM behaviors of tables, charts, layouts, placeholders, and slide operations \\
\midrule
\multicolumn{2}{l}{\emph{Validation and recovery}} \\
Snapshot diff & Stores the slide parse before a task and shows the validator only what changed after each attempt \\
Structured verdict & Asks the validator for named wrong and expected shapes, given the scope and the automatic adjustments \\
Verdict checks & Re-asks on unreadable verdicts and accepts rejections whose reasons say that nothing remains to do \\
Rollback decision & Rolls back an attempt that edited a flagged shape and otherwise continues from the partial result \\
Feedback & Adds a directive and a shape inventory from each failed attempt to the next prompt \\
Flip stop & Stops retrying when the verdict on a shape flips between attempts \\
Best attempt & Keeps the failed attempt with the fewest missed targets and no wrong edits, or restores the previous state \\
File rollback & Restores the deck from a saved copy and restarts PowerPoint if COM stops responding \\
Slide checks & Verifies slide additions, deletions, moves, transitions, and backgrounds directly through COM \\
Task isolation & Logs and skips a failing task so that the rest of the request still runs \\
\bottomrule
\end{tabular}
\caption{Supporting procedures for execution, validation, and recovery, continuing Table~\ref{tab:algorithms}.}
\label{tab:algorithms-2}
\end{table*}

\paragraph{Execution.}
The tools keep each edit within the requested scope and make only a few predefined adjustments of their own. A tool writes a formatting value only when it differs from the current one, because rewriting an equal value would turn inherited formatting into explicit formatting that no longer follows the layout. Style arguments that the request does not mention are dropped before the call. Character ranges given by the model are checked against the actual text and moved to the nearest exact match, and replacements assign the new text to the range instead of deleting and inserting, which in COM can also remove a neighboring space. When text is rewritten, as in a translation, a style mapper carries the formatting of the original runs over to the new text, the number of paragraphs is reconciled when the mapper merges or splits them, and each paragraph is written in one assignment so that its paragraph format survives. Text that outgrows its box is shrunk within a lower bound, and only the shapes a call targets are moved back inside the slide. A write-scope guard compares every shape on the slide before and after each tool call. It restores shapes outside the call's scope that moved and rejects the call when such a shape changed its text or disappeared. Calls with invalid arguments, picture edits on requests that do not mention pictures, and repeated calls are filtered out before execution, and a text specialist that stops before covering every assigned shape, or every paragraph in a translation, is asked to continue.

\paragraph{Validation and recovery.}
Validation compares structured snapshots instead of asking a model to inspect a slide freely. The parse of the slide is stored before a task starts and taken again after each attempt, and the validator sees in full only the shapes that changed, together with the request, the assigned scope, and the adjustments that the tools made automatically, so that these are not blamed on the agent. It returns a structured verdict that names the shapes that should have changed and those that should not have. Code then decides how to continue. When the verdict names shapes that the attempt edited, the attempt is rolled back, and when it names shapes the attempt left alone, the partial progress is kept and the next attempt continues from it. Each failed attempt adds a directive to the feedback for the next one, and retries stop early when the verdict on a shape flips between attempts. When all attempts fail, the attempt with the fewest missed targets and no wrong edits is kept if there is one, and the previous state is restored otherwise. Slide-level operations such as adding, moving, or deleting slides are checked directly through COM rather than by the validator.

\FloatBarrier
\section{Benchmark Details}
\label{app:benchmark}

\subsection{Task Construction}
\label{app:task-construction}
The tasks were written for real decks so that they reflect realistic editing requests. The 28 human-authored decks were provided by nine contributors from different fields, who consented to their release. The decks are lectures, seminars, paper reviews, proposals, and reports on AI and speech research (12 decks), STEM teaching and engineering (4), the humanities and general topics (4), business (3), architecture, urban planning, and visual design (3), and medicine and biology (2), and 21 of them are in English and 7 in Korean. The authors went through each deck and wrote requests suited to its content, favoring edits that repeat across slides, combine several changes, or leave room for interpretation. Instructions name their targets by content or function rather than by exact strings, as in ``On slide 5, delete the citation textbox,'' so a system has to find the targets itself.

Each task is classified by whether its instruction points to a single slide (explicit) or describes its targets by a pattern that may cover several slides (pattern), and by whether each target slide involves one target and edit (simple) or several (compound). The benchmark has 17 explicit-simple, 57 explicit-compound, 87 pattern-simple, and 22 pattern-compound tasks, and 92 of the 183 instructions instantiate 14 reusable templates while 91 are specific to their deck. The annotation of a task describes the end state it requires through target and acceptable shapes, checks, and preservation clauses, and it assumes no particular tool or execution path. Each annotation was checked against the outputs of all systems before it was finalized, and errors found in this way were corrected in the annotation itself, so the final criteria apply to every system in the same way, without task-specific exceptions for side effects.

\subsection{Statistics by Deck Length}
\label{app:tier-stats}
Task scope grows with deck length, not only the number of slides. As Table~\ref{tab:tier-stats} shows, a task on a long deck targets several times more slides than a task on a short deck and involves about ten times as many candidate shapes, while the share of the deck that a task targets stays roughly constant across tiers. Long decks also carry more pictures and more shapes per slide than the other tiers. The deck-length results in Section~\ref{sec:res-length} may therefore reflect these differences in task composition as well as deck length itself.

\begin{table}[t]
\centering
\small
\setlength{\tabcolsep}{4pt}
\resizebox{\columnwidth}{!}{%
\begin{tabular}{l rrrr}
\toprule
 & Short & Medium & Long & All \\
\midrule
Decks & 9 & 12 & 7 & 28 \\
Slides & 46 & 241 & 295 & 582 \\
Tasks & 54 & 81 & 48 & 183 \\
Slides per deck, median & 5 & 21 & 43 & 18 \\
Shapes per slide & 14.3 & 8.5 & 21.8 & 15.7 \\
Pictures per slide & 1.09 & 1.47 & 3.23 & 2.34 \\
\midrule
Target slides per task & 2.3 / 1 & 8.6 / 6 & 18.8 / 9 & 9.4 / 3 \\
Tasks with $\ge$10 target slides & 0\% & 40\% & 50\% & 31\% \\
Candidate shapes per task & 50 / 23.5 & 76 / 43 & 500 / 165.5 & 179 / 39 \\
Tasks with $\ge$100 candidates & 20\% & 27\% & 52\% & 32\% \\
Targeted share of the deck & 56\% & 42\% & 45\% & 47\% \\
\bottomrule
\end{tabular}}
\caption{Benchmark statistics by deck length: short decks have at most 10 slides, medium decks 11--30, and long decks more than 30. Counts follow the rules of Table~\ref{tab:bench-compare}. Per-task values are mean / median. The targeted share is the mean fraction of a deck's slides that a task targets.}
\label{tab:tier-stats}
\end{table}

\subsection{Example Tasks}
\label{app:examples}
Table~\ref{tab:examples} shows instructions that illustrate the range of \bench{}. Many of them name their targets by role or by content rather than by slide number, so a system has to find the targets before it can edit them, and a short instruction can reach every slide of a deck. Others require reading one slide in order to edit another, translating only the passages in one language, or judging which content matters, and some act on tables, charts, pictures, slide order, or transitions rather than on text.

\begin{table*}[t]
\centering
\small
\setlength{\tabcolsep}{4pt}
\begin{tabular}{p{0.2\textwidth} p{0.6\textwidth} l r}
\toprule
Kind & Instruction & Deck & Targets \\
\midrule
Role across the deck & Change every slide's title color to green. & Short & 9 \\
Body text across the deck & Change every slide's body font to Times New Roman. & Medium & 22 \\
Condition on content & On every slide that mentions `MingTok-Audio', bold every occurrence of the term. & Medium & 10 \\
Condition and addition & For every slide whose title starts with `DEMO', change the title color to dark red and add a small textbox `Phase 2 Demo' at the top-right corner of the slide. & Short & 4 \\
Cross-slide reference & On slides 6-15, adjust each shape's position and size to match slide 5's layout. & Medium & 10 \\
Selective translation & Translate every Vietnamese-language passage in the deck into Korean. Leave English-only slides unchanged. & Medium & 6 \\
Open-ended content & On slide 8, summarize the body content and replace the original body with the summary. & Long & 1 \\
Table & On slide 15's table, apply a different text color to each column. & Long & 1 \\
Chart & On slide 4 (Work Environment Satisfaction chart), set the vertical axis title to `Share of respondents (\%)' and the axis range to 0 to 50. & Short & 1 \\
Layout & On slide 7, make the two charts the same size and align their tops. & Short & 1 \\
Slide order & Move the three `Supplement' slides so they come after the `Thank You' slide. & Medium & 3 \\
Pictures & On slides 5-8, reduce every figure's size by 30\%. & Long & 4 \\
Transitions & Apply a different transition to each slide so that no two consecutive slides share the same transition. & Medium & 30 \\
\bottomrule
\end{tabular}
\caption{Example instructions from \bench{}, exactly as given to every system. Deck is the length tier of the source deck, and Targets is the number of annotated target slides.}
\label{tab:examples}
\end{table*}

\subsection{Notes on the Benchmark Comparison}
All PPTX corpus statistics in Table~\ref{tab:bench-compare} are recomputed from the publicly released source decks of each benchmark with one XML-based counting rule. PPTArena reports 2{,}125 slides, and our count of 999 covers its released source decks. Its target-slide counts are estimated from the differences between original and reference decks, and its candidate-shape counts are derived from these inferred target slides. Of the 26 \bench{} tasks that use model judgments, three compare pictures.

In Table~\ref{tab:bench-compare}, a full mark means that a benchmark annotates targets at the object level or checks every other object for preservation, and a partial mark means that it does so only in part. PPTArena~\citep{ofengenden2025pptarenabenchmarkagenticpowerpoint} judges preservation within a holistic visual score rather than object by object. PPT-Eval~\citep{ppteval2026} writes targets and preservation into rubric code for each task, where many preservation checks are non-critical branches that do not fail the task. PPTBench~\citep{pptbench2025} annotates a single shape per task, and in its released implementation only font-change tasks compare the rest of the slide.

\FloatBarrier
\section{Annotation and Metrics}
\label{app:golden}

\subsection{Annotation Schema}
Each task is a single record that states what must change, what may change, and what must stay as it was, so scoring needs no reference deck. The record names the target slides and, on each of them, the target shapes by their PowerPoint shape IDs. Shapes on any slide that may change but need not are listed as acceptable. A list of checks describes the state that the target shapes must reach, and a preservation clause names the properties compared on every other shape. Figure~\ref{fig:golden-example} shows an abridged record.

The evaluator reads the original and edited decks directly from their OOXML packages in Python and does not use the COM parser of the framework. PowerPoint's COM interface cannot serve several requests in parallel, so parsing every output of the twelve systems through it would take long, whereas reading the files lets outputs be scored in parallel and on any platform. The evaluator resolves formatting inherited from layouts, masters, themes, and table styles itself, so formatting is compared by the value the slide displays rather than by the attributes written in the file. Preservation compares the geometry, text, character and paragraph formatting, equations, fills, tables, charts, and pictures of each shape, together with shape additions and deletions.

Whether a shape is unchanged is decided by an explicit list of differences to ignore rather than by byte equality. PowerPoint rewrites some attributes on every save, such as spell-check and language tags, and stores geometry in fine units, so the comparison ignores these attributes, geometry changes below one point, and size changes of text boxes that PowerPoint resizes automatically. Other differences in the compared properties count as changes, and Appendix~\ref{app:evaluator-validity} reports how we checked this list. Color names in instructions map to fixed ranges of hue, saturation, and lightness, so a request for green accepts any clearly green color rather than one exact value.

\begin{figure}[t]
\begin{lstlisting}[basicstyle=\ttfamily\scriptsize,columns=fullflexible,frame=single,breaklines=true]
{
 "instruction": "Change every slide's title color to green.",
 "targets": {
  "slides": [1, 2, 3, 4, 5, 6, 7, 8, 9],
  "shapes": {"1": ["14"], "2": ["5"], "3": ["11"], ...},
  "acceptable": {"1": ["13"]}
 },
 "checks": [{"type": "color_in", "range": "green"}],
 "preservation": {
  "scope": "non_target_shapes",
  "compare": ["geometry", "text", "run_style",
              "math", "presence"]
 }
}
\end{lstlisting}
\caption{Abridged annotation of one task. On the cover slide, the English line under the Korean title is acceptable, so it may turn green as well.}
\label{fig:golden-example}
\end{figure}

\subsection{Check Types}
\label{app:check-types}
Most checks compare a property of the annotated target shapes with the original deck, and only outcomes without a single correct form are left to a model. Table~\ref{tab:check-types} groups the 69 check types by what they inspect. Tables, charts, and slide-level properties have their own types because their state lives in structures that text checks do not reach. Seven types rely on a model judgment, covering translation, summaries, the choice of important content, and two checks on pictures. Even in these tasks, rules decide whether the right shapes were edited and whether anything else changed, and a picture that was deleted or left as it was fails before any model is called.

\begin{table*}[t]
\centering
\small
\setlength{\tabcolsep}{4pt}
\begin{tabular}{l rr p{0.62\textwidth}}
\toprule
Category & Types & Tasks & What the checks verify \\
\midrule
Text content & 8 & 33 & Text matches a pattern, a rewrite rule applied to the original text, numbering prefixes, no characters of the source language left after translation \\
Character style & 7 & 33 & Bold, italic, underline, font, or size on whole shapes or on selected terms, and emphasis patterns or fonts kept through a rewrite \\
Color and fill & 8 & 29 & Text or fill color within a named range, recoloring from one range to another, distinct or ordered colors across shapes \\
Objects and pictures & 5 & 37 & Shapes added in a region with given text, shapes removed, pictures scaled, recolored, or replaced \\
Geometry and arrangement & 6 & 8 & Alignment, even spacing, equal size, matching another shape's geometry, reordering or swapping items \\
Tables & 11 & 12 & Header and row styles, alternating row colors, column colors, cell text, merged cells, added columns \\
Charts & 10 & 15 & Chart type, title, legend, axes, categories, series, data point colors, new charts \\
Slides and deck & 7 & 15 & Transitions, backgrounds, slide order, added, duplicated, or removed slides \\
\midrule
Model-judged & 7 & 26 & Translation fidelity and run-level formatting, summaries, choice of important content, text inside pictures, replaced pictures \\
\bottomrule
\end{tabular}
\caption{The 69 check types grouped by what they inspect. A task can use types from several categories, so task counts do not add up to 183.}
\label{tab:check-types}
\end{table*}

Model judgments use gpt-6-luna at high reasoning effort, with one verdict per item and the same prompts for every system (Appendix~\ref{app:prompts-judge}). Because gpt-6-luna is also one of the evaluated backbones, its verdicts may favor its own outputs, a tendency reported for LLM judges~\citep{zheng2023judging}. Rules decide targeting and preservation before the model is called, so a model judgment only decides whether a requested change is complete. As a sensitivity check, we compare rankings on the two groups of tasks. Restricting Task Success to the 157 rule-only tasks leaves the ranking of all twelve systems unchanged, and on the 26 model-judged tasks the systems that use gpt-6-luna do not rank above those that use other backbones.

\subsection{Evaluator Consistency Checks}
\label{app:evaluator-validity}
We ran two diagnostic checks while developing the evaluator. The first tests for spurious changes, since a deck that is only opened and saved must score as unchanged. Re-saving all 28 decks through PowerPoint and through python-pptx without any edit left none of their 10{,}024 shapes flagged. The second tests whether a score depends on how a result is read. We scored the outputs of four systems on 34 tasks twice with the same annotations, once from the OOXML package and once from the parser's output for the same decks. The two scorings agreed on 98.2\% of the requested changes, 90.3\% of slide grades, 91.6\% of damage decisions, and 92.1\% of targeting decisions, and the disagreements revealed three errors in the OOXML scorer, which we fixed. Both checks used earlier versions of the parser and the scorer than the final ones.

\subsection{Metric Definitions}
\label{app:metrics}
Let $T$ be the set of tasks ($|T|=183$). For a task $t$, $S_t$ is the set of its target slides, each graded $g(s)\in\{\text{pass},\text{partial},\text{fail}\}$ as in Section~\ref{sec:eval-scoring}, and $W_t$ is the set of non-target slides that changed.
\begin{itemize}[leftmargin=*,itemsep=1pt,topsep=2pt]
    \item $C_t \subseteq S_t$: Target slides where at least one intended object was modified, regardless of editing success, or where the requested conditions were already satisfied.
    \item $d(s)=1$ for $s\in C_t$ if an acceptable or forbidden shape on $s$ changed beyond what the request allows, or a shape on $s$ was added or deleted without being requested; otherwise $d(s)=0$.
    \item $p(s)=1$ if $g(s)=\text{pass}$ and $d(s)=0$; otherwise $p(s)=0$.
\end{itemize}
Outputs are handled as follows. A task without a usable result, because the system returned no deck or a deck that cannot be read, has all of its target slides graded fail and $C_t=W_t=\emptyset$. A returned deck that leaves a target slide untouched gets a fail on that slide, which stays outside $C_t$ unless the slide already met the request, while every non-target slide it changes enters $W_t$. A deck identical to the input is judged the same way, so it fails every target slide that did not already meet the request and adds nothing to $W_t$. In our runs, all 22 decks returned unchanged fail every target slide. Target slides whose checks cannot be judged are left out of $S_t$, and Task Success requires at least one judged target slide. A non-target slide whose only changes are allowed changes to acceptable shapes belongs to neither $C_t$ nor $W_t$.
\begin{align*}
\text{Task Success} &= \tfrac{1}{|T|}\,\bigl|\{t : |S_t|>0,\ W_t=\emptyset,\\
&\qquad\qquad p(s)=1\ \forall s\in S_t\}\bigr|\\
\text{Slide Success} &= \frac{\sum_t |\{s\in S_t : p(s)=1\}|}{\sum_t |S_t|}\\
\text{IF} &= \frac{\sum_t |\{s\in S_t : g(s)\neq\text{fail}\}|}{\sum_t |S_t|}\\
\text{Over-edit} &= \frac{\sum_t |W_t|}{\sum_t \bigl(|C_t|+|W_t|\bigr)}\\
\text{Preservation} &= \frac{\sum_t |\{s\in C_t : d(s)=0\}|}{\sum_t |C_t|}
\end{align*}
Because $C_t$ also contains target slides that already met the request, the denominator of Over-edit counts correctly targeted slides rather than only the slides a system edited.

\paragraph{Worked example.}
The five metrics separate failure modes that a single success rate would merge. Consider a task with three target slides. A system completes the edit on the first slide cleanly, completes it on the second but also changes a footer there, does only part of it on the third, and changes one slide that is not a target. The task fails. Slide Success credits only the first slide, one of three, while Instruction Following credits all three, since each received at least part of the edit. Preservation is two of three, because only the second slide changed a shape it should have left alone, and Over-edit is one of the four slides the system edited.

\paragraph{What makes a slide partial.}
Across the twelve systems, 1{,}893 target slides are graded partial. In 90\% of them a requested change is incomplete, most often a translation or formatting change that the model-based check rejects on some target shapes. In the other 10\% every requested change is complete but a target also changed in another way, mostly styling that spills beyond the requested terms or a moved or resized box. A slide that is both incomplete and changed beyond the request counts as incomplete.

\paragraph{Shape-level scores.}
Two further metrics grade every annotated target shape on its own. A shape passes when the requested change is complete and nothing else about the shape changed, and it counts as partial when the change is incomplete or the shape also changed beyond the request. Shape Success is the share of target shapes that pass, and Shape IF is the share that pass or are partial, over all target shapes on judged target slides. Slide-level tasks such as transitions have no target shapes and do not enter these scores. As Table~\ref{tab:shape-metrics} shows, the shape-level scores rank the systems much as Slide Success and Instruction Following do. Our framework completes the largest share of target shapes without side effects, while Claude Code with Sonnet~5 at least partly edits slightly more of them, in line with its lead in Instruction Following.

\begin{table}[t]
\centering
\small
\setlength{\tabcolsep}{4pt}
\begin{tabular}{ll rr}
\toprule
System & Model & Shape Success & Shape IF \\
\midrule
\multirow{3}{*}{PPTPilot} & gpt-6-sol & 73.4 & 82.7 \\
 & gpt-6-luna & 54.3 & 70.0 \\
 & Qwen3.8-27B & 36.2 & 37.5 \\
\midrule
\multirow{2}{*}{TTYS} & gpt-6-sol & 71.7 & 79.0 \\
 & gpt-6-luna & 38.9 & 42.4 \\
\midrule
\multirow{2}{*}{Claude Code} & Sonnet 5 & 89.1 & \textbf{96.1} \\
 & Haiku 4.5 & 62.8 & 84.4 \\
\midrule
\multirow{5}{*}{\editppt{}} & gpt-6-sol & \underline{91.9} & 95.0 \\
 & gpt-6-luna & 89.3 & 92.9 \\
 & Qwen3.8-27B & 89.6 & 93.7 \\
 & Sonnet 5 & \textbf{92.5} & \underline{95.8} \\
 & Haiku 4.5 & 88.5 & 93.2 \\
\bottomrule
\end{tabular}
\caption{Shape-level scores on all 183 tasks, over the 3{,}388 target shapes on judged target slides. Best value per column in bold, second underlined.}
\label{tab:shape-metrics}
\end{table}

\subsection{Sensitivity to Acceptable Handling}
\label{app:sensitivity}
Table~\ref{tab:sensitivity} reports all twelve systems under the released annotation and under the two extreme readings of the acceptable category, in which acceptable shapes are either forbidden (strict) or required to change (required). Neither reading changes the comparisons that the statistics support. Every \editppt{} backbone still completes significantly more tasks than every PPTPilot and TTYS configuration under McNemar's test with Holm correction, and no backbone differs significantly from Claude Code with Sonnet~5 in Task Success under any reading. The order among the five strongest systems does shift within these non-significant differences. Claude Code with Sonnet~5 ranks second under the strict reading and first when acceptable shapes are required, since it changes acceptable shapes on more slides than our framework does.

\begin{table*}[t]
\centering
\small
\setlength{\tabcolsep}{4pt}
\begin{tabular}{ll ccc ccc ccc}
\toprule
 & & \multicolumn{3}{c}{Task Success} & \multicolumn{3}{c}{Slide Success} & \multicolumn{3}{c}{Preservation} \\
\cmidrule(lr){3-5}\cmidrule(lr){6-8}\cmidrule(lr){9-11}
System & Model & Released & Strict & Required & Released & Strict & Required & Released & Strict & Required \\
\midrule
\multirow{3}{*}{PPTPilot} & gpt-6-sol & 51.4 & 49.7 & 49.7 & 69.1 & 64.7 & 66.2 & \underline{98.3} & 91.8 & \underline{98.2} \\
 & gpt-6-luna & 43.2 & 41.5 & 42.6 & 59.0 & 57.4 & 57.6 & 96.3 & \textbf{94.2} & 96.2 \\
 & Qwen3.8-27B & 50.3 & 48.6 & 49.7 & 47.3 & 46.1 & 46.7 & 93.1 & 90.8 & 93.1 \\
\midrule
\multirow{2}{*}{TTYS} & gpt-6-sol & 35.5 & 33.9 & 33.9 & 64.8 & 63.1 & 60.7 & 93.6 & 91.0 & 93.8 \\
 & gpt-6-luna & 29.0 & 27.3 & 27.9 & 41.3 & 40.4 & 39.5 & 95.5 & 92.9 & 95.4 \\
\midrule
\multirow{2}{*}{Claude Code} & Sonnet 5 & 73.8 & \underline{70.5} & \textbf{72.1} & 81.5 & 77.7 & 78.9 & 93.9 & 89.7 & 93.7 \\
 & Haiku 4.5 & 53.6 & 52.5 & 52.5 & 54.4 & 52.5 & 52.2 & 84.7 & 82.4 & 84.6 \\
\midrule
\multirow{5}{*}{\editppt{}} & gpt-6-sol & \textbf{74.9} & \textbf{72.1} & \underline{71.6} & \textbf{89.0} & \textbf{84.6} & 83.8 & 98.1 & 92.8 & 98.1 \\
 & gpt-6-luna & 73.8 & 69.4 & 69.9 & 88.5 & 83.7 & 83.8 & \textbf{98.9} & \underline{93.7} & \textbf{99.0} \\
 & Qwen3.8-27B & 72.1 & 68.9 & 70.5 & 87.4 & 82.6 & \underline{83.9} & 97.4 & 92.4 & 97.5 \\
 & Sonnet 5 & \underline{74.3} & 68.9 & 71.0 & \underline{88.8} & \underline{84.1} & \textbf{84.6} & 97.6 & 93.1 & 97.7 \\
 & Haiku 4.5 & 66.7 & 64.5 & 65.6 & 83.8 & 79.1 & 79.8 & 98.0 & 92.9 & 98.0 \\
\bottomrule
\end{tabular}
\caption{Results under the released annotation and the two extreme readings of the acceptable category, on all 183 tasks. Only the 32 tasks with acceptable annotations are rescored. Best value per column in bold, second underlined.}
\label{tab:sensitivity}
\end{table*}
Under the strict reading, the Slide Success and Preservation gaps to Claude Code with Sonnet~5 narrow. On the 32 tasks with acceptable annotations, Claude Code changes acceptable shapes on 119 slides, 278 shapes in all, against 87 to 94 slides and 142 to 172 shapes for our framework. The annotation permits these changes, so they indicate more of the allowed changes rather than more errors. The strict reading counts them as damage, but 49 of the 119 slides already count as damaged under the released annotation, most of them in four body-font tasks in which Claude Code also restyled running headers, page numbers, section labels, or titles. Because a slide cannot be penalized twice, the strict reading removes more previously clean slides from \editppt{}, 75 to 90 depending on the backbone, than from Claude Code, 70. The Slide Success and Preservation advantages over Claude Code remain positive for every backbone under this reading, although their confidence intervals include zero for most backbones.

\FloatBarrier
\section{Implementation and Reproducibility}
\label{app:implementation}

\subsection{Models and Reasoning Settings}
\label{app:models}
Reasoning effort is concentrated where targets are decided. As Table~\ref{tab:reasoning} shows, the planner and the dispatcher run at high effort with every backbone, and so does the style mapper with the gpt-6 backbones, while the specialists, the text validator, and picture captioning run at the lowest setting each backbone allows. That setting is no reasoning for gpt-6, thinking turned off for Qwen3.8-27B, and the lowest effort level of Claude Code. Within a configuration, every role uses the same backbone, so one model serves the whole pipeline. The only exception is picture generation and editing, which few tasks request and which use gemini-3-pro-image-preview with every backbone. Vision validation is off in all main runs.

The gpt-6 models are called through the OpenAI API. Qwen3.8-27B is served with vLLM~\citep{kwon2023vllm} on local servers with NVIDIA A100 or H200 GPUs, so its runs have no API cost. Its runs in Table~\ref{tab:main} add up to about 53 hours of task time, and those of TTYS in Appendix~\ref{app:baselines} to about 100 hours. The Claude backbones are called through Claude Code in non-interactive mode, which the framework uses as its model interface: tool schemas are given in the prompt, and tool calls come back as structured output. Cost is computed from the token counts of every call at list prices (Table~\ref{tab:prices}), including picture generation, and for the Claude backbones it is the list-price cost that Claude Code reports. The parameter counts of the gpt-6 and Claude models are not publicly disclosed. Runs that drive PowerPoint execute on a Windows machine, one task at a time.

\begin{table}[t]
\centering
\small
\setlength{\tabcolsep}{4pt}
\begin{tabular}{l ccc}
\toprule
Role & gpt-6 & Qwen3.8-27B & Claude Code \\
\midrule
Planner & high & thinking on & high \\
Dispatcher & high & thinking on & high \\
Specialists & none & thinking off & low \\
Style mapper & high & thinking off & low \\
Text validator & none & thinking off & low \\
Picture captions & none & thinking off & low \\
\bottomrule
\end{tabular}
\caption{Reasoning setting of each role by backbone family.}
\label{tab:reasoning}
\end{table}

\begin{table}[t]
\centering
\small
\setlength{\tabcolsep}{2.8pt}
\begin{tabular}{l rrr}
\toprule
Model & Input & Cached input & Output \\
\midrule
gpt-6-sol & 2.00 & 0.20 & 10.00 \\
gpt-6-luna & 0.10 & 0.01 & 0.50 \\
gemini-3-pro-image-preview & 2.00 & -- & 120.00 \\
\bottomrule
\end{tabular}
\caption{List prices used for cost, in USD per million tokens. The gpt-6 prices are from the provider's pricing page on 2026-09-27. Costs for the Claude backbones are those reported by Claude Code.}
\label{tab:prices}
\end{table}

\subsection{Baseline Configurations}
\label{app:baselines}
Every system, including ours, receives only the instruction and the source deck for each task and never sees the annotations. The baselines keep the prompts and settings of their repositories except for the model substitutions and harness adjustments described below.

\paragraph{Claude Code + PPTX skill.}
We run Claude Code 2.1.284 without its interactive interface, with Anthropic's PPTX skill installed and either claude-sonnet-5 or claude-haiku-4-5 as the model. The instruction is followed by a fixed note that names the deck file in the working directory and asks the agent to edit it in place without asking questions. Tool permissions are bypassed so that the agent can run commands without confirmation, and no limit is set on turns or budget. Cost is the list-price estimate that Claude Code reports for each run.

\paragraph{PPTPilot.}
We run the PPTPilot agent from the PPTArena repository at commit 7072dc9 and replace its main model with each backbone. The repository uses gpt-5-nano as router and slide planner whenever the main model is an OpenAI model, and the gpt-6 runs keep this setting. In the Qwen3.8-27B run, the router and planner also use Qwen3.8-27B so that no proprietary model takes part. Qwen3.8-27B is served with vLLM~\citep{kwon2023vllm} with a context window of 262{,}144 tokens and thinking enabled, and each task is stopped after one hour. All other settings, including the 120-second limit on generated scripts and the single retry, follow the repository.

\paragraph{Talk to Your Slides.}
We run TTYS at commit 8514945 with one gpt-6 model for every stage, in place of Gemini 1.5 Flash for planning and GPT-4.1 mini for the later stages, at the provider's default reasoning setting. Our harness passes the slide count of the open deck to the planner and saves the deck even when a later step fails, both of which help TTYS complete more tasks.

\paragraph{TTYS with Qwen3.8-27B.}
We also run TTYS at the same commit with Qwen3.8-27B for every stage, with thinking enabled for every call, a thinking budget of 4{,}096 tokens, at most 65{,}536 output tokens per call, and a limit of two hours per task. These runs take 39.5 minutes per task on average, and 17 tasks reach the time limit, so we stopped after 152 of the 183 tasks, which took about 100 hours. Of these 152 tasks, 36 ran on an H200 server and the rest on A100 servers. The 31 tasks left out come from five long decks and make up 31 of the 48 long-deck tasks. A task that was never run is not a failure of the system, so we leave these runs out of Table~\ref{tab:main} and compare all systems on the 152 tasks instead (Table~\ref{tab:ttys-qwen}). On these tasks, TTYS with Qwen3.8-27B returns a modified deck for 105 tasks and has the lowest Task Success and the highest Over-edit of all systems. Its Task Success of 16.4 is far below the 75.0 of our framework and the 53.9 of PPTPilot with the same backbone. Because the subset leaves out most long-deck tasks, every other configuration also reaches a higher Task Success on it than on all 183 tasks.

\begin{table}[t]
\centering
\small
\setlength{\tabcolsep}{3.5pt}
\resizebox{\columnwidth}{!}{%
\begin{tabular}{ll c rrrrr}
\toprule
System & Model & Modified & \textbf{TS}\,$\uparrow$ & Slide\,$\uparrow$ & IF\,$\uparrow$ & Over-edit\,$\downarrow$ & Pres\,$\uparrow$ \\
\midrule
\multirow{3}{*}{PPTPilot}
 & gpt-6-sol   & 126/152 & 54.6 & 70.2 & 86.7 & 0.87 & 98.3 \\
 & gpt-6-luna  & 119/152 & 46.7 & 66.0 & 81.2 & 0.90 & 95.0 \\
 & Qwen3.8-27B & 131/152 & 53.9 & 51.3 & 58.6 & 0.43 & 94.4 \\
\midrule
\multirow{3}{*}{TTYS}
 & gpt-6-sol   & 145/152 & 36.8 & 61.8 & 73.1 & 7.16 & 93.4 \\
 & gpt-6-luna  & 140/152 & 32.2 & 39.7 & 47.2 & 6.57 & 97.3 \\
 & Qwen3.8-27B & 105/152 & 16.4 & 18.6 & 21.9 & 10.31 & 92.0 \\
\midrule
\multirow{2}{*}{Claude Code}
 & Sonnet 5    & 152/152 & \underline{77.0} & 79.7 & \textbf{98.6} & 1.04 & 91.8 \\
 & Haiku 4.5   & 148/152 & 55.9 & 63.2 & 84.4 & 1.08 & 93.8 \\
\midrule
\multirow{5}{*}{\editppt{}}
 & gpt-6-sol   & 152/152 & \textbf{78.3} & \textbf{88.3} & 95.4 & 0.54 & 98.0 \\
 & gpt-6-luna  & 152/152 & 75.7 & 87.4 & 95.1 & 0.54 & \textbf{98.7} \\
 & Qwen3.8-27B & 152/152 & 75.0 & \underline{87.8} & 96.7 & \textbf{0.26} & 96.8 \\
 & Sonnet 5    & 152/152 & 75.7 & \textbf{88.3} & \underline{96.9} & \underline{0.35} & 97.5 \\
 & Haiku 4.5   & 152/152 & 69.1 & 86.0 & 94.1 & 0.36 & \underline{98.4} \\
\bottomrule
\end{tabular}}
\caption{All systems on the 152 tasks that TTYS with Qwen3.8-27B ran. The 31 tasks left out come from five long decks. Columns as in Table~\ref{tab:main}. Best value per quality metric in bold, second underlined.}
\label{tab:ttys-qwen}
\end{table}

\FloatBarrier
\section{Statistical Analysis}
\label{app:stats}
Tasks on the same deck share content and are correlated, so we resample decks rather than tasks. Every interval is a 95\% percentile interval from a cluster bootstrap that draws the 28 decks with replacement 10{,}000 times and recomputes each metric from the pooled counts of the drawn tasks~\citep{efron1993bootstrap}. A difference between two systems is computed on the same draws for both, so its interval accounts for the pairing. For Task Success we also run an exact two-sided McNemar test on the tasks that exactly one of the two systems completes~\citep{mcnemar1947}, and apply Holm's correction over the 66 pairs~\citep{holm1979}. McNemar's test treats tasks as independent, so we also run an exact deck-level sign-flip test, which compares the per-deck differences in completed tasks with all $2^{28}$ ways of swapping the two systems within decks, and apply the same correction. After correction, the two tests agree on 62 of the 66 pairs, and on the other four only McNemar's test finds a significant difference. As in Table~\ref{tab:main}, a task without a usable result counts as a failure. The intervals are not adjusted for multiple comparisons. Each system was run once per task, so the intervals reflect which tasks were sampled, not run-to-run variation.

Table~\ref{tab:ci-systems} adds intervals to every quality value in Table~\ref{tab:main}, and Tables~\ref{tab:ci-pairs} and~\ref{tab:ci-pairs-other} compare every pair of systems. Three findings follow. First, \editppt{} with gpt-6-sol, gpt-6-luna, Qwen3.8-27B, or Sonnet~5 completes significantly more tasks than every PPTPilot and TTYS configuration and than Claude Code with Haiku~4.5, with Holm-corrected $p<0.01$ under both tests. With Haiku~4.5, \editppt{} is significantly ahead of both TTYS configurations and of PPTPilot with gpt-6-luna under both tests, and of PPTPilot with gpt-6-sol and Qwen3.8-27B under McNemar's test only. Against Claude Code with the same Haiku~4.5, it completes 13.1 points more (CI $[3.4, 22.3]$, McNemar $p{=}0.003$), which does not survive the correction over 66 pairs ($p_{\text{Holm}}{=}0.079$). Second, Task Success does not separate the five strongest systems, \editppt{} with gpt-6-sol, gpt-6-luna, Qwen3.8-27B, or Sonnet~5 and Claude Code with Sonnet~5: the smallest McNemar $p$ among these ten pairs is 0.47, and every difference interval contains zero. Third, the faithfulness metrics do separate them. Every \editppt{} backbone, Haiku~4.5 included, has higher Preservation than Claude Code with Sonnet~5 ($+3.5$ to $+5.1$ points), with intervals that exclude zero. The four other backbones are also ahead on Slide Success ($+5.9$ to $+7.5$), and gpt-6-sol, Qwen3.8-27B, and Sonnet~5 have a significantly lower Over-edit rate. Claude Code's higher IF is significant only against \editppt{} with Haiku~4.5. The pattern is the same when both systems use Sonnet~5, which suggests that it is associated with the framework rather than with the backbone alone.

\begin{table*}[t]
\centering
\small
\setlength{\tabcolsep}{4pt}
\resizebox{\textwidth}{!}{%
\begin{tabular}{ll ccccc}
\toprule
System & Model & \textbf{TS}\,$\uparrow$ & Slide\,$\uparrow$ & IF\,$\uparrow$ & Over-edit\,$\downarrow$ & Pres\,$\uparrow$ \\
\midrule
\multirow{3}{*}{\textbf{PPTPilot}}
 & gpt-6-sol & 51.4 {\scriptsize[42.5, 60.1]} & 69.1 {\scriptsize[62.9, 76.6]} & 83.3 {\scriptsize[76.9, 90.3]} & 0.68 {\scriptsize[0.19, 1.26]} & \underline{98.3} {\scriptsize[97.4, 99.1]} \\
 & gpt-6-luna & 43.2 {\scriptsize[35.1, 50.6]} & 59.0 {\scriptsize[50.4, 66.4]} & 73.6 {\scriptsize[65.5, 81.5]} & 0.68 {\scriptsize[0.21, 1.41]} & 96.3 {\scriptsize[92.0, 98.9]} \\
 & Qwen3.8-27B & 50.3 {\scriptsize[38.9, 61.0]} & 47.3 {\scriptsize[30.2, 67.0]} & 53.2 {\scriptsize[34.4, 74.9]} & 0.42 {\scriptsize[0.00, 1.07]} & 93.1 {\scriptsize[87.4, 97.3]} \\
\midrule
\multirow{2}{*}{\textbf{Talk to Your Slides}}
 & gpt-6-sol & 35.5 {\scriptsize[28.2, 43.4]} & 64.8 {\scriptsize[58.0, 72.2]} & 76.6 {\scriptsize[70.0, 83.3]} & 6.23 {\scriptsize[4.37, 8.44]} & 93.6 {\scriptsize[89.3, 97.4]} \\
 & gpt-6-luna & 29.0 {\scriptsize[20.9, 38.2]} & 41.3 {\scriptsize[33.1, 50.4]} & 49.3 {\scriptsize[40.5, 58.5]} & 5.63 {\scriptsize[3.54, 8.16]} & 95.5 {\scriptsize[92.7, 97.7]} \\
\midrule
\multirow{2}{*}{\textbf{Claude Code + PPTX skill}}
 & Sonnet 5 & 73.8 {\scriptsize[66.3, 80.7]} & 81.5 {\scriptsize[75.5, 87.9]} & \textbf{97.8} {\scriptsize[95.5, 99.6]} & 0.88 {\scriptsize[0.40, 1.38]} & 93.9 {\scriptsize[89.5, 97.4]} \\
 & Haiku 4.5 & 53.6 {\scriptsize[45.8, 61.2]} & 54.4 {\scriptsize[42.5, 65.6]} & 81.0 {\scriptsize[73.5, 88.5]} & 1.55 {\scriptsize[0.58, 2.73]} & 84.7 {\scriptsize[73.6, 94.7]} \\
\midrule
\multirow{5}{*}{\textbf{\editppt{}}}
 & gpt-6-sol & \textbf{74.9} {\scriptsize[67.4, 82.1]} & \textbf{89.0} {\scriptsize[84.2, 92.9]} & 95.7 {\scriptsize[91.8, 98.3]} & 0.42 {\scriptsize[0.12, 0.75]} & 98.1 {\scriptsize[96.7, 99.3]} \\
 & gpt-6-luna & 73.8 {\scriptsize[67.4, 80.5]} & 88.5 {\scriptsize[83.8, 92.7]} & 95.5 {\scriptsize[91.9, 98.1]} & 0.54 {\scriptsize[0.23, 0.98]} & \textbf{98.9} {\scriptsize[98.2, 99.6]} \\
 & Qwen3.8-27B & 72.1 {\scriptsize[65.8, 78.3]} & 87.4 {\scriptsize[84.1, 91.6]} & 95.8 {\scriptsize[93.8, 98.0]} & \underline{0.36} {\scriptsize[0.06, 0.73]} & 97.4 {\scriptsize[94.8, 99.6]} \\
 & Sonnet 5 & \underline{74.3} {\scriptsize[68.6, 80.3]} & \underline{88.8} {\scriptsize[85.4, 92.6]} & \underline{97.0} {\scriptsize[95.2, 98.5]} & \textbf{0.30} {\scriptsize[0.09, 0.54]} & 97.6 {\scriptsize[95.4, 99.7]} \\
 & Haiku 4.5 & 66.7 {\scriptsize[59.2, 74.2]} & 83.8 {\scriptsize[78.5, 88.4]} & 93.3 {\scriptsize[88.9, 96.9]} & 0.55 {\scriptsize[0.15, 1.02]} & 98.0 {\scriptsize[96.2, 99.7]} \\
\bottomrule
\end{tabular}}
\caption{The quality metrics of Table~\ref{tab:main} with 95\% deck-cluster bootstrap intervals (28 decks, 10{,}000 resamples). Best value per column in bold, second underlined, as in Table~\ref{tab:main}.}
\label{tab:ci-systems}
\end{table*}

\begin{table*}[t]
\centering
\scriptsize
\setlength{\tabcolsep}{3pt}
\resizebox{\textwidth}{!}{%
\begin{tabular}{ll c cc lllll}
\toprule
A & B & A/B only & $p$ & $p_{\text{Holm}}$ & $\Delta$TS & $\Delta$Slide & $\Delta$IF & $\Delta$Over-edit & $\Delta$Pres \\
\midrule
\editppt{} sol & PPTPilot sol & 55/12 & $<$0.001 & $<$0.001 & +23.5 [15.1, 31.9] & +19.9 [11.5, 27.1] & +12.4 [3.1, 20.3] & $-$0.26 [$-$0.62, 0.07] & $-$0.2 [$-$1.7, 1.2] \\
 & PPTPilot luna & 66/8 & $<$0.001 & $<$0.001 & +31.7 [23.2, 40.3] & +30.1 [21.2, 39.3] & +22.1 [12.6, 30.7] & $-$0.26 [$-$0.99, 0.25] & +1.9 [$-$1.1, 6.1] \\
 & PPTPilot Qwen & 56/11 & $<$0.001 & $<$0.001 & +24.6 [14.4, 35.6] & +41.7 [21.7, 59.3] & +42.5 [20.7, 62.1] & 0.00 [$-$0.54, 0.43] & +5.0 [0.9, 10.7] \\
 & TTYS sol & 77/5 & $<$0.001 & $<$0.001 & +39.3 [31.8, 47.0] & +24.2 [17.3, 30.9] & +19.1 [12.5, 26.2] & $-$5.81 [$-$8.07, $-$3.97] & +4.6 [0.4, 8.9] \\
 & TTYS luna & 87/3 & $<$0.001 & $<$0.001 & +45.9 [38.1, 54.0] & +47.7 [37.6, 57.0] & +46.4 [36.2, 56.3] & $-$5.21 [$-$7.81, $-$3.05] & +2.7 [$-$0.2, 5.6] \\
 & Claude Code Sonnet & 20/18 & 0.87 & 1.00 & +1.1 [$-$5.6, 8.3] & +7.5 [0.4, 14.7] & $-$2.1 [$-$6.5, 1.0] & $-$0.46 [$-$0.73, $-$0.19] & +4.3 [0.7, 8.3] \\
 & Claude Code Haiku & 49/10 & $<$0.001 & $<$0.001 & +21.3 [12.5, 30.6] & +34.6 [21.9, 46.8] & +14.7 [5.7, 22.4] & $-$1.13 [$-$2.31, $-$0.23] & +13.4 [3.1, 24.7] \\
\addlinespace[2pt]
\editppt{} luna & PPTPilot sol & 51/10 & $<$0.001 & $<$0.001 & +22.4 [14.1, 31.3] & +19.4 [11.7, 25.8] & +12.2 [3.1, 20.1] & $-$0.15 [$-$0.90, 0.61] & +0.6 [$-$0.2, 1.5] \\
 & PPTPilot luna & 67/11 & $<$0.001 & $<$0.001 & +30.6 [22.0, 39.3] & +29.5 [20.7, 39.1] & +21.9 [12.8, 30.4] & $-$0.14 [$-$0.74, 0.30] & +2.7 [0.1, 6.8] \\
 & PPTPilot Qwen & 53/10 & $<$0.001 & $<$0.001 & +23.5 [14.2, 33.5] & +41.2 [22.1, 57.9] & +42.3 [20.8, 61.6] & +0.11 [$-$0.48, 0.62] & +5.8 [1.7, 11.4] \\
 & TTYS sol & 76/6 & $<$0.001 & $<$0.001 & +38.3 [32.8, 44.0] & +23.7 [17.5, 29.7] & +18.9 [12.0, 26.1] & $-$5.70 [$-$7.82, $-$3.82] & +5.4 [1.5, 9.5] \\
 & TTYS luna & 87/5 & $<$0.001 & $<$0.001 & +44.8 [37.6, 52.2] & +47.2 [37.8, 56.2] & +46.2 [36.2, 56.0] & $-$5.09 [$-$7.57, $-$3.02] & +3.5 [1.1, 6.1] \\
 & Claude Code Sonnet & 21/21 & 1.00 & 1.00 & 0.0 [$-$7.1, 7.9] & +7.0 [1.0, 12.9] & $-$2.3 [$-$6.4, 0.8] & $-$0.34 [$-$0.94, 0.30] & +5.1 [1.7, 9.1] \\
 & Claude Code Haiku & 45/8 & $<$0.001 & $<$0.001 & +20.2 [12.6, 27.9] & +34.1 [21.6, 46.6] & +14.5 [5.6, 22.4] & $-$1.02 [$-$2.23, 0.11] & +14.2 [4.5, 25.1] \\
\addlinespace[2pt]
\editppt{} Qwen & PPTPilot sol & 54/16 & $<$0.001 & $<$0.001 & +20.8 [12.2, 29.4] & +18.3 [11.5, 24.4] & +12.6 [5.3, 19.6] & $-$0.32 [$-$1.06, 0.31] & $-$0.9 [$-$3.6, 1.3] \\
 & PPTPilot luna & 61/8 & $<$0.001 & $<$0.001 & +29.0 [21.7, 36.6] & +28.5 [21.5, 36.9] & +22.2 [15.5, 29.1] & $-$0.32 [$-$1.05, 0.21] & +1.1 [$-$1.1, 3.9] \\
 & PPTPilot Qwen & 50/10 & $<$0.001 & $<$0.001 & +21.9 [12.2, 32.2] & +40.1 [22.3, 56.2] & +42.6 [21.4, 61.1] & $-$0.07 [$-$0.74, 0.46] & +4.3 [$-$0.4, 10.2] \\
 & TTYS sol & 74/7 & $<$0.001 & $<$0.001 & +36.6 [30.4, 42.9] & +22.6 [16.3, 29.3] & +19.2 [12.3, 26.0] & $-$5.88 [$-$8.08, $-$4.01] & +3.8 [0.2, 6.9] \\
 & TTYS luna & 84/5 & $<$0.001 & $<$0.001 & +43.2 [35.0, 51.1] & +46.1 [38.3, 54.0] & +46.5 [37.3, 55.7] & $-$5.27 [$-$7.83, $-$3.17] & +1.9 [$-$1.4, 5.0] \\
 & Claude Code Sonnet & 22/25 & 0.77 & 1.00 & $-$1.6 [$-$7.9, 5.1] & +5.9 [0.3, 11.7] & $-$2.0 [$-$3.9, 0.5] & $-$0.52 [$-$1.08, $-$0.05] & +3.5 [1.0, 6.8] \\
 & Claude Code Haiku & 45/11 & $<$0.001 & $<$0.001 & +18.6 [11.4, 25.9] & +33.0 [22.7, 44.1] & +14.8 [8.1, 21.7] & $-$1.20 [$-$2.33, $-$0.31] & +12.6 [2.8, 23.6] \\
\addlinespace[2pt]
\editppt{} Sonnet & PPTPilot sol & 53/11 & $<$0.001 & $<$0.001 & +23.0 [14.6, 31.3] & +19.7 [13.2, 25.4] & +13.7 [6.0, 20.5] & $-$0.39 [$-$1.07, 0.20] & $-$0.7 [$-$2.9, 1.4] \\
 & PPTPilot luna & 70/13 & $<$0.001 & $<$0.001 & +31.1 [23.5, 39.1] & +29.8 [22.0, 39.1] & +23.4 [15.5, 31.3] & $-$0.38 [$-$1.05, 0.06] & +1.3 [$-$1.8, 5.3] \\
 & PPTPilot Qwen & 56/12 & $<$0.001 & $<$0.001 & +24.0 [13.9, 34.7] & +41.5 [22.6, 57.9] & +43.8 [22.1, 62.5] & $-$0.13 [$-$0.73, 0.30] & +4.5 [0.3, 9.9] \\
 & TTYS sol & 77/6 & $<$0.001 & $<$0.001 & +38.8 [32.8, 44.5] & +24.0 [17.2, 30.6] & +20.4 [14.0, 26.9] & $-$5.94 [$-$8.08, $-$4.13] & +4.0 [0.0, 8.0] \\
 & TTYS luna & 89/6 & $<$0.001 & $<$0.001 & +45.4 [37.6, 53.1] & +47.4 [38.4, 56.0] & +47.7 [38.3, 56.8] & $-$5.33 [$-$7.81, $-$3.31] & +2.1 [$-$0.8, 4.6] \\
 & Claude Code Sonnet & 22/21 & 1.00 & 1.00 & +0.5 [$-$6.8, 8.4] & +7.3 [2.1, 12.6] & $-$0.8 [$-$2.7, 1.2] & $-$0.58 [$-$1.11, $-$0.10] & +3.7 [0.5, 7.5] \\
 & Claude Code Haiku & 49/11 & $<$0.001 & $<$0.001 & +20.8 [12.8, 28.9] & +34.4 [22.9, 46.6] & +16.0 [8.6, 23.4] & $-$1.26 [$-$2.42, $-$0.28] & +12.8 [3.5, 23.1] \\
\addlinespace[2pt]
\editppt{} Haiku & PPTPilot sol & 49/21 & 0.001 & 0.03 & +15.3 [4.3, 25.7] & +14.7 [6.6, 21.9] & +10.1 [1.7, 16.8] & $-$0.13 [$-$0.96, 0.64] & $-$0.3 [$-$2.2, 1.6] \\
 & PPTPilot luna & 61/18 & $<$0.001 & $<$0.001 & +23.5 [14.6, 32.2] & +24.9 [16.5, 34.3] & +19.7 [10.9, 28.5] & $-$0.13 [$-$0.97, 0.52] & +1.7 [$-$1.3, 5.8] \\
 & PPTPilot Qwen & 49/19 & $<$0.001 & 0.01 & +16.4 [5.2, 28.0] & +36.5 [17.5, 53.5] & +40.1 [17.7, 60.1] & +0.13 [$-$0.57, 0.71] & +4.9 [1.2, 9.9] \\
 & TTYS sol & 68/11 & $<$0.001 & $<$0.001 & +31.1 [24.2, 38.2] & +19.0 [9.3, 27.5] & +16.7 [9.1, 24.5] & $-$5.68 [$-$7.96, $-$3.71] & +4.4 [1.0, 8.3] \\
 & TTYS luna & 78/9 & $<$0.001 & $<$0.001 & +37.7 [30.1, 45.3] & +42.5 [32.8, 51.6] & +44.0 [33.4, 54.4] & $-$5.08 [$-$7.68, $-$2.92] & +2.5 [0.5, 4.4] \\
 & Claude Code Sonnet & 18/31 & 0.09 & 1.00 & $-$7.1 [$-$16.0, 1.8] & +2.3 [$-$6.3, 10.1] & $-$4.5 [$-$9.7, $-$0.1] & $-$0.33 [$-$0.97, 0.26] & +4.1 [0.5, 8.2] \\
 & Claude Code Haiku & 43/19 & 0.003 & 0.08 & +13.1 [3.4, 22.3] & +29.4 [16.1, 42.6] & +12.3 [2.4, 21.4] & $-$1.00 [$-$2.14, $-$0.10] & +13.2 [4.1, 23.2] \\
\bottomrule
\end{tabular}}
\caption{Each \editppt{} backbone against each baseline on the 183 tasks. Differences are A $-$ B in percentage points with 95\% deck-cluster bootstrap intervals. \emph{A/B only} counts the tasks that only A or only B completes. $p$ is an exact two-sided McNemar test on Task Success, and $p_{\text{Holm}}$ applies the Holm correction over all 66 pairs in this table and Table~\ref{tab:ci-pairs-other}. Model names are shortened to sol, luna, Qwen, Sonnet, and Haiku.}
\label{tab:ci-pairs}
\end{table*}

\begin{table*}[t]
\centering
\scriptsize
\setlength{\tabcolsep}{3pt}
\resizebox{\textwidth}{!}{%
\begin{tabular}{ll c cc lllll}
\toprule
A & B & A/B only & $p$ & $p_{\text{Holm}}$ & $\Delta$TS & $\Delta$Slide & $\Delta$IF & $\Delta$Over-edit & $\Delta$Pres \\
\midrule
\editppt{} sol & \editppt{} luna & 16/14 & 0.86 & 1.00 & +1.1 [$-$5.9, 8.1] & +0.5 [$-$2.3, 2.8] & +0.2 [$-$1.2, 1.3] & $-$0.12 [$-$0.66, 0.35] & $-$0.8 [$-$2.3, 0.3] \\
 & \editppt{} Qwen & 18/13 & 0.47 & 1.00 & +2.7 [$-$3.2, 8.4] & +1.6 [$-$3.6, 5.4] & $-$0.1 [$-$4.8, 3.0] & +0.06 [$-$0.35, 0.49] & +0.8 [$-$1.4, 3.0] \\
 & \editppt{} Sonnet & 18/17 & 1.00 & 1.00 & +0.5 [$-$6.7, 7.9] & +0.2 [$-$3.8, 3.9] & $-$1.3 [$-$4.6, 1.5] & +0.12 [$-$0.22, 0.51] & +0.6 [$-$1.6, 2.4] \\
 & \editppt{} Haiku & 25/10 & 0.02 & 0.38 & +8.2 [0.5, 16.3] & +5.2 [0.3, 10.6] & +2.4 [$-$0.7, 6.5] & $-$0.13 [$-$0.69, 0.42] & +0.2 [$-$1.8, 2.0] \\
\addlinespace[2pt]
\editppt{} luna & \editppt{} Qwen & 16/13 & 0.71 & 1.00 & +1.6 [$-$3.2, 6.5] & +1.0 [$-$3.3, 4.4] & $-$0.3 [$-$4.6, 2.5] & +0.18 [$-$0.19, 0.64] & +1.5 [$-$0.2, 3.8] \\
 & \editppt{} Sonnet & 13/14 & 1.00 & 1.00 & $-$0.5 [$-$4.8, 4.4] & $-$0.3 [$-$3.2, 2.4] & $-$1.5 [$-$4.5, 1.2] & +0.24 [$-$0.05, 0.65] & +1.4 [$-$0.2, 3.0] \\
 & \editppt{} Haiku & 20/7 & 0.02 & 0.40 & +7.1 [1.1, 13.3] & +4.7 [0.3, 10.3] & +2.2 [$-$0.8, 6.3] & $-$0.01 [$-$0.46, 0.49] & +1.0 [$-$0.3, 2.3] \\
\addlinespace[2pt]
\editppt{} Qwen & \editppt{} Sonnet & 15/19 & 0.61 & 1.00 & $-$2.2 [$-$7.2, 2.9] & $-$1.3 [$-$4.0, 1.8] & $-$1.2 [$-$3.4, 1.6] & +0.06 [$-$0.17, 0.32] & $-$0.2 [$-$1.7, 0.8] \\
 & \editppt{} Haiku & 23/13 & 0.13 & 1.00 & +5.5 [0.0, 11.0] & +3.6 [$-$1.5, 10.2] & +2.5 [$-$1.9, 7.9] & $-$0.19 [$-$0.45, 0.05] & $-$0.6 [$-$2.4, 0.9] \\
\addlinespace[2pt]
\editppt{} Sonnet & \editppt{} Haiku & 23/9 & 0.02 & 0.40 & +7.7 [2.8, 12.8] & +4.9 [0.7, 10.1] & +3.7 [0.1, 7.9] & $-$0.25 [$-$0.67, 0.13] & $-$0.4 [$-$1.4, 0.3] \\
\midrule
PPTPilot sol & PPTPilot luna & 25/10 & 0.02 & 0.38 & +8.2 [2.1, 14.5] & +10.2 [0.2, 22.1] & +9.7 [0.7, 20.7] & 0.00 [$-$0.94, 0.76] & +2.0 [$-$0.7, 6.4] \\
 & PPTPilot Qwen & 22/20 & 0.88 & 1.00 & +1.1 [$-$6.5, 8.9] & +21.8 [4.7, 36.0] & +30.1 [10.2, 46.8] & +0.26 [$-$0.53, 0.94] & +5.2 [0.9, 10.9] \\
 & TTYS sol & 51/22 & $<$0.001 & 0.03 & +15.8 [7.2, 24.6] & +4.3 [$-$3.4, 12.1] & +6.7 [$-$2.0, 16.0] & $-$5.55 [$-$7.88, $-$3.74] & +4.7 [0.7, 9.2] \\
 & TTYS luna & 63/22 & $<$0.001 & $<$0.001 & +22.4 [12.5, 32.3] & +27.8 [17.7, 38.0] & +34.0 [22.9, 45.5] & $-$4.95 [$-$7.71, $-$2.67] & +2.8 [0.3, 5.7] \\
 & Claude Code Sonnet & 10/51 & $<$0.001 & $<$0.001 & $-$22.4 [$-$30.5, $-$14.7] & $-$12.4 [$-$18.4, $-$6.7] & $-$14.5 [$-$21.9, $-$6.6] & $-$0.20 [$-$0.63, 0.14] & +4.4 [0.8, 8.9] \\
 & Claude Code Haiku & 26/30 & 0.69 & 1.00 & $-$2.2 [$-$9.8, 5.3] & +14.7 [2.1, 29.0] & +2.3 [$-$7.8, 12.9] & $-$0.87 [$-$2.20, 0.17] & +13.5 [3.8, 24.6] \\
\addlinespace[2pt]
PPTPilot luna & PPTPilot Qwen & 20/33 & 0.10 & 1.00 & $-$7.1 [$-$16.7, 2.9] & +11.6 [$-$7.8, 30.1] & +20.4 [$-$2.2, 40.6] & +0.26 [$-$0.25, 0.89] & +3.2 [$-$2.7, 9.7] \\
 & TTYS sol & 45/31 & 0.14 & 1.00 & +7.7 [$-$2.2, 16.9] & $-$5.9 [$-$17.3, 4.7] & $-$3.0 [$-$15.3, 8.5] & $-$5.55 [$-$7.63, $-$3.68] & +2.7 [$-$0.6, 5.4] \\
 & TTYS luna & 53/27 & 0.005 & 0.12 & +14.2 [3.0, 24.8] & +17.6 [4.8, 29.2] & +24.3 [10.4, 38.0] & $-$4.95 [$-$7.39, $-$2.81] & +0.8 [$-$3.9, 5.0] \\
 & Claude Code Sonnet & 10/66 & $<$0.001 & $<$0.001 & $-$30.6 [$-$38.9, $-$22.2] & $-$22.6 [$-$33.5, $-$12.6] & $-$24.2 [$-$31.0, $-$17.2] & $-$0.20 [$-$0.80, 0.57] & +2.4 [$-$0.8, 6.0] \\
 & Claude Code Haiku & 23/42 & 0.02 & 0.47 & $-$10.4 [$-$18.2, $-$2.8] & +4.5 [$-$3.0, 10.9] & $-$7.4 [$-$16.1, 0.2] & $-$0.87 [$-$2.04, 0.34] & +11.5 [0.1, 23.7] \\
\addlinespace[2pt]
PPTPilot Qwen & TTYS sol & 48/21 & 0.002 & 0.04 & +14.8 [5.1, 23.6] & $-$17.5 [$-$31.3, $-$0.9] & $-$23.4 [$-$39.9, $-$2.6] & $-$5.81 [$-$8.08, $-$3.77] & $-$0.5 [$-$6.1, 5.2] \\
 & TTYS luna & 53/14 & $<$0.001 & $<$0.001 & +21.3 [11.2, 30.6] & +6.0 [$-$7.4, 21.4] & +3.9 [$-$11.3, 22.8] & $-$5.21 [$-$7.83, $-$2.94] & $-$2.4 [$-$6.6, 0.7] \\
 & Claude Code Sonnet & 9/52 & $<$0.001 & $<$0.001 & $-$23.5 [$-$33.5, $-$14.1] & $-$34.2 [$-$47.9, $-$17.5] & $-$44.6 [$-$63.9, $-$22.7] & $-$0.46 [$-$0.94, 0.07] & $-$0.7 [$-$8.1, 5.6] \\
 & Claude Code Haiku & 25/31 & 0.50 & 1.00 & $-$3.3 [$-$12.0, 5.1] & $-$7.1 [$-$25.5, 11.9] & $-$27.8 [$-$46.3, $-$7.4] & $-$1.13 [$-$2.34, 0.02] & +8.4 [$-$0.5, 17.0] \\
\addlinespace[2pt]
TTYS sol & TTYS luna & 22/10 & 0.05 & 0.90 & +6.6 [2.0, 11.9] & +23.5 [14.5, 32.2] & +27.3 [17.8, 36.6] & +0.60 [$-$1.67, 2.68] & $-$1.9 [$-$5.9, 1.1] \\
 & Claude Code Sonnet & 8/78 & $<$0.001 & $<$0.001 & $-$38.3 [$-$45.2, $-$30.8] & $-$16.7 [$-$21.6, $-$12.3] & $-$21.2 [$-$28.5, $-$13.9] & +5.36 [3.53, 7.61] & $-$0.3 [$-$4.4, 5.0] \\
 & Claude Code Haiku & 17/50 & $<$0.001 & 0.002 & $-$18.0 [$-$26.5, $-$9.0] & +10.4 [$-$2.0, 23.2] & $-$4.4 [$-$12.8, 4.7] & +4.68 [2.63, 7.16] & +8.8 [$-$2.4, 20.3] \\
\addlinespace[2pt]
TTYS luna & Claude Code Sonnet & 7/89 & $<$0.001 & $<$0.001 & $-$44.8 [$-$52.5, $-$36.9] & $-$40.2 [$-$48.9, $-$31.8] & $-$48.5 [$-$58.1, $-$38.5] & +4.75 [2.51, 7.44] & +1.6 [$-$3.2, 6.7] \\
 & Claude Code Haiku & 16/61 & $<$0.001 & $<$0.001 & $-$24.6 [$-$33.1, $-$15.2] & $-$13.1 [$-$25.6, 0.3] & $-$31.7 [$-$43.3, $-$20.2] & +4.08 [2.15, 6.41] & +10.7 [1.8, 20.4] \\
\addlinespace[2pt]
Claude Code Sonnet & Claude Code Haiku & 47/10 & $<$0.001 & $<$0.001 & +20.2 [12.6, 27.5] & +27.1 [16.1, 39.8] & +16.8 [10.3, 22.8] & $-$0.68 [$-$1.91, 0.31] & +9.1 [$-$2.2, 21.6] \\
\bottomrule
\end{tabular}}
\caption{Pairwise comparisons among the \editppt{} backbones (top) and among the baselines (bottom), in the format of Table~\ref{tab:ci-pairs}.}
\label{tab:ci-pairs-other}
\end{table*}

\paragraph{Ablations.}
Table~\ref{tab:ci-ablation} gives the intervals behind the ablation results in Section~\ref{sec:res-ablation}, as differences between each variant and the full system on the same tasks with gpt-6-luna. Removing the dispatcher and specialists reliably lowers Task Success and Slide Success and raises Over-edit, while Preservation does not change. Disabling the validation loop reliably lowers only Preservation, both over all tasks and on the 57 tasks in which the loop triggered a retry.

\begin{table*}[t]
\centering
\small
\setlength{\tabcolsep}{3pt}
\resizebox{\textwidth}{!}{%
\begin{tabular}{l c lllll cc}
\toprule
Variant & Tasks & $\Delta$TS & $\Delta$Slide & $\Delta$IF & $\Delta$Over-edit & $\Delta$Pres & Full / variant only & $p$ \\
\midrule
Single agent & 183 & $-$12.0 [$-$17.4, $-$6.9] & $-$6.0 [$-$10.0, $-$2.7] & $-$2.2 [$-$5.3, 0.4] & +5.05 [2.17, 8.19] & $-$0.2 [$-$0.9, 0.5] & 27/5 & $<$0.001 \\
No validation & 183 & $-$3.3 [$-$6.7, 0.0] & $-$2.2 [$-$5.5, 0.6] & $-$0.6 [$-$2.3, 0.6] & $-$0.06 [$-$0.41, 0.15] & $-$2.1 [$-$5.0, $-$0.1] & 9/3 & 0.15 \\
No validation, retried tasks & 57 & $-$10.5 [$-$21.4, 0.0] & $-$4.9 [$-$11.8, 1.4] & $-$1.5 [$-$4.7, 1.5] & $-$0.14 [$-$0.92, 0.37] & $-$4.7 [$-$10.4, $-$0.1] & 9/3 & 0.15 \\
\bottomrule
\end{tabular}}
\caption{Ablations of \editppt{} with gpt-6-luna: each variant minus the full system, in points, with 95\% deck-cluster bootstrap intervals. \emph{Full / variant only} counts the tasks completed only by the full system or only by the variant, and $p$ is an exact two-sided McNemar test on Task Success. The 183-task no-validation row reuses the results of the 126 tasks in which the loop never triggered a retry.}
\label{tab:ci-ablation}
\end{table*}

\FloatBarrier
\section{Additional Results}
\label{app:additional}

\subsection{Task Success by Editing Scope}
\label{app:scope}
Table~\ref{tab:scope} groups the tasks by their number of target slides instead of deck length. Averaged over the twelve systems, Task Success falls from 66.6\% on single-slide tasks to 41.0\% on tasks with 20 or more target slides. The two groupings overlap, since short decks hold no task with 20 or more target slides, but the decline also appears within medium and long decks. PPTPilot is the exception, because it returns no usable deck for many single-slide tasks with the gpt-6 backbones and for half of the widest tasks with Qwen3.8-27B. With the gpt-6, Qwen3.8-27B, and Sonnet~5 backbones, our framework still passes 86.9--89.7\% of the target slides in the widest group, so its lower Task Success there is consistent with the stricter task-level criterion, which requires every target slide of a task to succeed.

\begin{table}[t]
\centering
\small
\setlength{\tabcolsep}{4pt}
\begin{tabular}{ll cccc}
\toprule
System & Backbone & 1 & 2--5 & 6--19 & $\ge$20 \\
\midrule
\multirow{3}{*}{PPTPilot}
 & gpt-6-sol   & 45.1 & 53.8 & 69.2 & 44.4 \\
 & gpt-6-luna  & 37.8 & 38.5 & 53.8 & 47.2 \\
 & Qwen3.8-27B & 50.0 & 57.7 & 61.5 & 33.3 \\
\midrule
\multirow{2}{*}{TTYS}
 & gpt-6-sol   & 56.1 & 34.6 & 12.8 & 13.9 \\
 & gpt-6-luna  & 52.4 & 15.4 & 12.8 & 2.8 \\
\midrule
\multirow{2}{*}{Claude Code}
 & Sonnet 5    & 80.5 & \underline{73.1} & \textbf{82.1} & 50.0 \\
 & Haiku 4.5   & 59.8 & 57.7 & 51.3 & 38.9 \\
\midrule
\multirow{5}{*}{\editppt{}}
 & gpt-6-sol   & \textbf{86.6} & \underline{73.1} & 69.2 & \underline{55.6} \\
 & gpt-6-luna  & \underline{84.1} & \textbf{84.6} & 64.1 & 52.8 \\
 & Qwen3.8-27B & \underline{84.1} & \underline{73.1} & 64.1 & 52.8 \\
 & Sonnet 5    & 79.3 & \textbf{84.6} & \underline{71.8} & \textbf{58.3} \\
 & Haiku 4.5   & 82.9 & 65.4 & 56.4 & 41.7 \\
\bottomrule
\end{tabular}
\caption{Task Success by the number of target slides per task, with 82 tasks on one slide, 26 on 2--5, 39 on 6--19, and 36 on 20 or more. Best per column in bold, second underlined.}
\label{tab:scope}
\end{table}

\subsection{Cost and Runtime Distributions}
\label{app:cost}
Table~\ref{tab:cost-dist} gives the distribution of per-task cost and wall-clock time behind the means in Table~\ref{tab:main}, over every task a system ran. Both are right-skewed for every system, with the mean above the median. For \editppt{}, time grows with deck length (with gpt-6-sol, a median of 1.2, 2.7, and 4.6 minutes on short, medium, and long decks), and the slowest tenth of tasks accounts for 52--58\% of the total time with every backbone. The slowest task is the same for every backbone, translating all text of a 38-slide deck into Korean, which takes from 64.5 minutes with gpt-6-luna to 190.1 minutes with Haiku~4.5. A typical task therefore costs and takes much less than the mean: with gpt-6-sol, the median task costs \$0.142 and takes 2.4 minutes, against means of \$0.429 and 6.1 minutes.

\begin{table*}[t]
\centering
\small
\setlength{\tabcolsep}{4pt}
\begin{tabular}{ll rrrr rrrr}
\toprule
 & & \multicolumn{4}{c}{Cost per task (USD)} & \multicolumn{4}{c}{Time per task (min)} \\
\cmidrule(lr){3-6} \cmidrule(lr){7-10}
System & Model & Mean & SD & Median & Max & Mean & SD & Median & Max \\
\midrule
\multirow{3}{*}{\textbf{PPTPilot}}
 & gpt-6-sol & \$0.298 & 0.341 & 0.154 & 1.705 & 1.7 & 1.5 & 1.1 & 8.2 \\
 & gpt-6-luna & \$0.017 & 0.017 & 0.009 & 0.091 & 1.6 & 1.3 & 1.1 & 7.7 \\
 & Qwen3.8-27B & -- & -- & -- & -- & 10.2 & 9.6 & 6.9 & 56.6 \\
\midrule
\multirow{2}{*}{\textbf{Talk to Your Slides}}
 & gpt-6-sol & \$0.625 & 0.909 & 0.195 & 4.363 & 7.4 & 10.2 & 2.4 & 44.4 \\
 & gpt-6-luna & \$0.027 & 0.035 & 0.010 & 0.188 & 6.7 & 8.3 & 2.8 & 33.5 \\
\midrule
\multirow{2}{*}{\textbf{Claude Code + PPTX skill}}
 & Sonnet 5 & \$0.387 & 0.458 & 0.271 & 3.152 & 1.6 & 3.1 & 0.8 & 31.8 \\
 & Haiku 4.5 & \$0.182 & 0.140 & 0.130 & 0.935 & 1.6 & 1.5 & 1.1 & 11.5 \\
\midrule
\multirow{5}{*}{\textbf{\editppt{}}}
 & gpt-6-sol & \$0.429 & 0.777 & 0.142 & 6.085 & 6.1 & 11.0 & 2.4 & 89.3 \\
 & gpt-6-luna & \$0.036 & 0.142 & 0.007 & 1.691 & 5.1 & 8.9 & 2.0 & 64.5 \\
 & Qwen3.8-27B & -- & -- & -- & -- & 7.3 & 13.3 & 3.0 & 113.0 \\
 & Sonnet 5 & \$0.919 & 1.622 & 0.313 & 11.993 & 4.9 & 9.2 & 1.8 & 75.0 \\
 & Haiku 4.5 & \$1.031 & 2.042 & 0.336 & 13.541 & 12.2 & 26.4 & 4.0 & 190.1 \\
\bottomrule
\end{tabular}
\caption{Per-task cost at list prices and wall-clock time over every task each system ran (183 per system; 181 with a recorded time for PPTPilot with Qwen3.8-27B). SD is the sample standard deviation. Qwen runs on local GPUs and has no list price (--).}
\label{tab:cost-dist}
\end{table*}

\subsection{Validation Loop Analysis}
\label{app:validation}
Table~\ref{tab:validator} compares the text validator's verdicts with the evaluator on every slide the agents edited in the \editppt{} run with gpt-6-luna. A slide accepted at the first attempt keeps that attempt, so its grade shows how often an accepted edit is correct: 91.7\% of these slides pass, and most of the remaining 8.3\% are partial or failed edits (111 of 127) rather than damage or edits to the wrong slide. When the validator rejects an attempt and a retry is then accepted, 76.2\% of the slides end as a pass. Slides on which every attempt is rejected end worse, with 41.9\% passing when the best attempt is kept and 38.5\% when the slide is restored.

\begin{table}[t]
\centering
\small
\setlength{\tabcolsep}{3pt}
\resizebox{\columnwidth}{!}{%
\begin{tabular}{l rrrrrr r}
\toprule
Validator path & Slides & Pass & Partial & Fail & Damage & Wrong & Pass rate \\
\midrule
Accepted at first attempt & 1525 & 1398 & 56 & 55 & 9 & 7 & \textbf{91.7\%} \\
Rejected, then accepted & 101 & 77 & 18 & 4 & 2 & 0 & \underline{76.2\%} \\
Rejected, best attempt kept & 74 & 31 & 30 & 6 & 6 & 1 & 41.9\% \\
Rejected, slide restored & 13 & 5 & 0 & 7 & 1 & 0 & 38.5\% \\
\bottomrule
\end{tabular}}
\caption{Text validator verdicts against the evaluator for \editppt{} with gpt-6-luna, over the slides the agents edited. \emph{Damage}: the requested change was graded but a non-target shape on the slide also changed. \emph{Wrong}: a non-target slide. Slides that are neither target nor wrong-target slides are excluded. Best pass rate in bold, second underlined.}
\label{tab:validator}
\end{table}

\subsection{TTYS Failure Analysis}
\label{app:ttys}
To help interpret the TTYS results, we describe the failure patterns observed in its runs, as read from its code, the result decks, and the evaluator's slide labels. Most of its failed tasks leave some target slides unedited rather than editing the right slides incorrectly. Of the 1{,}720 target slides across all tasks, TTYS misses 261 with gpt-6-sol and 815 with gpt-6-luna.

Three behaviors account for most of these misses. First, TTYS processes one slide at a time, and when the model's output for a slide cannot be parsed, the run stops with an error before all planned edits are applied. This happens in 51 tasks with gpt-6-luna and 9 with gpt-6-sol, and these tasks hold most of the slides that gpt-6-luna misses. Second, its slide parser reports only the name and type of the first three shapes in a group and none of their text, so the model does not see text inside groups. Of the 186 target shapes that sit in groups, TTYS leaves 180 unchanged with gpt-6-sol, whereas our framework with the same backbone changes 179. Third, a task can run to completion without an error and still leave its targets unchanged. With gpt-6-sol, more than half of the missed slides belong to such tasks.

Edits to non-target slides occur only in tasks that span multiple slides. The planner receives only the file name and the number of slides and is instructed to write one task per slide, so it often plans an edit for every slide. On slides that needed no change, the generated code still rewrites text formatting, and on some of them it replaces automatic slide numbers with fixed text.

\FloatBarrier
\section{Case Studies}
\label{app:qualitative}
This section examines individual tasks, including several on which \editppt{} fails.

\begin{figure*}[p]
\centering
\begin{subfigure}{\textwidth}
\centering
\includegraphics[width=\textwidth]{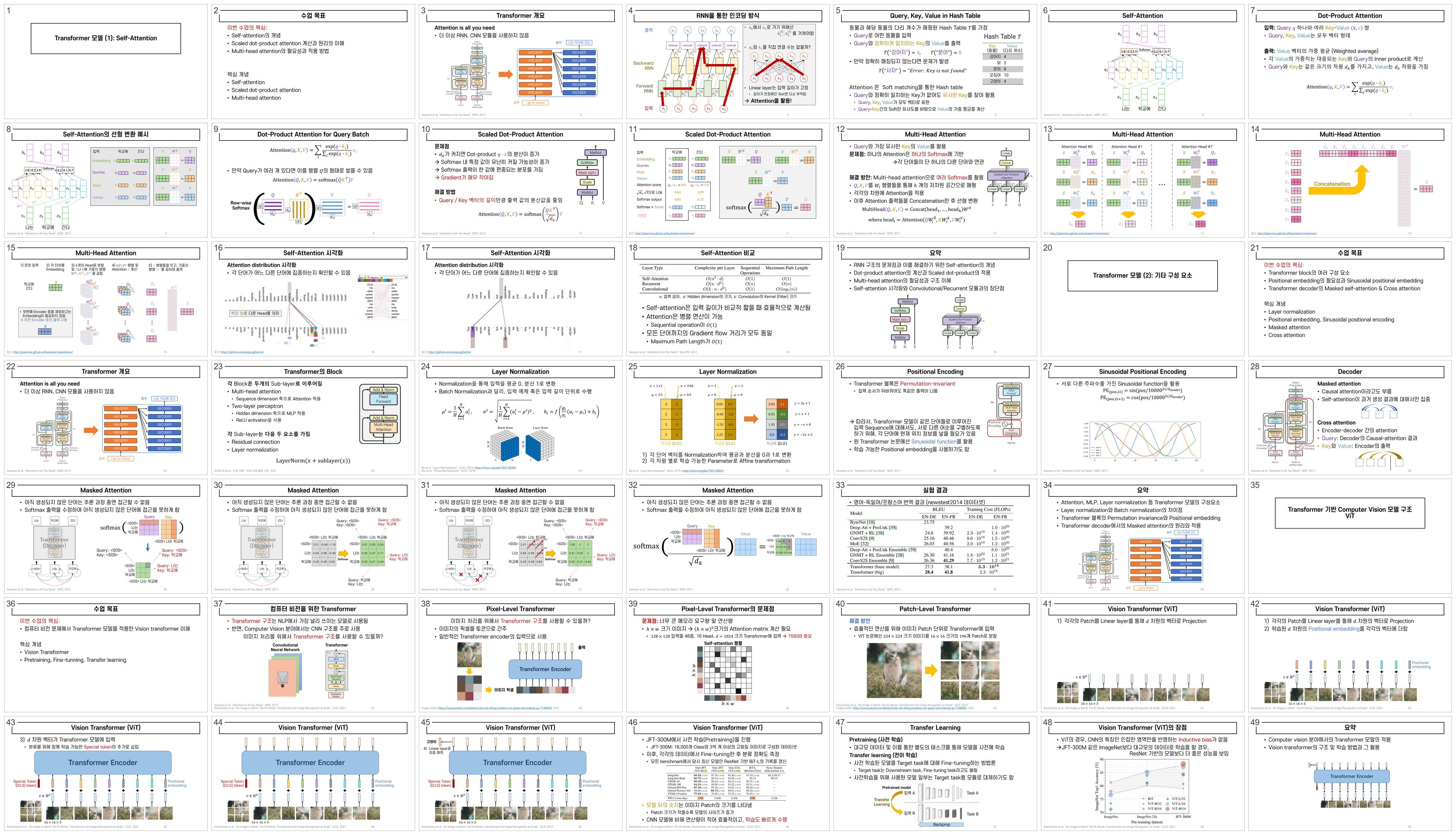}
\caption{Original deck.}
\end{subfigure}

\vspace{0.6em}
\begin{subfigure}{\textwidth}
\centering
\includegraphics[width=\textwidth]{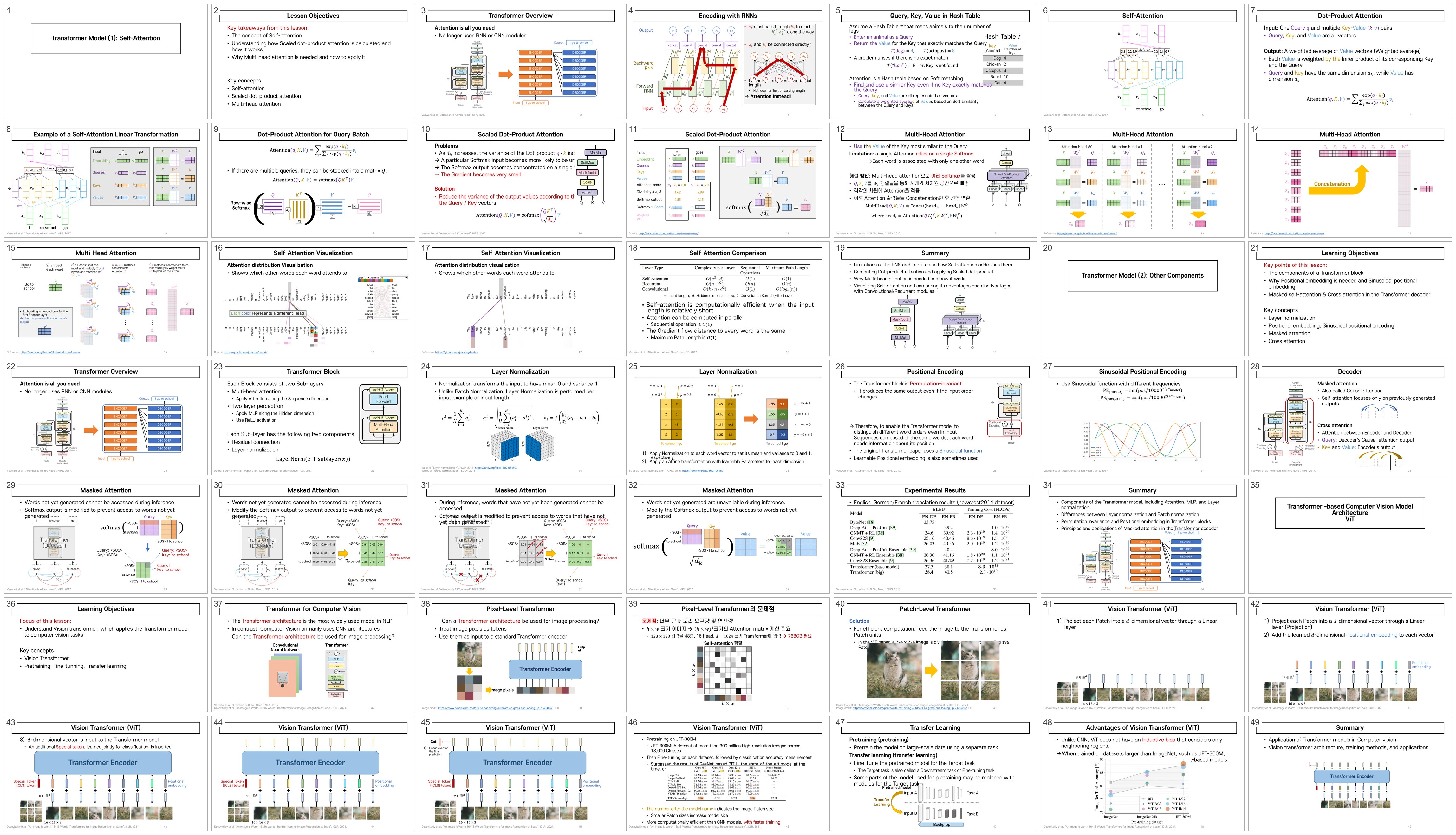}
\caption{After \editppt{} with gpt-6-sol.}
\end{subfigure}
\caption{All 49 slides of a deck before and after the request ``Translate every Korean passage in the deck into English.'' Slides are numbered in order, and slide 44, which has no Korean text, is not a target.}
\label{fig:case-transformer}
\end{figure*}

\paragraph{Translating a lecture deck.}
\citet{jung2026talkslideshighefficiencyslide} translate a 49-slide Korean lecture deck on the Transformer into English in their appendix. They note that the model occasionally turns mathematical formulas into illegible text and that the longer English text can overflow its box. Figure~\ref{fig:case-transformer} shows a deck of this kind before and after the same translation with \editppt{} (gpt-6-sol). Most of the text is translated, and almost all equations stay intact, with Korean words inside them translated in place. A few passages remain in Korean or lose the formatting of individual words, so the task does not count as completed under our criterion. The edit also introduces new rendering defects, for example on slide 40, where the translated text is longer than the original and runs into the picture below it. Defects of this kind are not visible to object-level checks (Appendix~\ref{app:vision}).

\begin{figure}[t]
\centering
\begin{subfigure}{\columnwidth}
\centering
\includegraphics[width=\columnwidth]{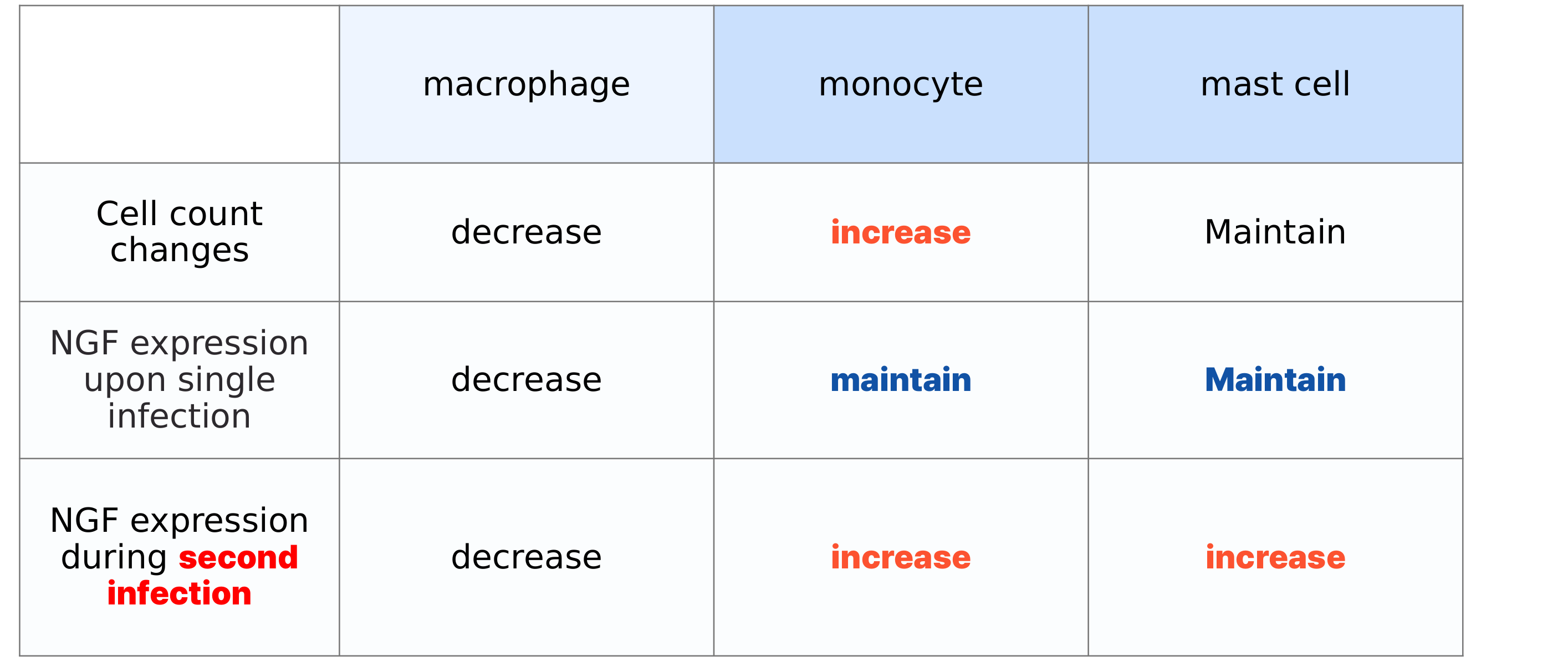}
\caption{Original table.}
\end{subfigure}

\vspace{0.4em}
\begin{subfigure}{\columnwidth}
\centering
\includegraphics[width=\columnwidth]{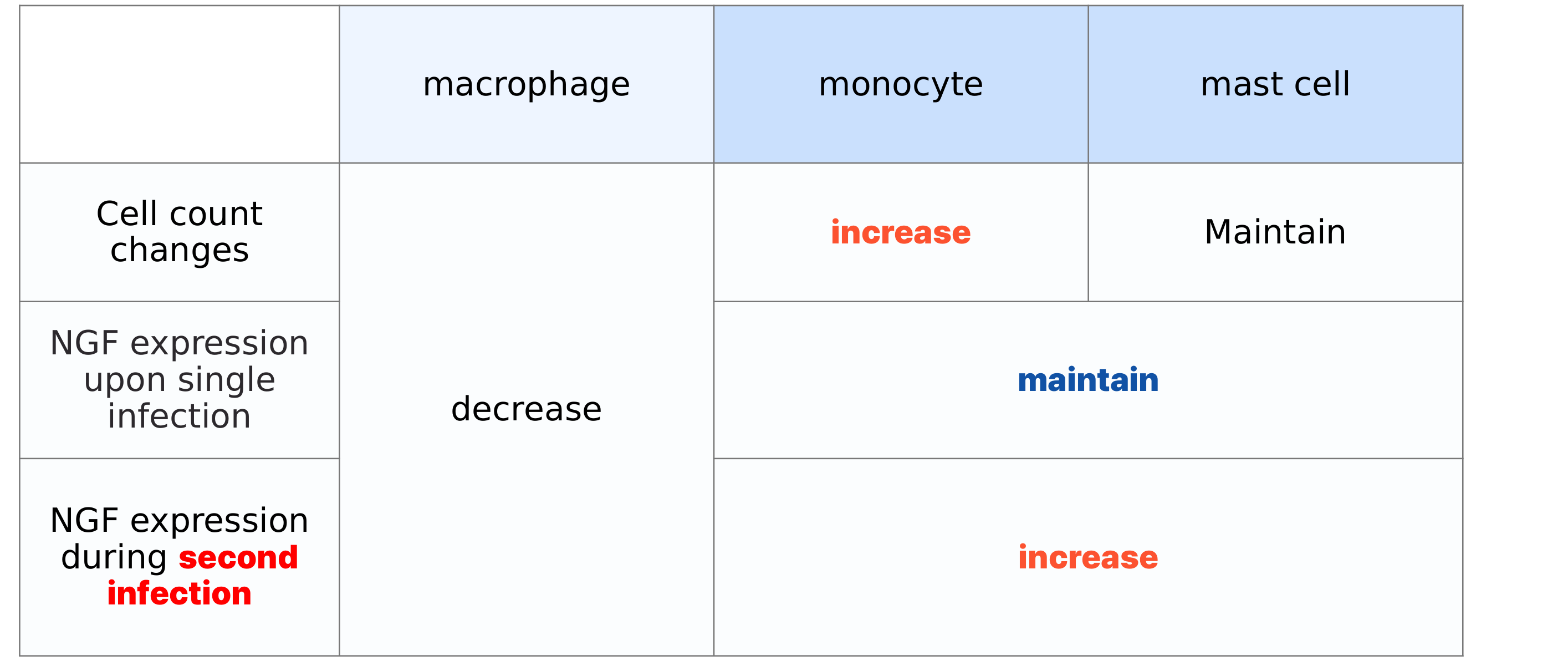}
\caption{\editppt{} with Qwen3.8-27B (completed).}
\end{subfigure}

\vspace{0.4em}
\begin{subfigure}{\columnwidth}
\centering
\includegraphics[width=\columnwidth]{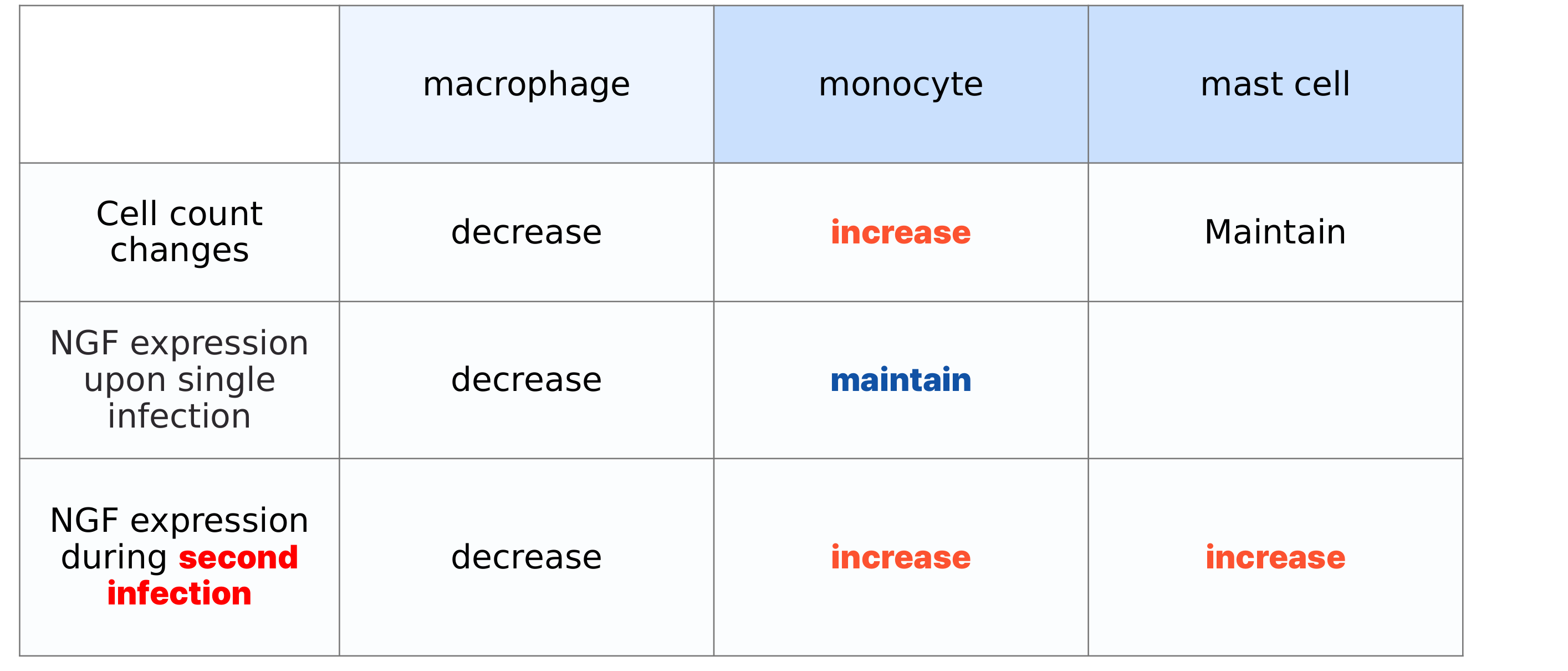}
\caption{\editppt{} with gpt-6-sol (not completed).}
\end{subfigure}
\caption{A table before and after the request ``On slide 15's table, merge adjacent cells that contain the same text, ignoring letter case. If a cell could be merged either horizontally or vertically, merge it horizontally.''}
\label{fig:case-table}
\end{figure}

\paragraph{Merging table cells.}
Figure~\ref{fig:case-table} shows a request to merge adjacent cells of a table that contain the same text, ignoring letter case, and to merge horizontally when a cell could be merged in either direction. The agent sees the table as a list of cells with their row and column indices and text. It has to work out from these indices which cells are vertical or horizontal neighbors, apply the stated preference, and express each merge as a tool call on the right pair of cells. The preference matters in the last column, where the lower of the two \emph{Maintain} cells could join the cell above it or the \emph{maintain} cell to its left, and the request decides for the left one. With Qwen3.8-27B, \editppt{} merges the three \emph{decrease} cells vertically and the \emph{maintain} and \emph{increase} pairs horizontally, which completes the request. With gpt-6-sol, the agent first clears the text of the lower \emph{Maintain} cell instead of merging it. Its later calls name the same three merges as the Qwen run but are malformed, so no merge takes effect, and the run ends with the cleared cell.

\begin{figure}[t]
\centering
\begin{subfigure}{\columnwidth}
\centering
\includegraphics[width=\columnwidth]{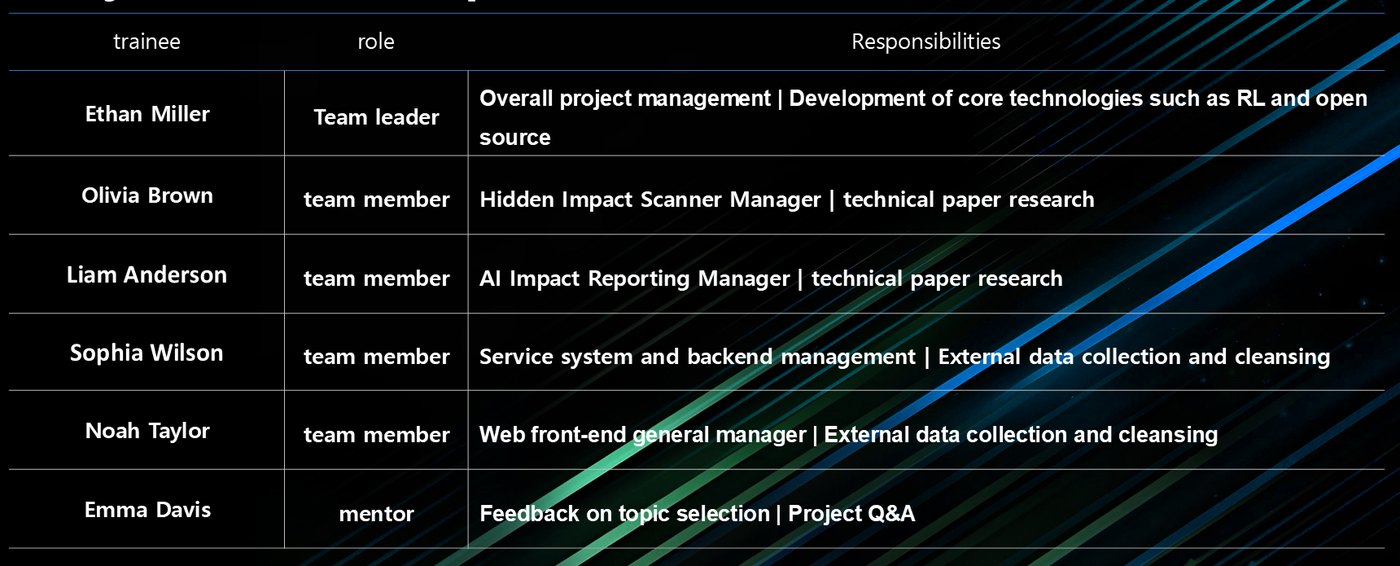}
\caption{Original table.}
\end{subfigure}

\vspace{0.4em}
\begin{subfigure}{\columnwidth}
\centering
\includegraphics[width=\columnwidth]{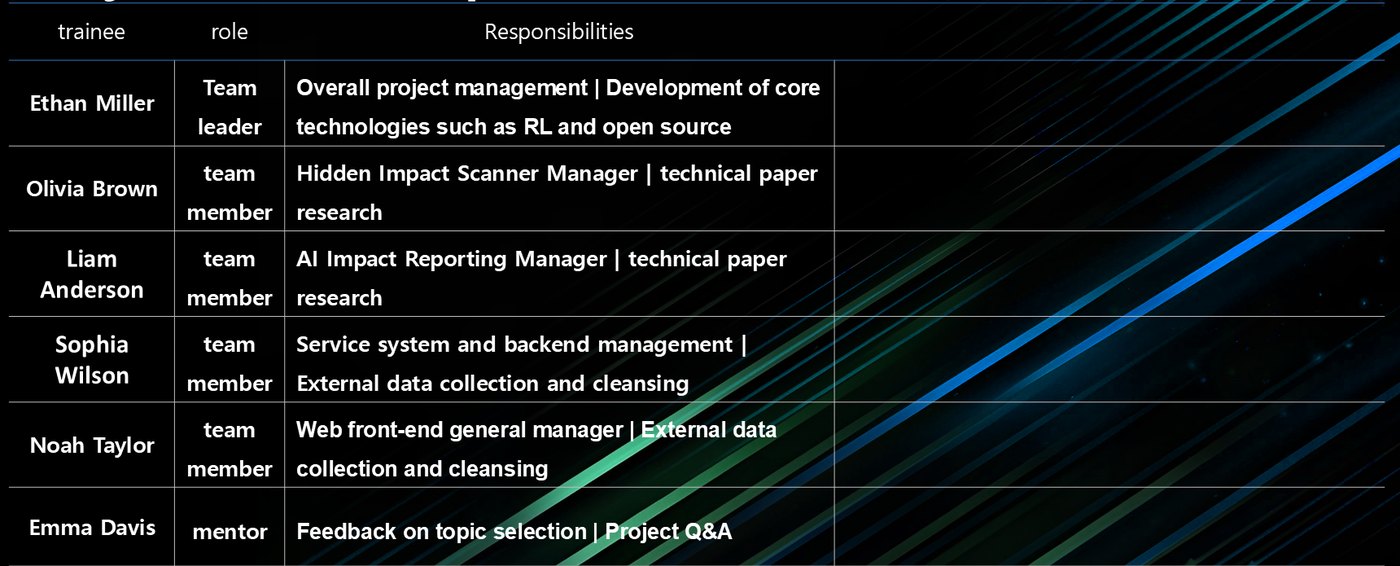}
\caption{\editppt{} with gpt-6-sol (not completed).}
\end{subfigure}

\vspace{0.4em}
\begin{subfigure}{\columnwidth}
\centering
\includegraphics[width=\columnwidth]{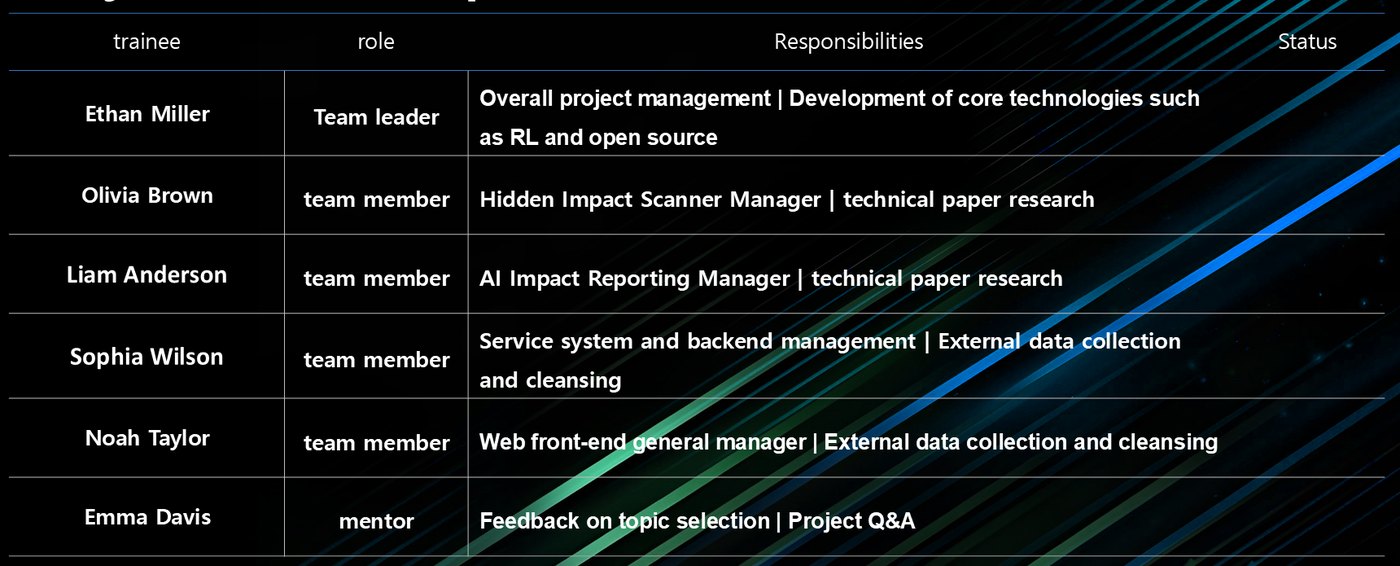}
\caption{Claude Code with Sonnet~5 (completed).}
\end{subfigure}
\caption{A table before and after the request ``On slide 29, add a new column at the right end of the table with the header `Status'.'' Names in the table are placeholders from the source deck.}
\label{fig:case-column}
\end{figure}

\paragraph{Adding a table column.}
Figure~\ref{fig:case-column} shows a task that \editppt{} did not complete with any backbone: adding a column with the header \emph{Status} at the right end of a table. The edit takes two tool calls, one that adds the column and one that writes the header cell. With gpt-6-sol, the agent added the column but left its header empty. The validator found the empty cell and asked for the tool that writes cell text, but after each rollback the agent issued the same call to add the column, and when the retries ran out the deck kept a column without a header. Adding the column also narrowed the existing columns, so their text now wraps onto more lines. The runs with gpt-6-luna and Qwen3.8-27B failed in the same way. With Sonnet~5 and Haiku~4.5, the agent wrote the header, but adding the column had also moved the table slightly, which counts as an unrequested change. Claude Code with Sonnet~5 and PPTPilot with Qwen3.8-27B completed the task. The case shows a limit of the recovery loop: a correct diagnosis does not help when the agent keeps choosing the same tool.

\paragraph{Different transitions on consecutive slides.}
A request asks for a transition on every slide of a 30-slide deck such that no two consecutive slides share the same one. The planner split it into one task per slide and told each task to pick an effect that differs from those of its neighbors. Each specialist sees only its own slide, however, so it cannot know which effects its neighbors received. With gpt-6-sol, the agent gave the first slide \emph{random bars} and the other 29 slides \emph{checkerboard}, and with gpt-6-luna it gave all 30 slides \emph{fade}. The runs with Haiku~4.5 and Sonnet~5 also repeated effects on some consecutive slides. With Qwen3.8-27B, the planner instead fixed the effect of every slide in advance, alternating \emph{fade} and \emph{push} between odd and even slides, and every slide passed. A constraint that links slides is thus kept only when the planner turns it into a concrete value for each slide. Claude Code with both backbones and PPTPilot with gpt-6-sol and gpt-6-luna completed the task.

\begin{figure}[t]
\centering
\begin{subfigure}{\columnwidth}
\centering
\includegraphics[width=\columnwidth]{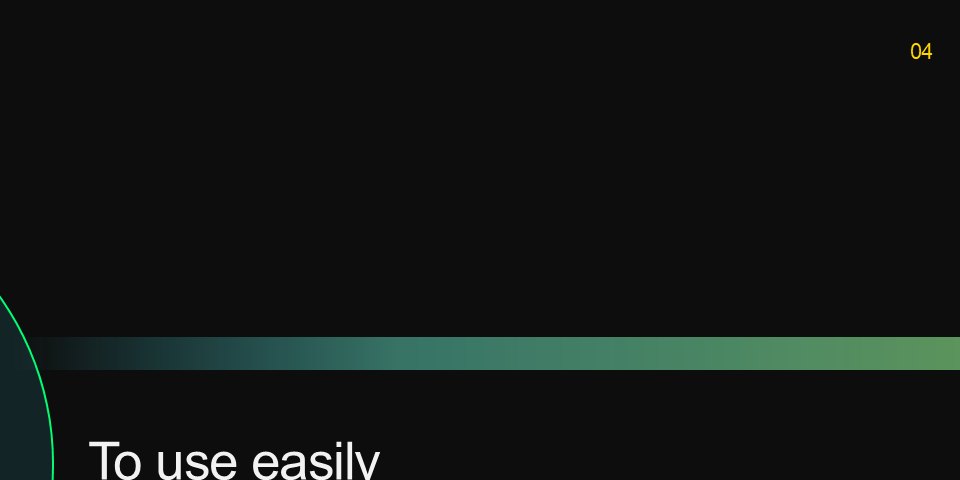}
\caption{Slide 6: the badge \emph{04} at the top right was recolored.}
\end{subfigure}

\vspace{0.4em}
\begin{subfigure}{\columnwidth}
\centering
\includegraphics[width=\columnwidth]{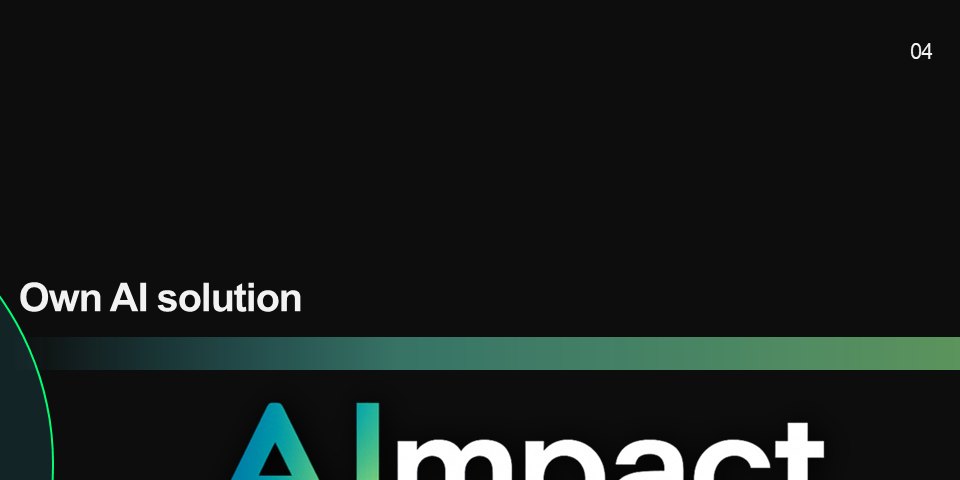}
\caption{Slide 7: the badge of the same section was left unchanged.}
\end{subfigure}
\caption{Two slides after \editppt{} with gpt-6-sol on the request ``On every slide that has a small section-number badge, change the badge color to gold.'' Only the upper right part of each slide is shown.}
\label{fig:case-badge}
\end{figure}

\paragraph{Recoloring section badges.}
A request asks to color the small section-number badge gold on every slide that has one, which is 18 slides of the deck (Figure~\ref{fig:case-badge}). With gpt-6-sol, the dispatcher selected the badge on every one of these slides, and the agent recolored it on 14. On the other four, the specialist returned no tool call. The framework treats an empty answer as a sign that the slide already meets the request and skips validation, so these slides were left unchanged and nothing flagged them. The other four \editppt{} backbones completed the task.

\FloatBarrier
\section{Vision Validation}
\label{app:vision}
Object-level checks compare the parsed shapes before and after an edit, so they cannot see defects that appear only when a slide is rendered, such as text that overflows its box, shapes that cover each other, content pushed off the slide, or text that loses contrast against its background. This section reports how often editing introduced such defects in our runs (\ref{app:vision-audit}), describes the data and training of an open vision validator for them (\ref{app:vision-data}--\ref{app:vision-extra}), and lists the validator's repairs on audited slides (\ref{app:vision-repairs}). For an edited slide, the vision validator judges whether the edit introduced a critical rendering defect, names the defect and the shapes it affects, and proposes a repair as tool calls. The vision validator is disabled in all main benchmark runs and is evaluated separately in this appendix. We will release it with its weights.

We use four defect categories that can be verified in the rendered image: content that lies more than one point outside the slide, content hidden behind another shape, overlaps that impair reading, and text whose contrast with the surface behind it is too low to read. Broken line breaks and paragraph structure are left out, because they are failures of text structure that the pixels alone cannot verify.

\subsection{Defect Audit}
\label{app:vision-audit}
\begin{figure*}[t]
\centering
\includegraphics[width=0.75\textwidth]{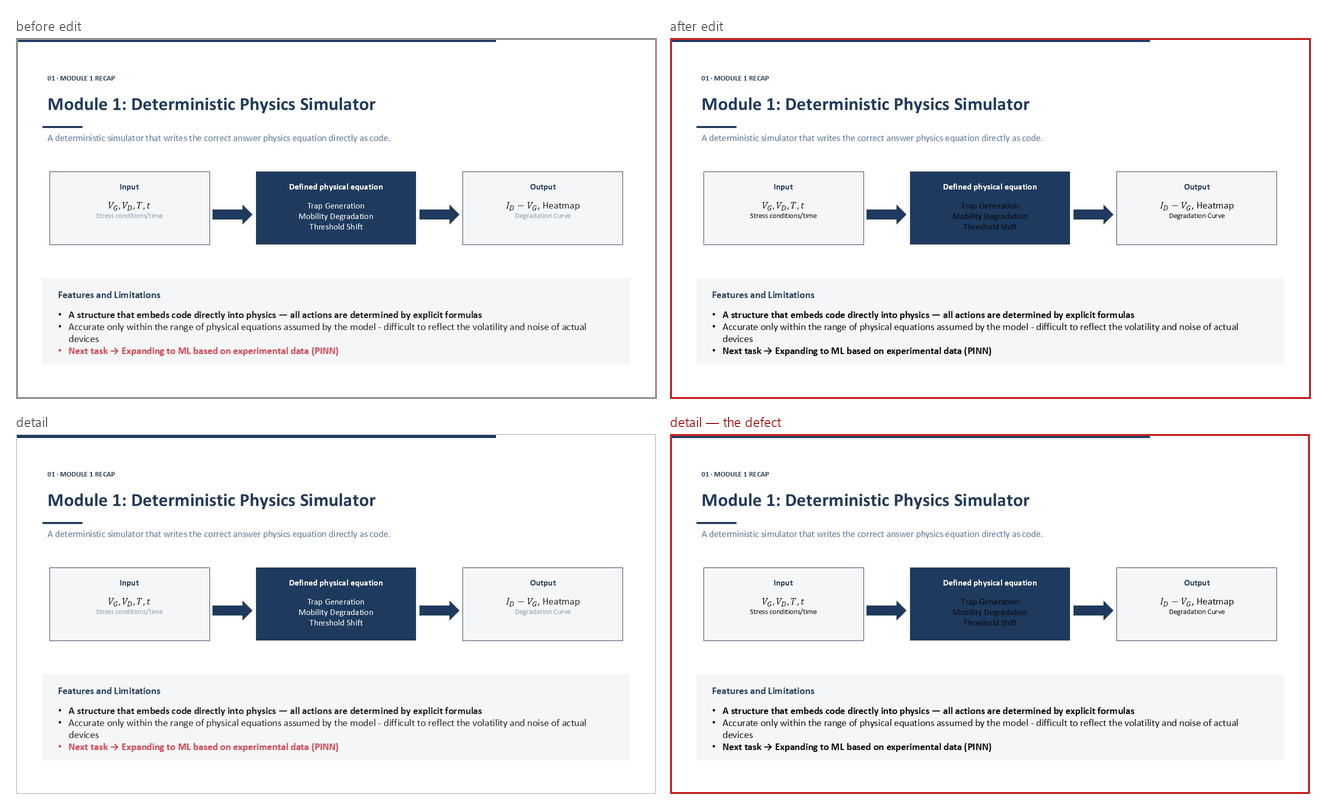}
\caption{A low-contrast defect found in the audit, shown before and after the edit with the changed region magnified below. After \editppt{} with gpt-6-luna changed the body font of every slide to Calibri, the white text in the dark box turned black, and the annotator noted that it is hard to see on the dark background.}
\label{fig:vision-audit-examples}
\end{figure*}

Rendering defects introduced by editing are rare but not negligible. We rendered the 3{,}615 slides edited in two runs with gpt-6-luna and gpt-6-sol, both with vision validation disabled, and kept the 3{,}149 slides whose rendering changed. One author first labeled every slide while viewing the before and after renders side by side, with the changed shapes outlined and the changed pixels marked. A slide was labeled defective when the edit left a serious visual defect, and marginal when it could still be read without trouble but a viewer would likely want it improved. An undergraduate annotator then went through all of the author's labels and pointed out those they disagreed with, and the two discussed each disagreement until they agreed, which yielded a single adjudicated label set. The annotator took part voluntarily and gave verbal consent. The adjudicated set marks 121 slides as defective and 167 as marginal. We set the marginal slides aside, because whether they need a repair depends largely on taste, and use only the defective ones in what follows. Five of the defects were already present in the source deck, which leaves 116 slides, 3.7\% of the inspected slides, with a newly introduced defect. Notably, of the tasks that these two runs succeed on under Task Success, 129 of 135 with gpt-6-luna and 131 of 137 with gpt-6-sol contain no slide labeled defective. This supports the view that success under our object-level evaluator largely carries over to the rendered slides, even though the evaluator does not inspect them. Figure~\ref{fig:vision-audit-examples} shows one of the defects.

A survey then checked whether these labels hold without the annotation aids. Raters saw the same before and after renders that a model receives, with no outlines or pixel measurements, and judged each slide as defective, unsure, or clean. The instructions (Figure~\ref{fig:survey-instructions}) asked whether the edit left a problem that actually hinders reading or viewing rather than a matter of design taste, listed typical cases with examples, said that some slides were intact, and asked the raters to spend 20 to 30 seconds on each slide. The list of cases is broader than the four categories above, since it also names broken line breaks and misaligned layouts. An answer of unsure does not count as confirming a defect. Three raters took part, none of them an author. They were recruited through personal communication from among research colleagues and acquaintances, participated voluntarily without compensation, and gave verbal consent after being informed of the evaluation task. They did not know the details of the experiment. In a first condition, one rater judged all 116 slides in order to narrow them to the defects a reader notices without help, and marked 52 as defective. In a second condition, the two other raters worked separately on different days. Each judged 88 of the 116 slides mixed with 30 slides that the adjudicated set marks clean, without knowing which were which. They confirmed 50 and 66 of the 88 defects and flagged only one and two of the 30 clean slides, with a Cohen's $\kappa$ of 0.51 between them. At least one of the two confirmed 72 slides, which is 2.3\% of the inspected slides, and both confirmed 44. To tell submissions apart, the survey recorded only a self-chosen nickname, a hashed network address, and the browser type. With these we kept one independent submission per rater, so a later submission from the device of one rater was not used.

\begin{figure*}[t]
\centering
\fbox{\begin{minipage}{0.96\textwidth}\small
\textbf{Slide defect judgment.} Look at the slide before the edit and the slide after the edit side by side, and judge whether the edit produced a serious visual defect. Judge whether it actually causes a problem for reading or viewing, not design taste.

\smallskip
\textbf{Count as defects:}
text overflows its box or is cut off so that it cannot be read;
text or shapes overlap and hide each other;
content goes beyond the edge of the slide;
the text color is so close to the background that it is hard to read;
broken line breaks or paragraphs collapse the structure of a list or a sentence;
the alignment is noticeably off so that the layout looks broken;
small elements such as page numbers, footers, and captions are hidden or overlapped (they are small and easy to miss).

\smallskip
\textbf{A defect:} the title was pushed to two lines and covers the shape below it; a translated sentence is cut off at the bottom of its box and the last line is not visible.
\textbf{Not a defect:} the text color changed but it reads well; a box became slightly larger but does not overlap anything; the margins changed slightly but reading is not affected.

\smallskip
If the slide before the edit has the same problem, still mark it as a defect if you see the problem after the edit. You do not need to decide whether this edit caused it.
Some slides are intact, so not every slide in the set is defective.
If you cannot decide, press ``Unsure''. You do not have to force a decision. When you choose Unsure or Defect, a one-line note on what bothered you helps a lot.
Please look at each item for 20 to 30 seconds, because defects in small elements are missed when you skim. You do not have to finish. Every answer is saved as soon as you click, so you can close the window, and if you come back with the same name, you continue where you left off. The name only tells responses apart, and any nickname is fine.
\end{minipage}}
\caption{Instructions shown to the survey raters, translated from Korean.}
\label{fig:survey-instructions}
\end{figure*}

Independent raters thus confirmed a large part of the annotated defects, and they rarely flagged clean slides. Each rater confirmed between about half and three quarters of the defects, while gpt-6-luna flagged 69 of the same 88. Many of the defects that a rater did not confirm involve a page number, a footer, or a URL, while few of the confirmed ones do, so small elements account for much of the disagreement.

\subsection{Training Data and Evaluation Protocol}
\label{app:vision-data}
The vision validator is trained on defects synthesized on 2{,}251 public decks released under CC BY or CC0, so no benchmark slide is used for training. Of these decks, 1{,}159 come from Zenodo10K~\citep{zheng-etal-2025-pptagent}, and the others are presentations we collected from Zenodo and figshare. Generators edit a deck through PowerPoint COM, and six of them produce 92\% of the samples by replacing text with a longer string, adding a text box, appending text, adding a summary box, enlarging a font, or adding a page number. A candidate defect is kept only if the rendered slide changes in the way its category predicts. Pairs that pass this check and the automatic filters are reviewed by a person as described in Appendix~\ref{app:vision-annotation}, and only confirmed pairs are used. In the training data, content hidden behind another shape is labeled as an overlap, so the validator reports the other three categories. The resulting data set has 8{,}384 samples from 2{,}251 source decks, of which 4{,}709 are defective. We split it by deck into 7{,}634 samples for training and 750 held-out samples, and use 374 of the training samples for validation.

A sample shows the slide before and after an edit, together with the tool calls of the edit and of any repair. If defective slides always came without repair calls and clean slides always with them, a model could answer from the call list without looking at the image. We therefore include all four combinations of slide state and label: edited slides with and without a defect, repaired slides that are still flawed, and repaired slides on which the defect is gone. All images are rendered at 960 by 540 pixels, where one point equals one pixel on a 16:9 slide, so the geometric amounts in the labels match the image exactly. The training set contains only 16:9 slides.

\subsection{Annotation of Training Pairs}
\label{app:vision-annotation}
\begin{figure*}[t]
\centering
\includegraphics[width=\textwidth]{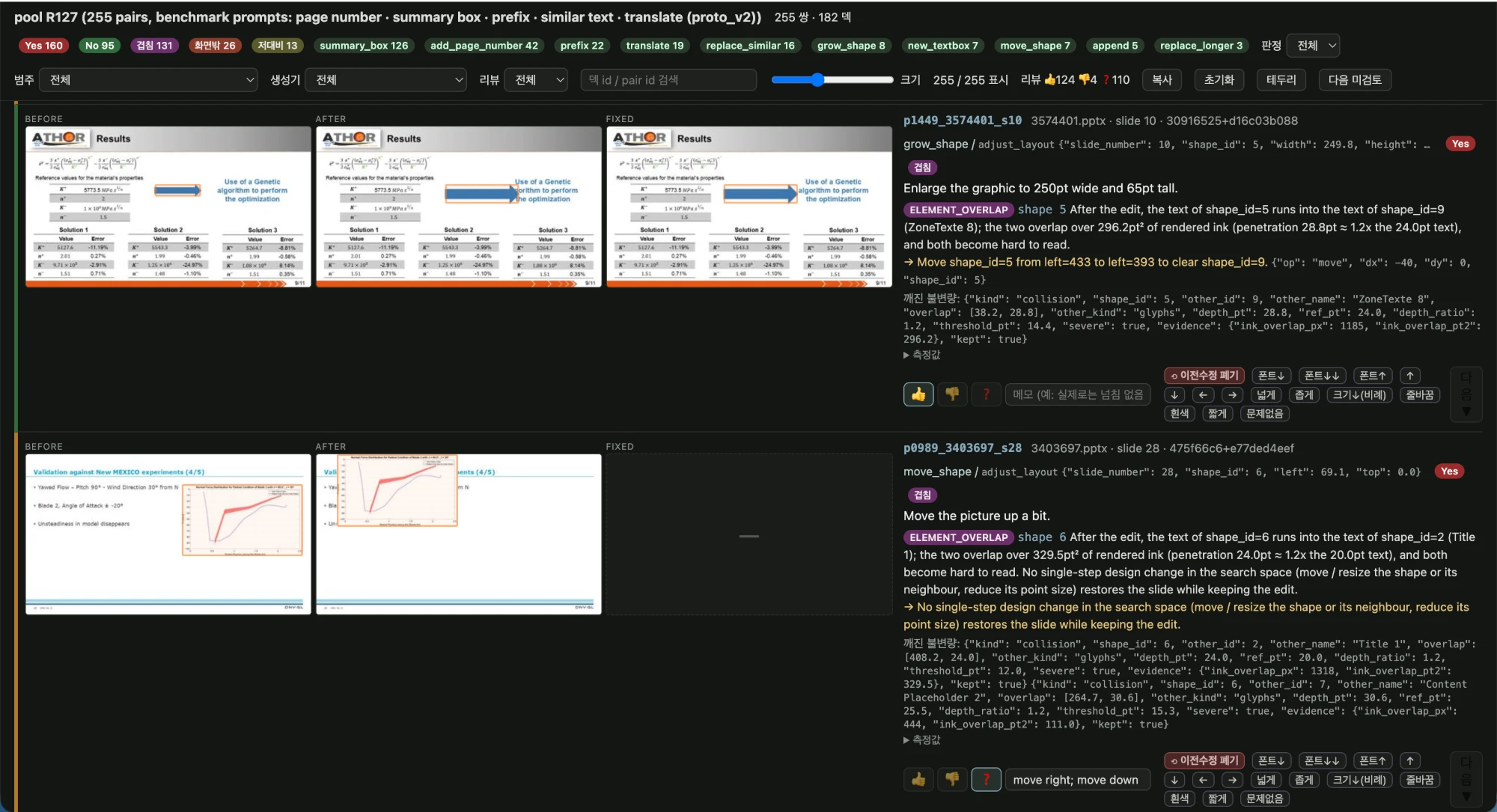}
\caption{Review page for training pairs. Each row shows the slide before an edit, after it, and after the proposed repair, with the request, the defect category, the diagnosis, and the proposed tool calls. The reviewer accepts, rejects, or questions a pair with the buttons at the bottom and enters corrections with the buttons on the right. In the second row no single change repairs the slide, so the repair column is empty and the reviewer has asked for a move to the right and down. The interface labels are in Korean. Slides from ``Creep Characterization of Refractory Materials at High Temperatures Using the Integrated Digital Image Correlation Technique'' by L.~Breder Teixeira, J.~Gillibert, E.~Blond, and T.~Sayet and ``Development of a second order dynamic stall model'' by N.C.~Adema, M.~Kloosterman, and J.G.~Schepers, both CC BY 4.0, edited.}
\label{fig:vision-review}
\end{figure*}
Every training pair was confirmed by a person, and the reviewer's corrections also shaped the repairs that the validator learns. An automatic pipeline drafts each pair. A generator applies one edit to a corpus slide through PowerPoint COM, a probe measures on the rendered slide whether the edit broke the layout and writes a diagnosis with its measurements, and a search over single changes, such as moving or resizing a shape or reducing its font size, proposes the smallest repair that the probe accepts on a new render. One author then judged the drafts on review pages like the one in Figure~\ref{fig:vision-review}, where each row shows the slide before the edit, after the edit, and after the proposed repair, next to the request, the defect categories, the diagnosis, and the proposed tool calls.

The reviewer accepts a pair, rejects it, or marks it as uncertain, and may add a correction. An accepted pair is confirmed, and a rejected pair without a correction is discarded. A correction or an uncertain mark sends the pair to another round, in which the pipeline replays the edit on a fresh copy of the deck, applies the corrected repair together with the repairs of earlier rounds, renders the result again, and shows the pair to the reviewer once more. Buttons on the page enter common corrections, such as a smaller font, a move in one direction, or a wider or narrower box, and a note that the slide has no defect turns the pair into a clean example. Later rounds also showed the repair proposed by a model trained on earlier confirmed pairs, and the reviewer chose between that repair and the pipeline's or rejected both.

The drafts were reviewed in pages of 250 pairs. Over all rounds the reviewer left 6{,}370 recorded judgments on 5{,}755 pairs. A pair that the reviewer viewed and left unmarked was discarded, and such views were not recorded, so the number of pairs the reviewer looked at is larger than the number judged.

The reviews also corrected the pipeline. Text that spills out of a frame without a visible border was rarely judged a defect, so the probe counts such a spill only when the frame is visible, and content hidden behind another shape is labeled as an overlap. Repairs for low contrast try a new text color before any move, because moving a shape cannot restore contrast. Once the reviewer had confirmed nearly every clean label of the probe, later pages concentrated on pairs on which the probe and a trained model disagreed, on generators not yet reviewed, and on defects without a repair, with a random sample of clean pairs kept as a check.

Pairs that needed several rounds also provide defective samples. A pair is sent back when a confirmed defect is still visible after a repair, because the repair went too far or not far enough. For example, the reviewer may ask to move a shape to the right to clear an overlap, and if the next render moves it too far, the reviewer asks for a move back to the left. The slide after such a repair is itself flawed, even though the pair is eventually resolved. Each round keeps its render and the repair calls it applied, so these states can be recovered from the review log. We selected such intermediate states as defective samples whose target is the repair that finally worked, and they make up 1{,}606 of the 8{,}384 samples in the data set described in Appendix~\ref{app:vision-data}.

\subsection{Model Training}
\label{app:vision-extra}
We fine-tune Qwen3.5-9B~\citep{qwen2026qwen35} with LoRA on the training split. The adapters have rank 16, a scaling factor of 32, and dropout 0.05. They are attached to the attention and feed-forward layers of the language model, to its linear-attention layers, and to the vision encoder, which gives 48.1M trainable parameters, 0.51\% of the model. We train with AdamW at a learning rate of $10^{-4}$ without weight decay, with 50 warm-up steps followed by cosine decay, 8 samples per step, and gradient clipping at 1.0. Training runs for at most three epochs with early stopping, and we select checkpoints by validation accuracy rather than validation loss. The selected checkpoint came at step 1{,}000, about 1.1 epochs, and training stopped at step 1{,}600. Each training, evaluation, and inference job runs on a single NVIDIA B200 GPU, and we count GPU hours as the time a GPU was held. Training the selected model takes about 2.3 GPU hours, and training it together with two alternative configurations takes 5.5 GPU hours. Including earlier training rounds, evaluation, and inference, the work on the vision validator used about 15.3 GPU hours. On the 497 held-out samples that show an edit without a repair, the fine-tuned model reaches an accuracy of 0.932 on the yes-or-no judgment, against about 0.60 for the base model with thinking disabled and an example of the output format in the prompt. We report accuracy on these samples because samples with a repair list the repair calls in the prompt, which can make them easier to classify.

\subsection{Repairs on Audited Slides}
\label{app:vision-repairs}
We also ran the vision validator on audited slides: the 121 slides that the audit of Appendix~\ref{app:vision-audit} marks as defective and 121 clean slides from the same decks. When it reported a defect, its tool calls were applied to the edited deck through PowerPoint and the slide was rendered again. We ran gpt-6-luna in the same role and collected the slides on which both models proposed a repair on one review page. From this page, one author picked 40 slides on which the result of the validator's repair can be seen clearly and judged the outcome, 36 from the gpt-6-luna run and 4 from the gpt-6-sol run. These cases illustrate the repairs and are not a random sample, so we list all of them instead of reporting rates. Table~\ref{tab:vision-repairs} gives the defect and the repair for each slide, and Figures~\ref{fig:vision-repairs-1}--\ref{fig:vision-repairs-3} show each slide before and after the repair.

Of the 37 slides with a defect, the repair removed it on 20 and reduced it on 6. On the other 11 the repair did not help, and on one of them it made the slide worse. The other three slides are clean slides that the validator flagged by mistake, and its repair left one of them unchanged and damaged the other two. Every repair either reduces a font size (31 cases) or moves or resizes a shape (9 cases). Moving an added page number away from the body text and shrinking text that outgrew its box account for the resolved cases. Several unresolved cases need a different kind of change, such as text hidden behind a picture or black text on a dark box, which a color change would fix. On one slide the validator asked for a smaller font on a picture, which has no text, so the call failed.

\newcommand{\repaircase}[3]{\begin{minipage}[t]{0.47\textwidth}\centering\includegraphics[width=\linewidth]{figures/vision_repairs/repair_#1.jpg}\\[1pt]{\scriptsize (#2) #3}\end{minipage}}

\begin{table*}[p]
\centering\scriptsize
\setlength{\tabcolsep}{4pt}\renewcommand{\arraystretch}{1.08}
\begin{tabular}{@{}r p{0.245\textwidth} p{0.245\textwidth} p{0.3\textwidth} p{0.11\textwidth}@{}}
\toprule
\# & Task, slide (run) & Defect noted in the audit & Repair by the validator & Outcome \\
\midrule
1 & 01\_OverseasNews-04, s.\,10 (luna) & Added page number overlaps the body text. & \texttt{adjust\_layout}: shape 2, left 886, top 525 & Resolved \\
2 & 01\_OverseasNews-04, s.\,17 (luna) & Added page number overlaps the body text. & \texttt{adjust\_layout}: shape 6, left 878, top 495 & Resolved \\
3 & 01\_OverseasNews-04, s.\,27 (luna) & Added page number overlaps the body text. & \texttt{adjust\_layout}: shape 3, left 894, top 523 & Resolved \\
4 & 01\_Transformer-01, s.\,24 (luna) & A graphic overlaps the text. & \texttt{set\_shape\_font\_size}: shape 5, 20\,pt & Resolved \\
5 & 01\_Transformer-01, s.\,29 (luna) & A graphic overlaps the text. & \texttt{set\_shape\_font\_size}: shape 5, 20\,pt & Resolved \\
6 & 01\_Transformer-01, s.\,30 (luna) & A graphic overlaps the text. & \texttt{set\_shape\_font\_size}: shape 5, 20\,pt & Resolved \\
7 & 01\_Transformer-01, s.\,31 (luna) & A graphic overlaps the text. & \texttt{set\_shape\_font\_size}: shape 5, 20\,pt & Resolved \\
8 & 01\_Transformer-01, s.\,38 (luna) & An arrow graphic overlaps the text. & \texttt{set\_shape\_font\_size}: shape 36, 16\,pt & Resolved \\
9 & 01\_Transformer-01, s.\,39 (luna) & A table overlaps the text. & \texttt{set\_shape\_font\_size}: shape 5, 20\,pt & Resolved \\
10 & 01\_Transformer-01, s.\,40 (luna) & A graphic overlaps the text. & \texttt{set\_shape\_font\_size}: shape 5, 20\,pt & Resolved \\
11 & 01\_Transformer-01, s.\,47 (luna) & A graphic overlaps the text. & \texttt{set\_shape\_font\_size}: shape 5, 20\,pt & Resolved \\
12 & 02\_SpeechAI-01, s.\,5 (luna) & A graphic overlaps the text. & \texttt{set\_shape\_font\_size}: shape 246, 12\,pt & Resolved \\
13 & 02\_UTI-06, s.\,11 (luna) & Text exceeds its box. & \texttt{set\_shape\_font\_size}: shape 14, 38\,pt & Resolved \\
14 & 02\_UTI-06, s.\,22 (luna) & Summary text slightly exceeds its box. & \texttt{set\_shape\_font\_size}: shape 12, 39\,pt & Resolved \\
15 & 02\_UTI-06, s.\,24 (luna) & Summary text slightly exceeds its box. & \texttt{set\_shape\_font\_size}: shape 4, 38\,pt & Resolved \\
16 & 02\_UTI-06, s.\,35 (luna) & Summary text slightly exceeds its box. & \texttt{set\_shape\_font\_size}: shape 7, 42\,pt & Resolved \\
17 & 06\_Algorithm-01, s.\,30 (luna) & The title overlaps the subtitle. & \texttt{set\_shape\_font\_size}: shape 2, 45\,pt & Resolved \\
18 & 11\_FourthIndustryStartup-01, s.\,1 (luna) & A text box at the bottom extends off the slide. & \texttt{adjust\_layout}: shape 13, left 676.5, top 474.1 & Resolved \\
19 & 01\_OverseasNews-04, s.\,12 (sol) & Added page number overlaps the body text. & \texttt{adjust\_layout}: shape 3, left 890, top 503 & Resolved \\
20 & 02\_SpeechAI-01, s.\,5 (sol) & A graphic overlaps the text. & \texttt{set\_shape\_font\_size}: shape 246, 12\,pt & Resolved \\
21 & 01\_Transformer-01, s.\,4 (luna) & A graph overlaps the body text. & \texttt{set\_shape\_font\_size}: shape 196, 18\,pt & Partly resolved \\
22 & 02\_SpeechAI-01, s.\,9 (luna) & A graphic overlaps the text. & \texttt{set\_shape\_font\_size}: shape 246, 12\,pt & Partly resolved \\
23 & 02\_UTI-01, s.\,29 (luna) & A text box extends off the slide. & \texttt{set\_shape\_font\_size}: shape 8, 48\,pt & Partly resolved \\
24 & 02\_UTI-06, s.\,16 (luna) & Body text slightly exceeds its box. & \texttt{set\_shape\_font\_size}: shape 6, 30\,pt & Partly resolved \\
25 & 03\_AbandonedSpace-01, s.\,19 (luna) & Two text boxes overlap. & \texttt{set\_shape\_font\_size}: shape 9, 16\,pt & Partly resolved \\
26 & 06\_SAMAudio-01, s.\,4 (luna) & A text box at the top overlaps a black box. & \texttt{set\_shape\_font\_size}: shape 540, 14\,pt & Partly resolved \\
27 & 02\_SpeechAI-08, s.\,22 (luna) & A graphic overlaps the table. & \texttt{adjust\_layout}: shape 4, top 198.7 & No change \\
28 & 02\_UTI-06, s.\,36 (luna) & Two passages of text overlap. & \texttt{set\_shape\_font\_size}: shape 7, 37\,pt & No change \\
29 & 07\_HousingComplex-01, s.\,20 (luna) & Texts are hidden behind their pictures. & \texttt{set\_shape\_font\_size}: shape 6, 10\,pt & No change \\
30 & 07\_HousingComplex-01, s.\,21 (luna) & Texts are hidden behind their pictures. & \texttt{set\_shape\_font\_size}: shape 2, 12.5\,pt & No change \\
31 & 07\_PINN\_LSTM-07, s.\,3 (luna) & Black body text is hard to read on a dark box. & \texttt{set\_shape\_font\_size}: shape 213, 11\,pt & No change \\
32 & 10\_MOSFET\_HCI-01, s.\,3 (luna) & A picture covers other shapes. & \texttt{set\_shape\_font\_size}: shape 13, 13\,pt (failed: the shape has no text) & No change \\
33 & 11\_FourthIndustryStartup-01, s.\,14 (luna) & Two text boxes overlap. & \texttt{set\_shape\_font\_size}: shape 24, 15\,pt & No change \\
34 & 11\_FourthIndustryStartup-01, s.\,3 (luna) & Two text boxes at the side edges extend off the slide. & \texttt{set\_shape\_font\_size}: shape 51, 20\,pt & No change \\
35 & 11\_FourthIndustryStartup-02, s.\,1 (luna) & A text box at the bottom extends off the slide. & \texttt{adjust\_layout}: shape 13, left 706.5, top 474.1 & No change \\
36 & 01\_OverseasNews-04, s.\,15 (sol) & Added page number overlaps a photo; white text would suit the background. & \texttt{adjust\_layout}: shape 7, left 883, top 493 & No change \\
37 & 01\_OverseasNews-07, s.\,16 (sol) & A table overlaps the first line of the text. & \texttt{set\_shape\_font\_size}: shape 5, 16\,pt & Worse \\
38 & 06\_Algorithm-01, s.\,17 (luna) & (clean) & \texttt{adjust\_layout}: shape 12, left 133.4 & Clean, unchanged \\
39 & 06\_Algorithm-01, s.\,2 (luna) & (clean) & \texttt{set\_shape\_font\_size}: shape 2, 34\,pt & Clean, damaged \\
40 & 11\_FourthIndustryStartup-01, s.\,13 (luna) & (clean) & \texttt{set\_shape\_font\_size}: shape 24, 24\,pt & Clean, damaged \\
\bottomrule
\end{tabular}
\caption{The 40 judged repairs of the vision validator, grouped by outcome. Notes are the auditor's, shortened and translated where needed. Each repair is one tool call on one shape, identified by its ID on the slide. \texttt{set\_shape\_font\_size} sets the font size, and \texttt{adjust\_layout} sets the position in points from the top-left corner of the slide. Each repair is shown in Figures~\ref{fig:vision-repairs-1}--\ref{fig:vision-repairs-3}.}
\label{tab:vision-repairs}
\end{table*}

\begin{figure*}[p]
\centering
\textbf{\small Resolved}\par\smallskip
\repaircase{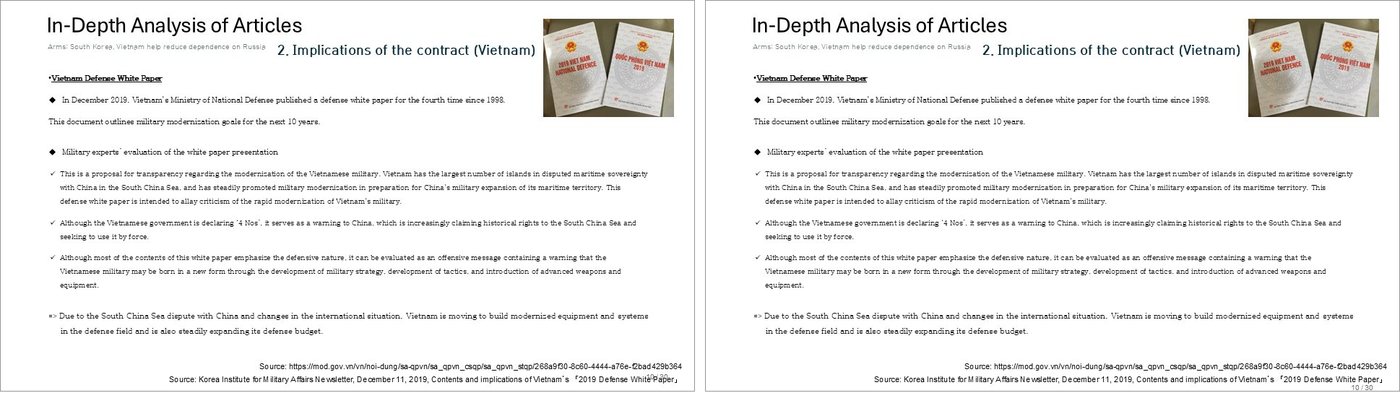}{1}{01\_OverseasNews-04, s.\,10 (luna)}\hfill\repaircase{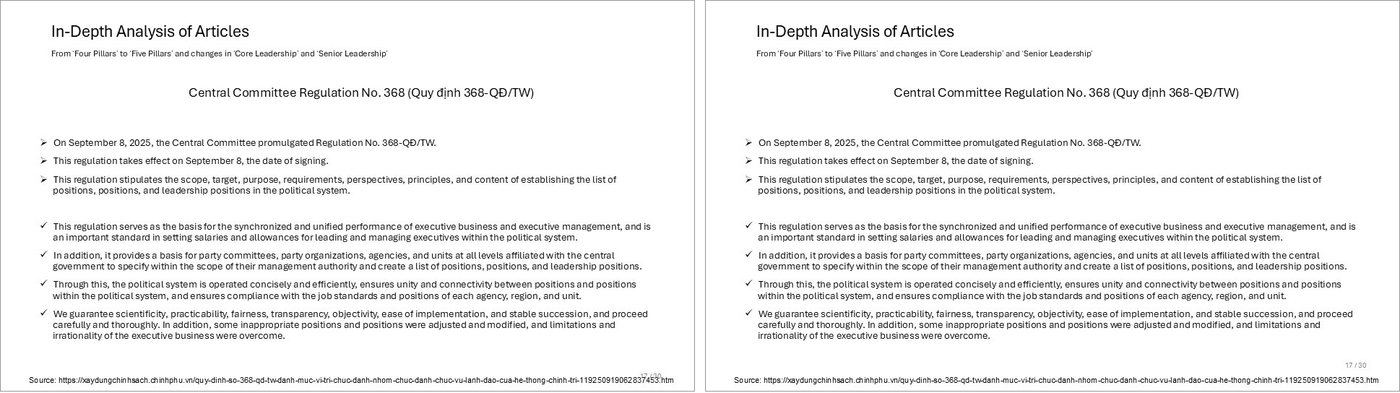}{2}{01\_OverseasNews-04, s.\,17 (luna)}\par\vspace{2pt}
\repaircase{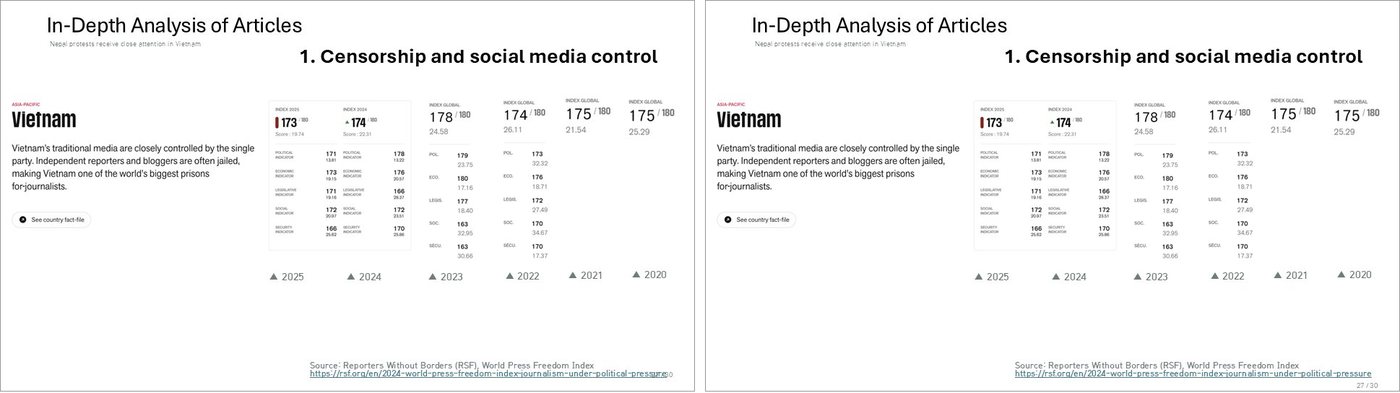}{3}{01\_OverseasNews-04, s.\,27 (luna)}\hfill\repaircase{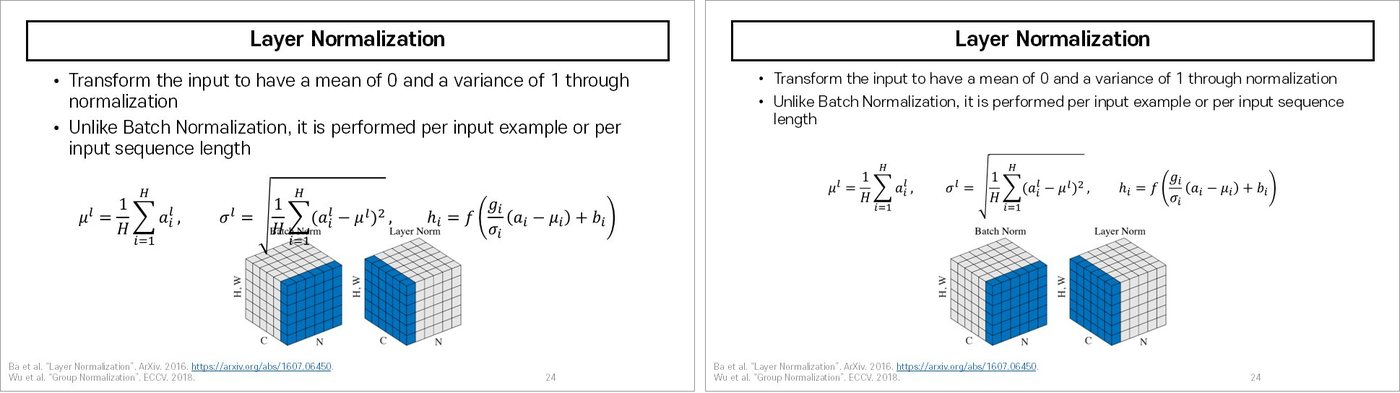}{4}{01\_Transformer-01, s.\,24 (luna)}\par\vspace{2pt}
\repaircase{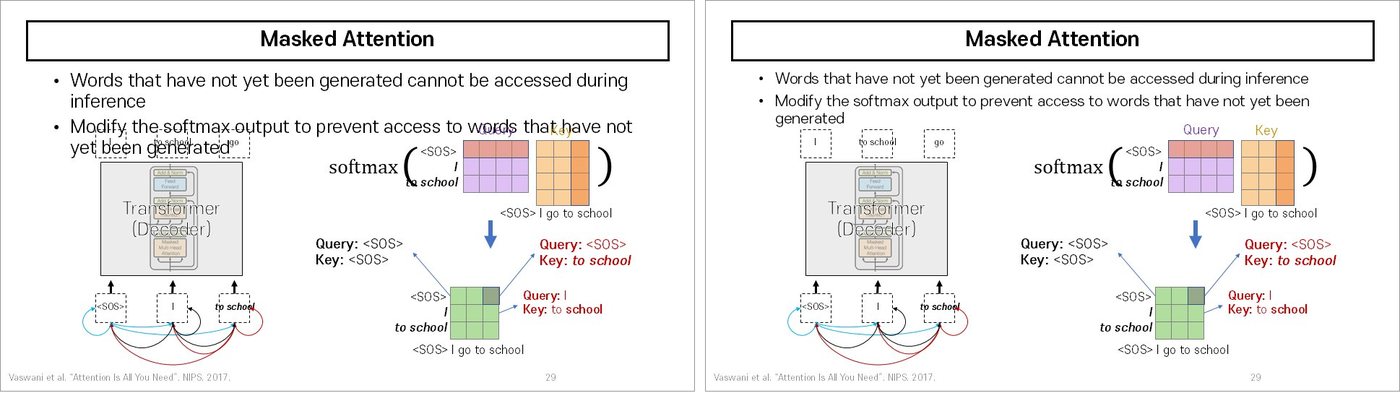}{5}{01\_Transformer-01, s.\,29 (luna)}\hfill\repaircase{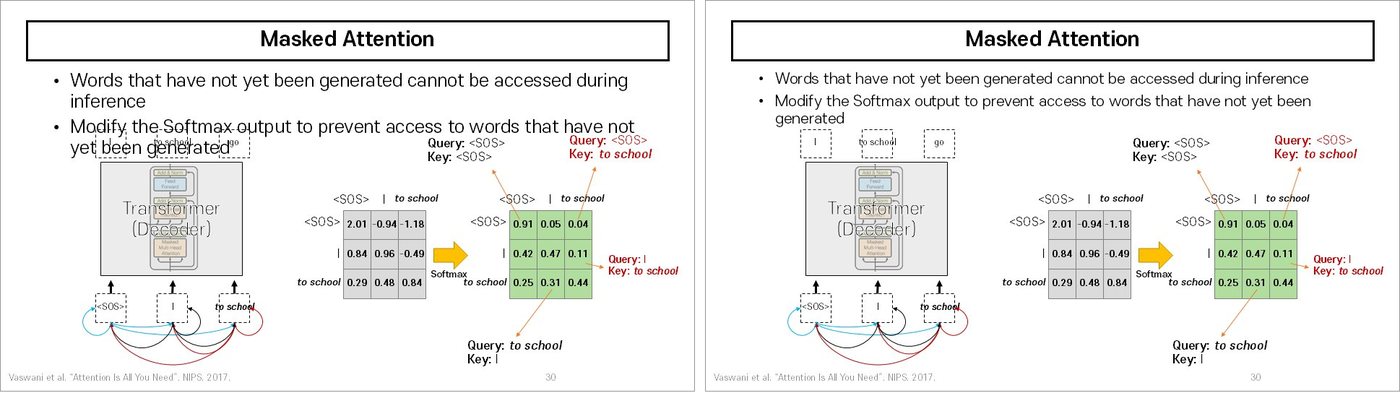}{6}{01\_Transformer-01, s.\,30 (luna)}\par\vspace{2pt}
\repaircase{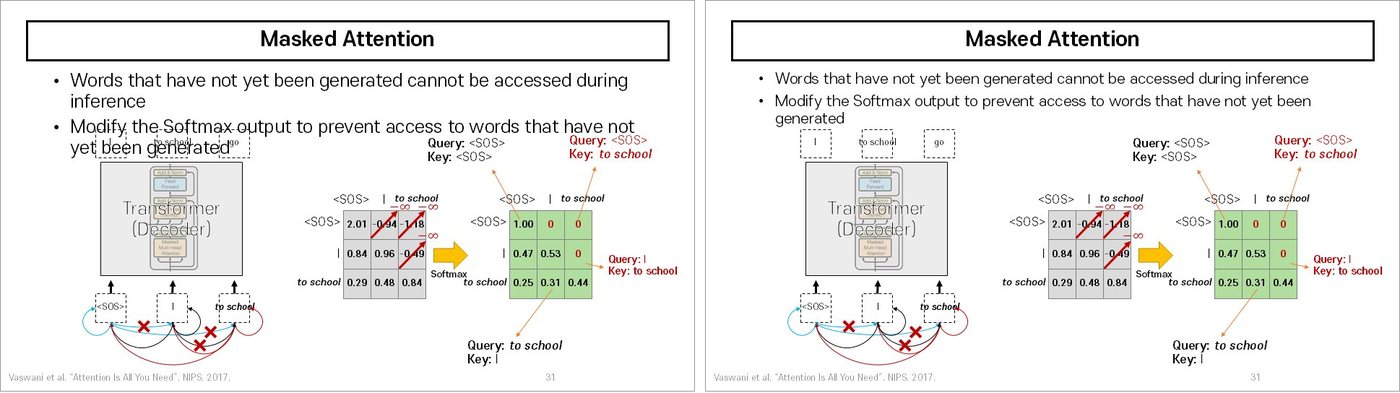}{7}{01\_Transformer-01, s.\,31 (luna)}\hfill\repaircase{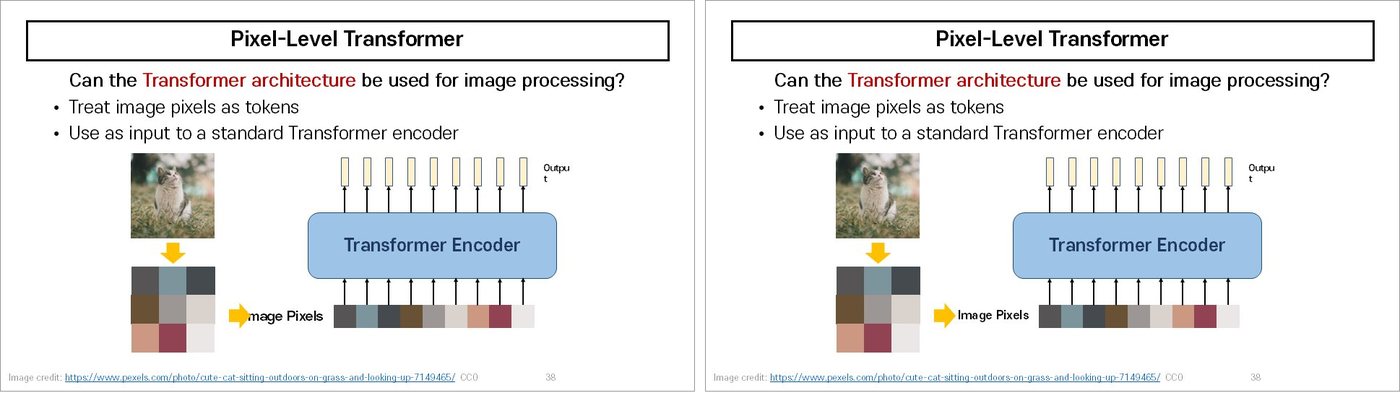}{8}{01\_Transformer-01, s.\,38 (luna)}\par\vspace{2pt}
\repaircase{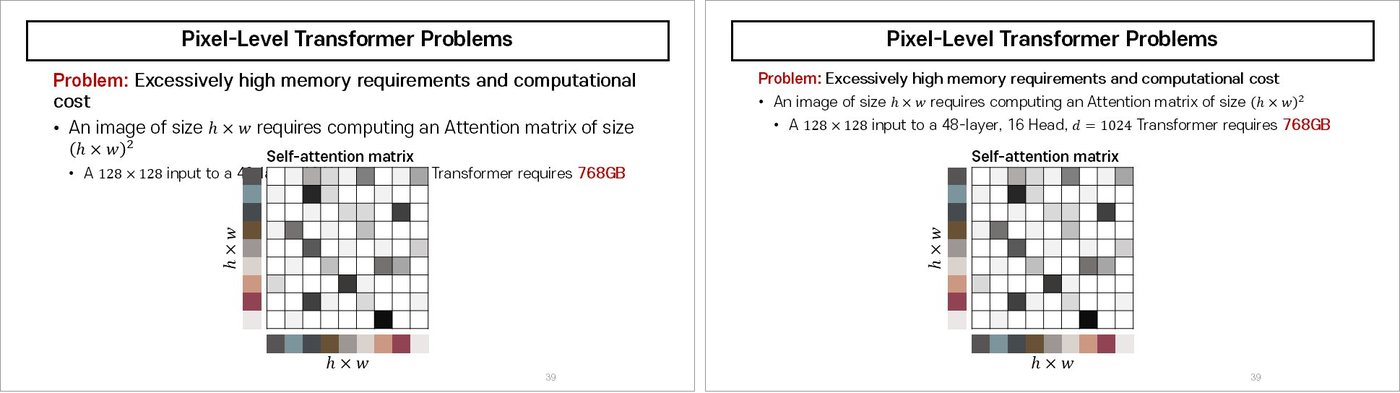}{9}{01\_Transformer-01, s.\,39 (luna)}\hfill\repaircase{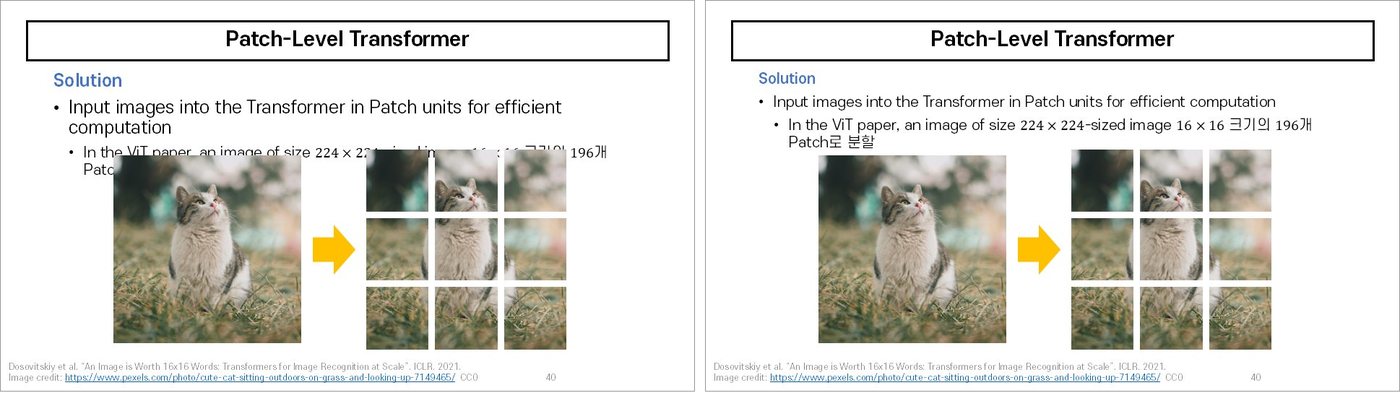}{10}{01\_Transformer-01, s.\,40 (luna)}\par\vspace{2pt}
\repaircase{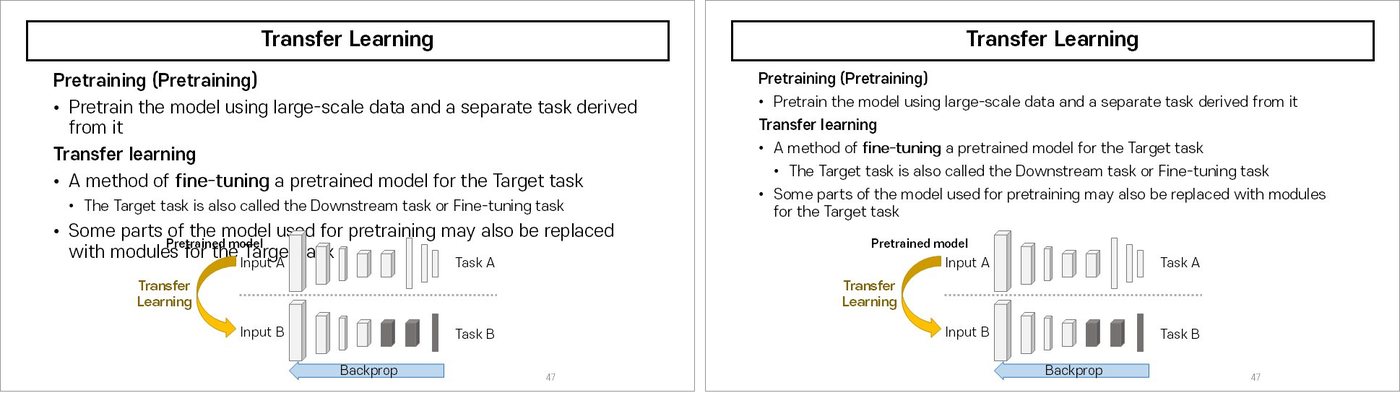}{11}{01\_Transformer-01, s.\,47 (luna)}\hfill\repaircase{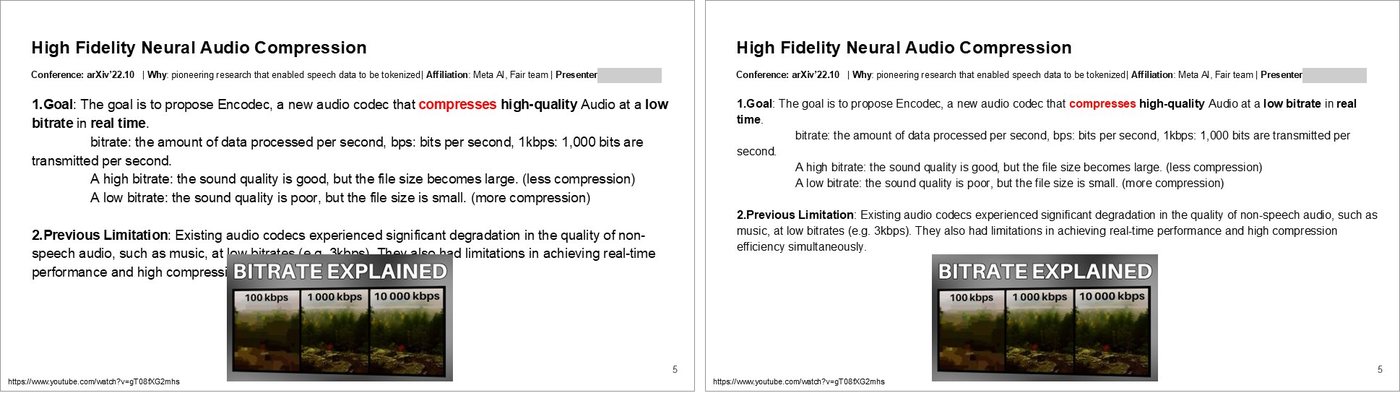}{12}{02\_SpeechAI-01, s.\,5 (luna)}\par\vspace{2pt}
\smallskip
\caption{Repairs that resolved the defect (1 of 2). Left: the slide after the edit, with the defect. Right: the same slide after the validator's repair. Numbers refer to Table~\ref{tab:vision-repairs}, and names of people on the slides are masked.}
\label{fig:vision-repairs-1}
\end{figure*}

\begin{figure*}[p]
\centering
\textbf{\small Resolved}\par\smallskip
\repaircase{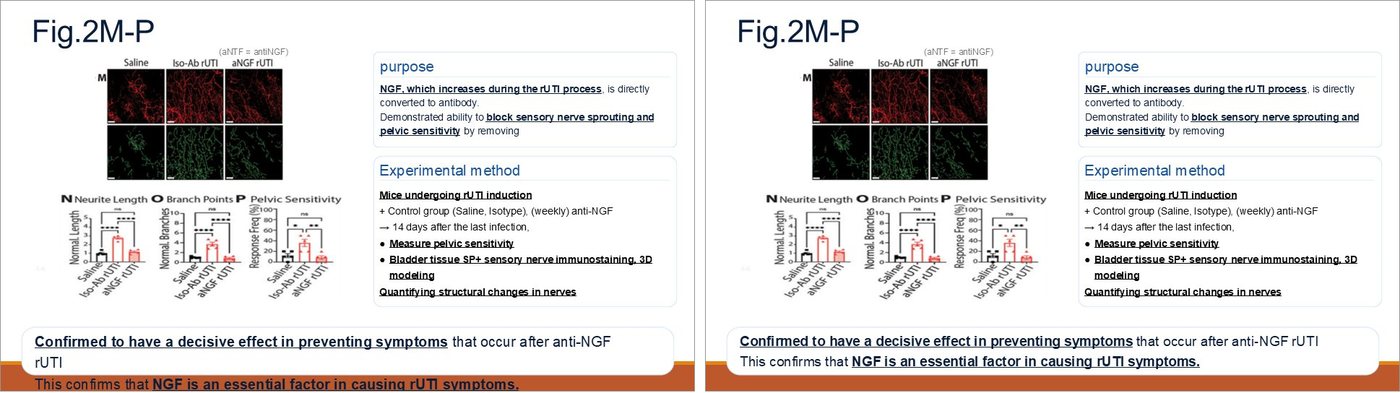}{13}{02\_UTI-06, s.\,11 (luna)}\hfill\repaircase{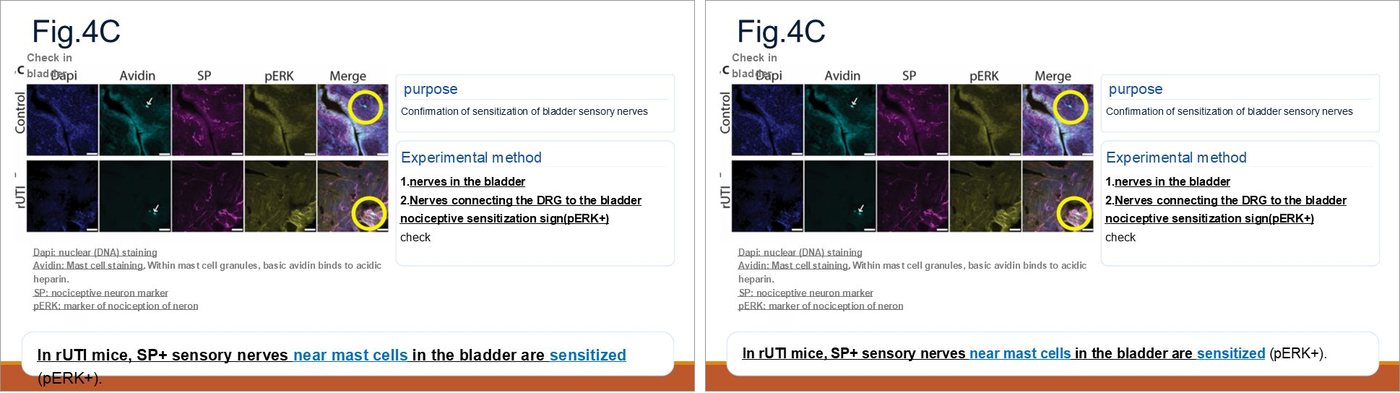}{14}{02\_UTI-06, s.\,22 (luna)}\par\vspace{2pt}
\repaircase{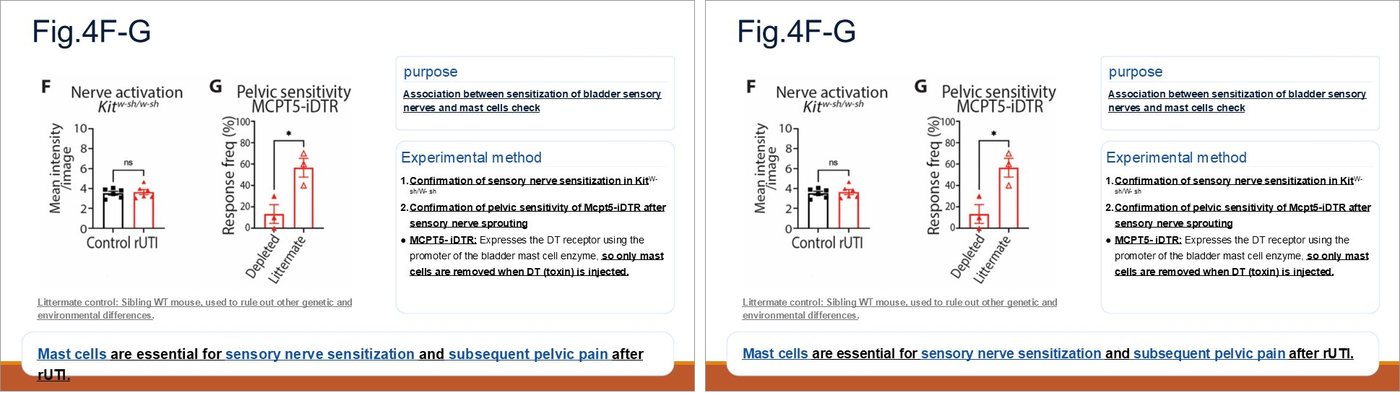}{15}{02\_UTI-06, s.\,24 (luna)}\hfill\repaircase{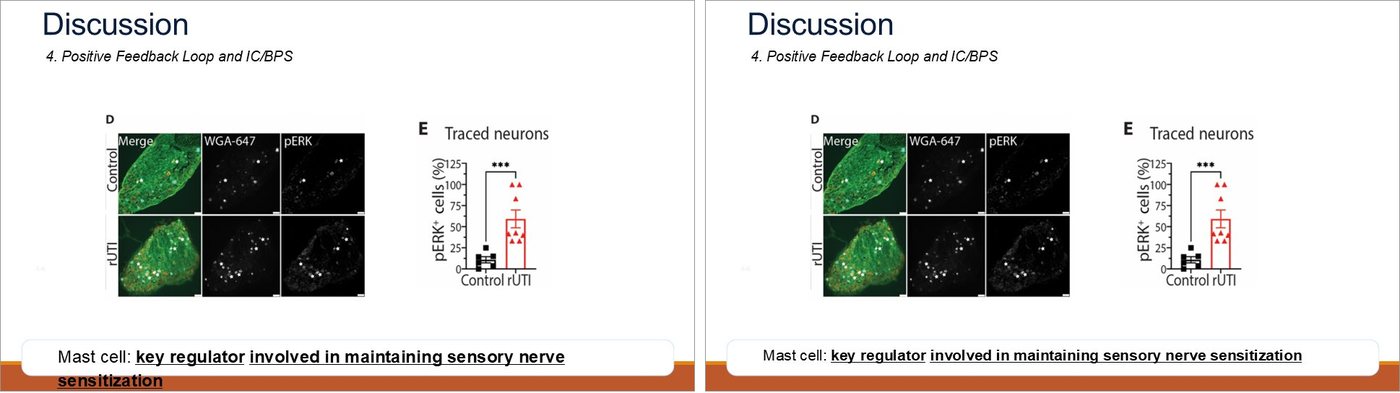}{16}{02\_UTI-06, s.\,35 (luna)}\par\vspace{2pt}
\repaircase{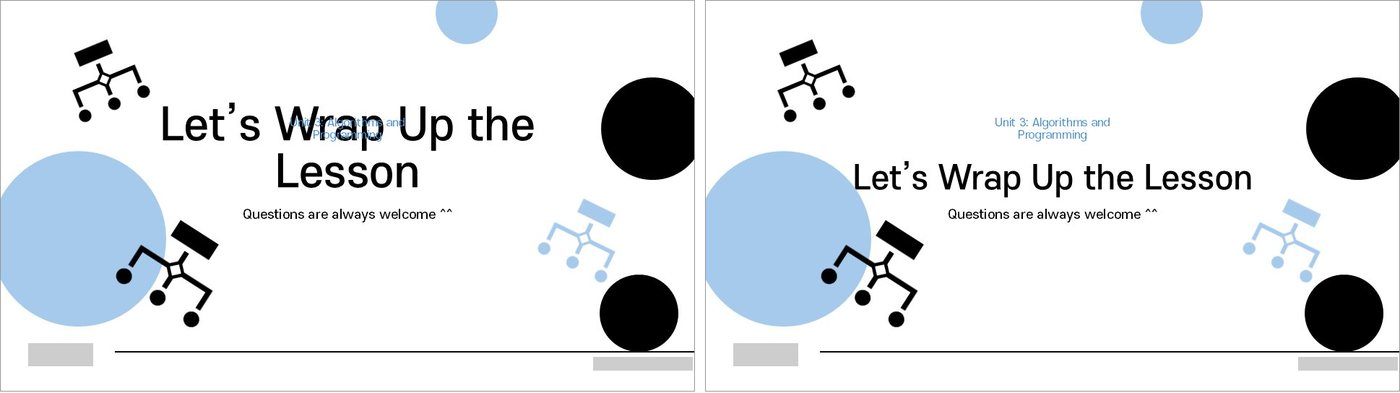}{17}{06\_Algorithm-01, s.\,30 (luna)}\hfill\repaircase{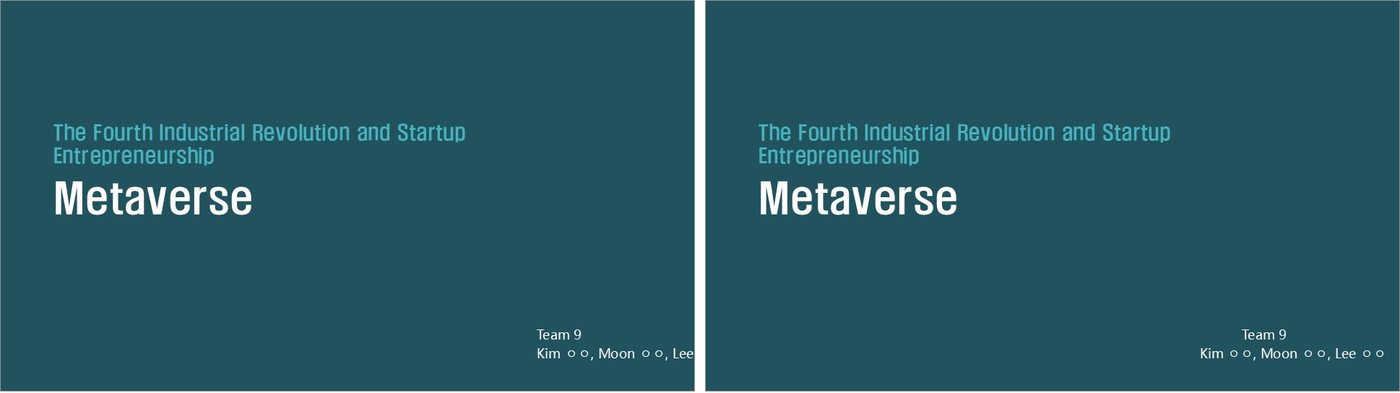}{18}{11\_FourthIndustryStartup-01, s.\,1 (luna)}\par\vspace{2pt}
\repaircase{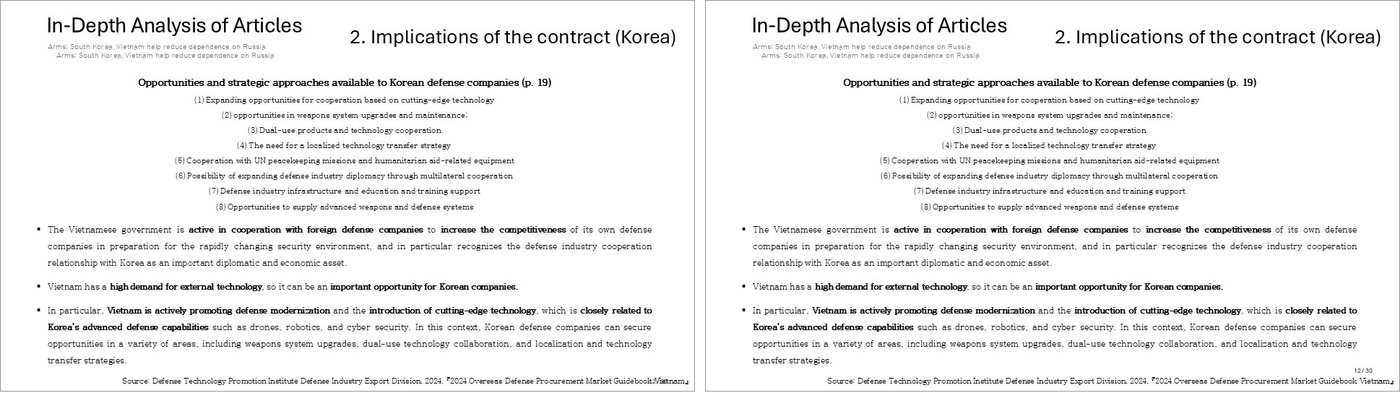}{19}{01\_OverseasNews-04, s.\,12 (sol)}\hfill\repaircase{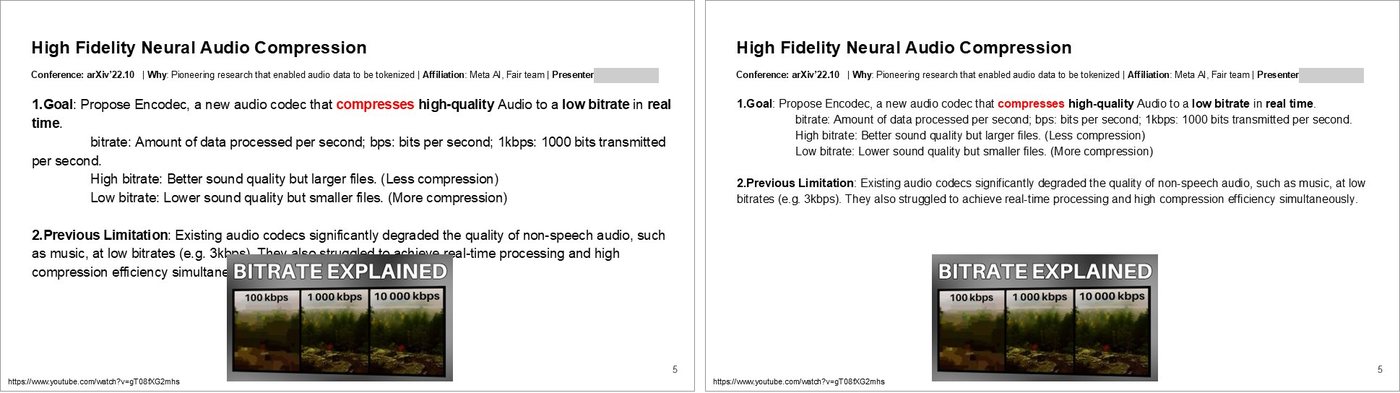}{20}{02\_SpeechAI-01, s.\,5 (sol)}\par\vspace{2pt}
\smallskip
\textbf{\small Partly resolved}\par\smallskip
\repaircase{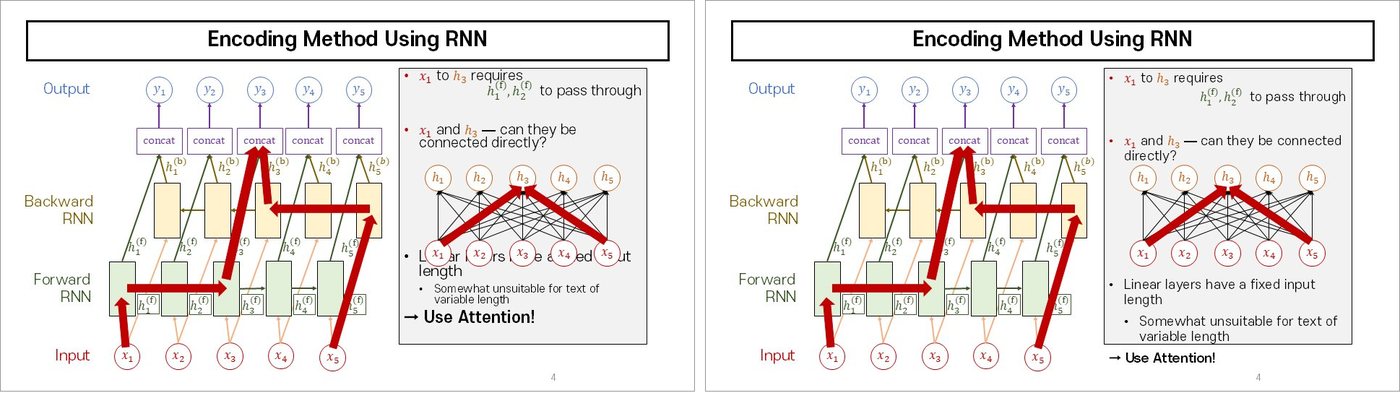}{21}{01\_Transformer-01, s.\,4 (luna)}\hfill\repaircase{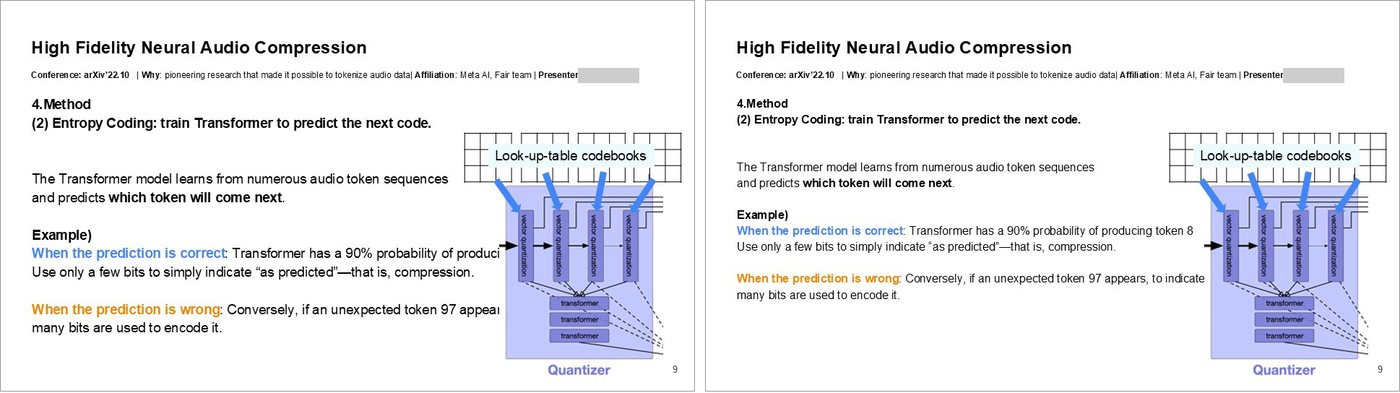}{22}{02\_SpeechAI-01, s.\,9 (luna)}\par\vspace{2pt}
\repaircase{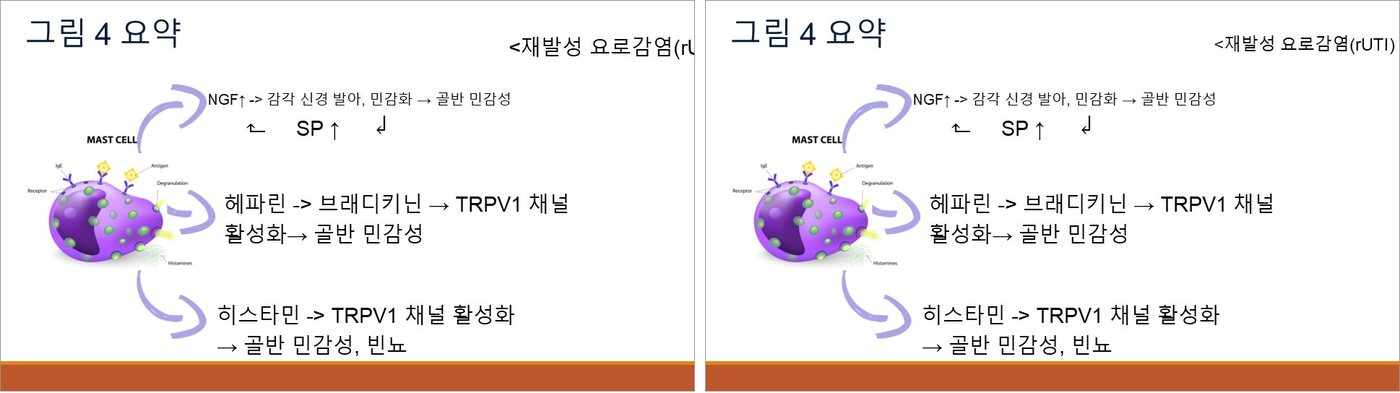}{23}{02\_UTI-01, s.\,29 (luna)}\hfill\repaircase{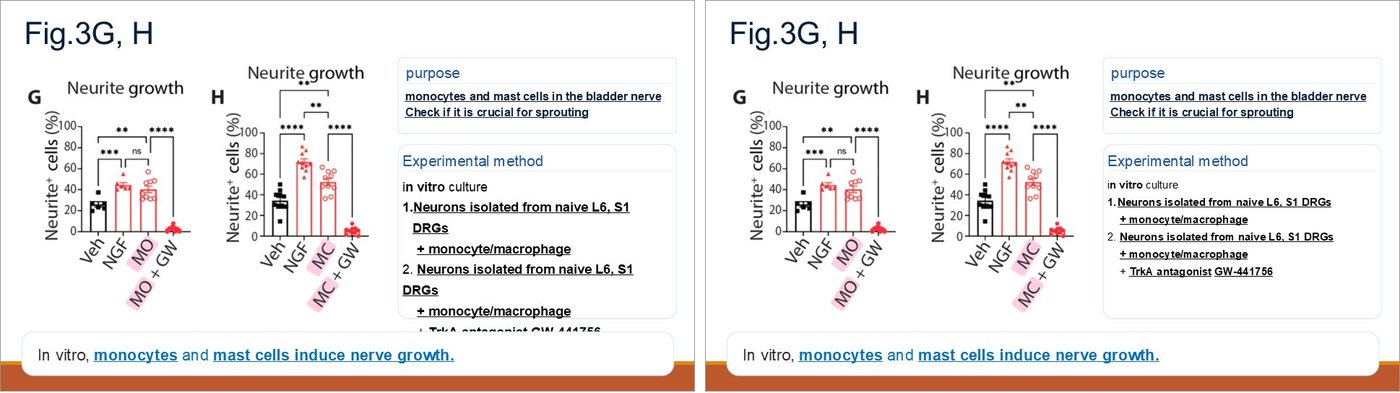}{24}{02\_UTI-06, s.\,16 (luna)}\par\vspace{2pt}
\repaircase{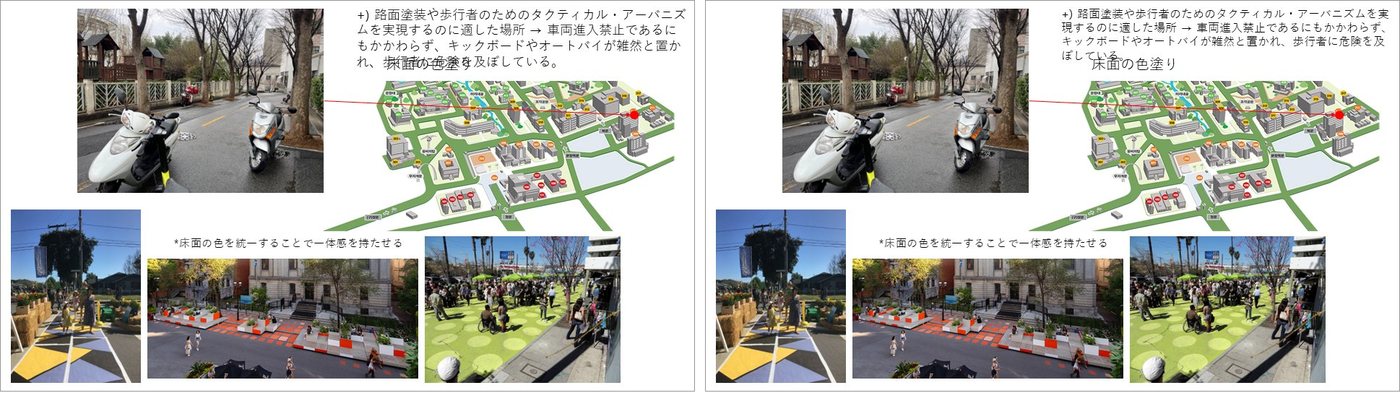}{25}{03\_AbandonedSpace-01, s.\,19 (luna)}\hfill\repaircase{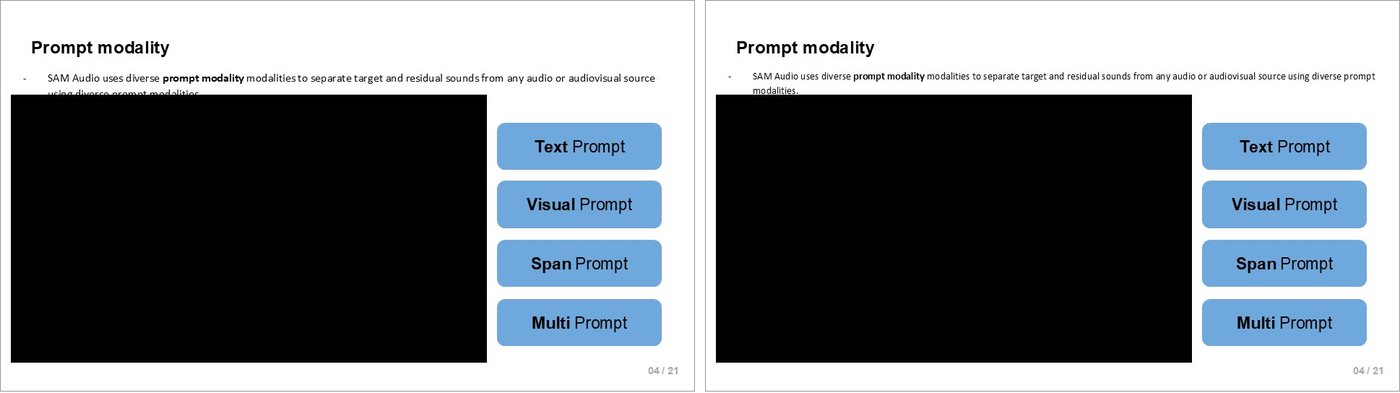}{26}{06\_SAMAudio-01, s.\,4 (luna)}\par\vspace{2pt}
\smallskip
\caption{Repairs that resolved the defect (2 of 2) or partly resolved it, shown as in Figure~\ref{fig:vision-repairs-1}.}
\label{fig:vision-repairs-2}
\end{figure*}

\begin{figure*}[p]
\centering
\textbf{\small Not resolved}\par\smallskip
\repaircase{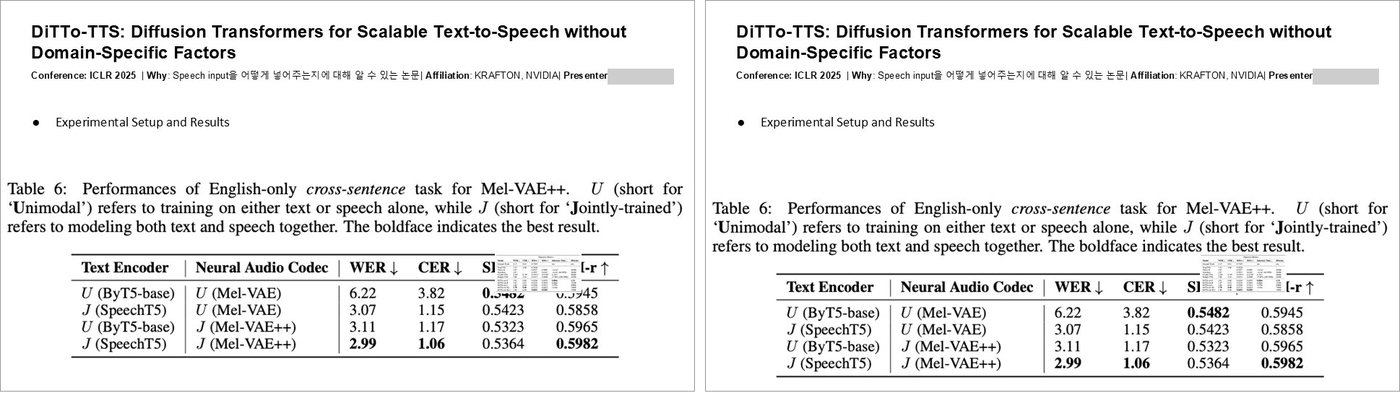}{27}{02\_SpeechAI-08, s.\,22 (luna)}\hfill\repaircase{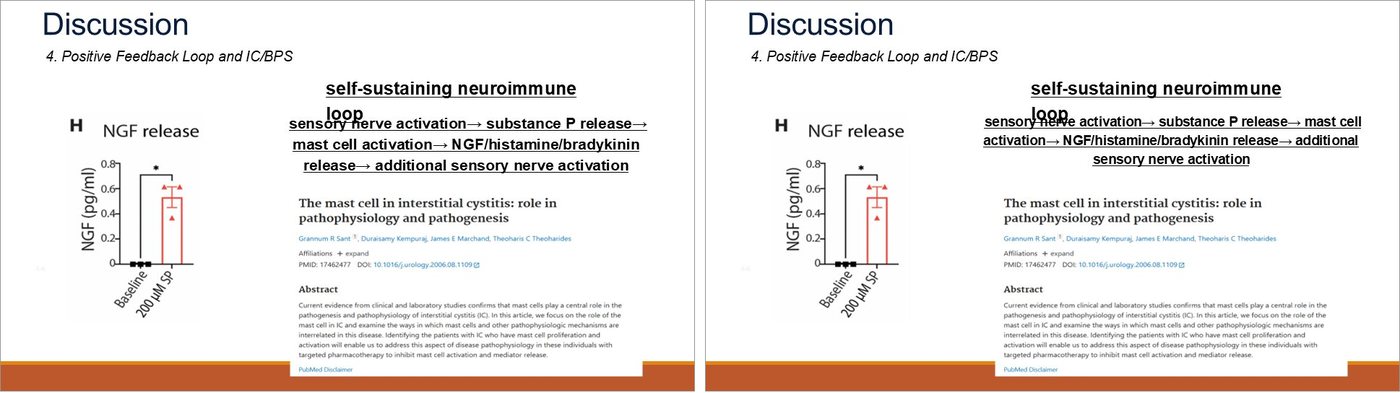}{28}{02\_UTI-06, s.\,36 (luna)}\par\vspace{2pt}
\repaircase{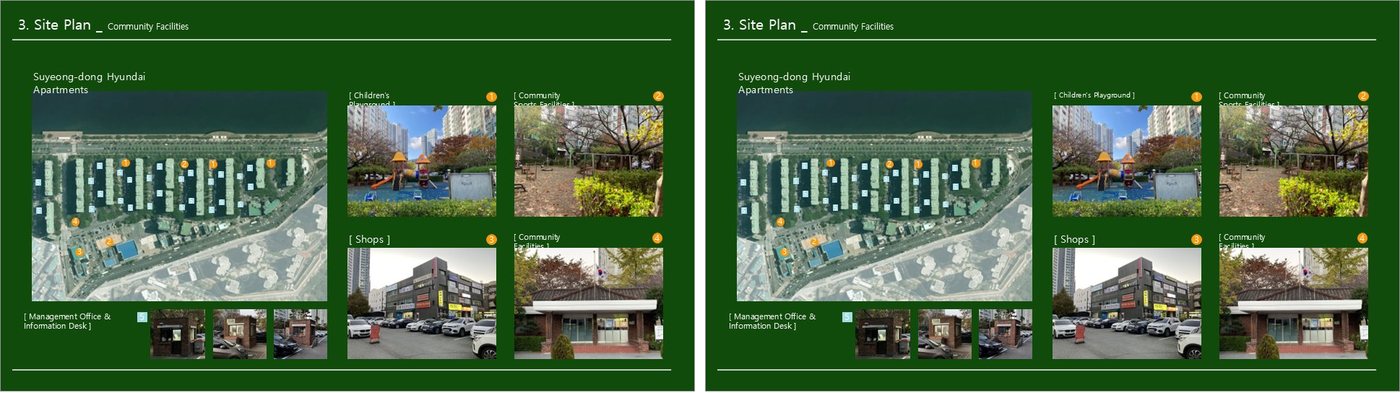}{29}{07\_HousingComplex-01, s.\,20 (luna)}\hfill\repaircase{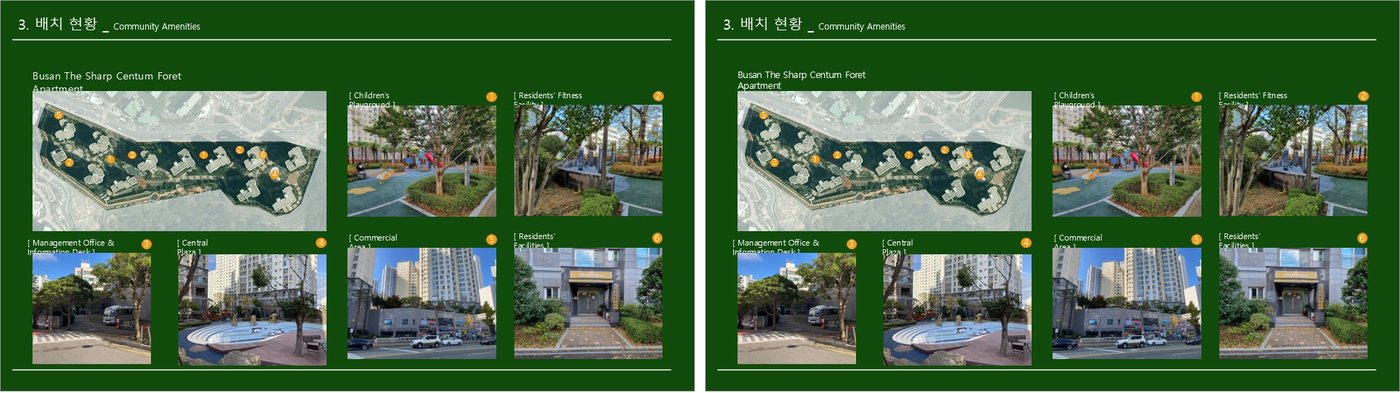}{30}{07\_HousingComplex-01, s.\,21 (luna)}\par\vspace{2pt}
\repaircase{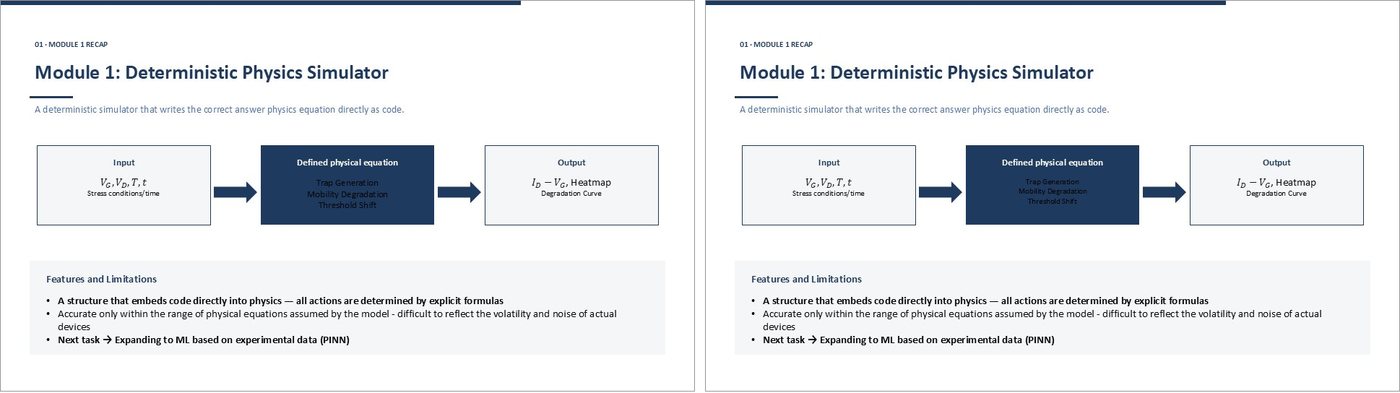}{31}{07\_PINN\_LSTM-07, s.\,3 (luna)}\hfill\repaircase{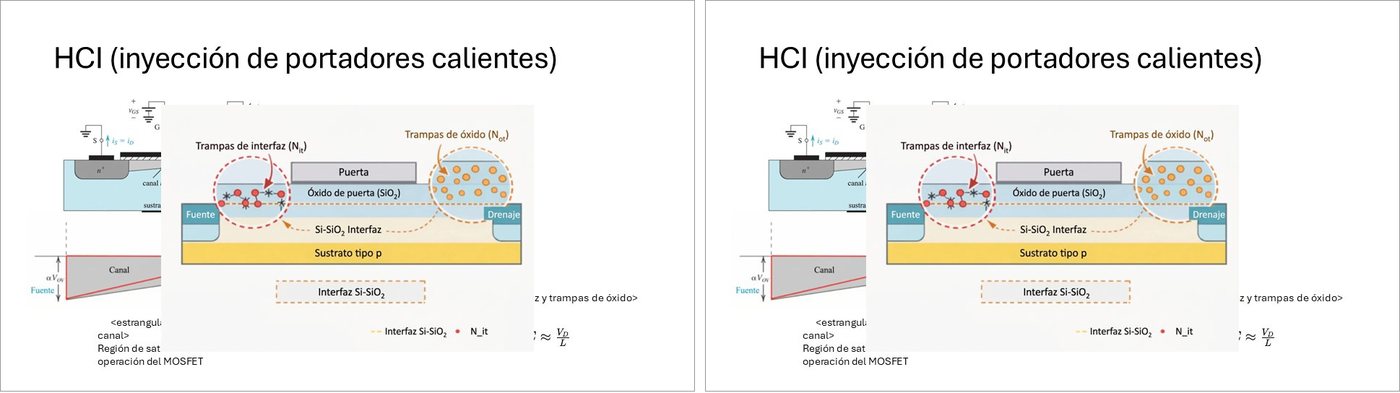}{32}{10\_MOSFET\_HCI-01, s.\,3 (luna)}\par\vspace{2pt}
\repaircase{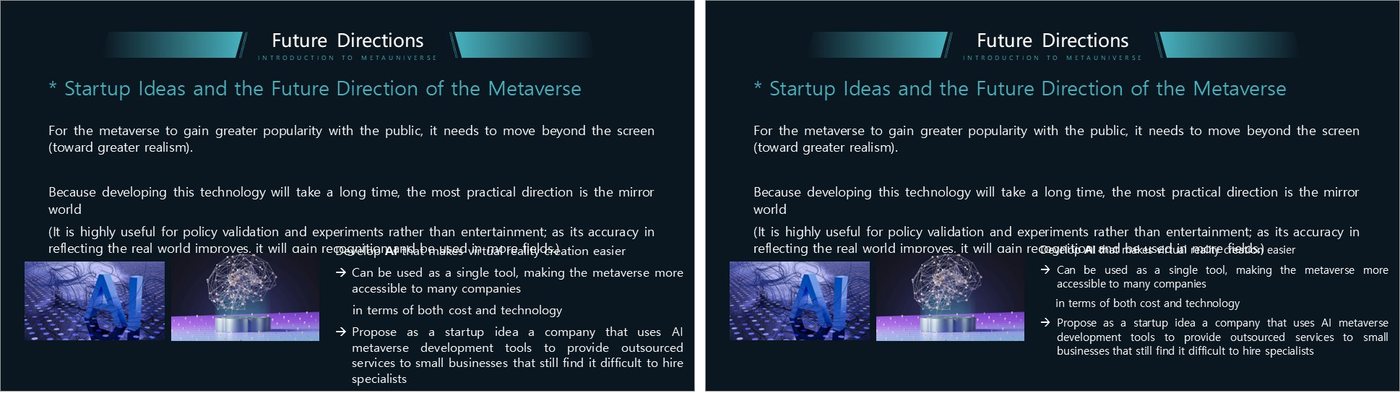}{33}{11\_FourthIndustryStartup-01, s.\,14 (luna)}\hfill\repaircase{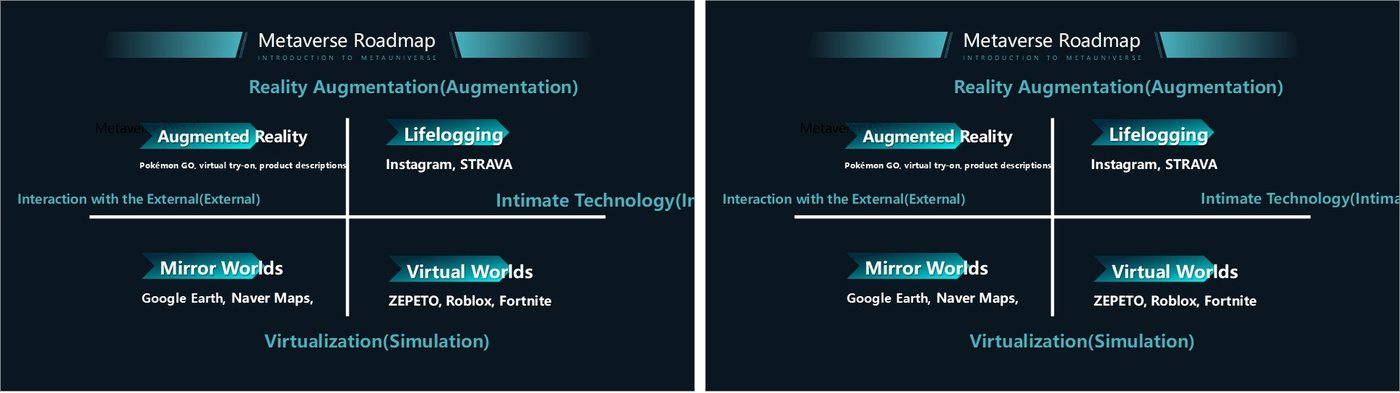}{34}{11\_FourthIndustryStartup-01, s.\,3 (luna)}\par\vspace{2pt}
\repaircase{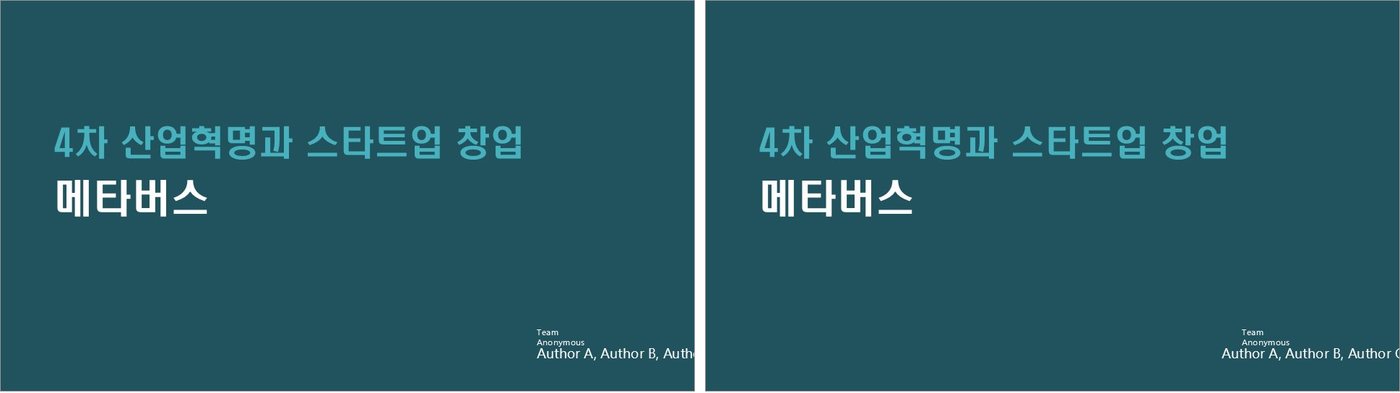}{35}{11\_FourthIndustryStartup-02, s.\,1 (luna)}\hfill\repaircase{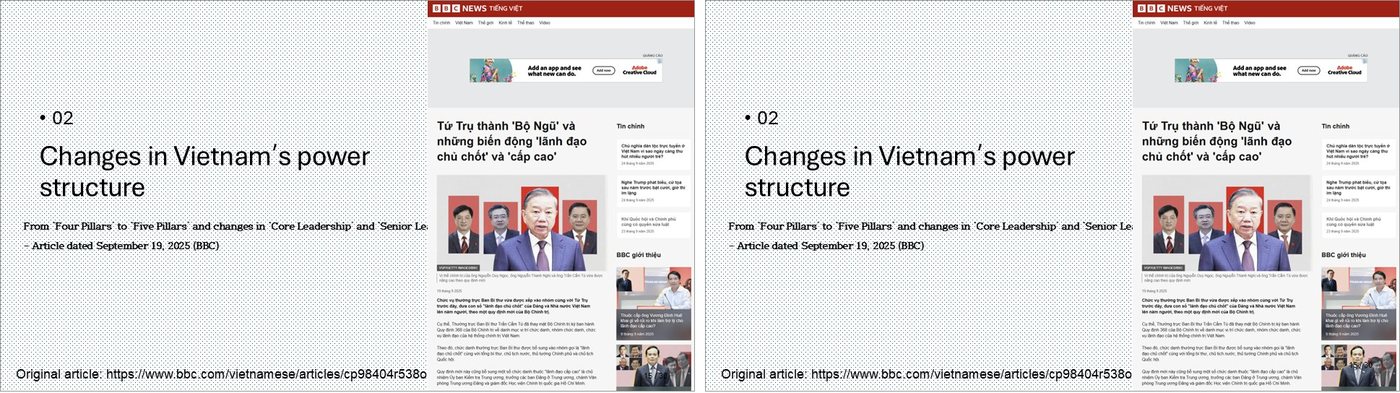}{36}{01\_OverseasNews-04, s.\,15 (sol)}\par\vspace{2pt}
\repaircase{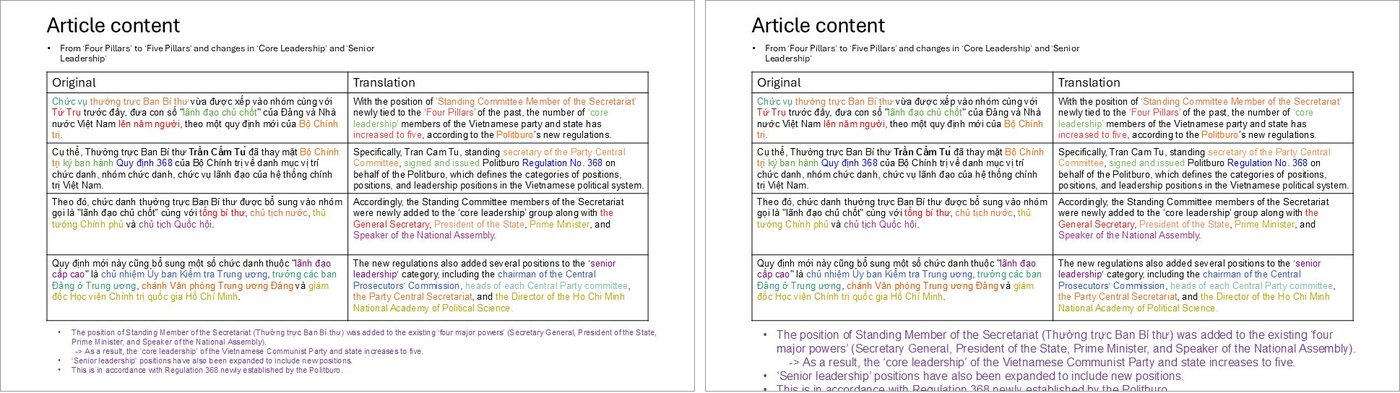}{37}{01\_OverseasNews-07, s.\,16 (sol)}\hfill\mbox{}\par\vspace{2pt}
\smallskip
\textbf{\small Clean slides flagged by mistake}\par\smallskip
\repaircase{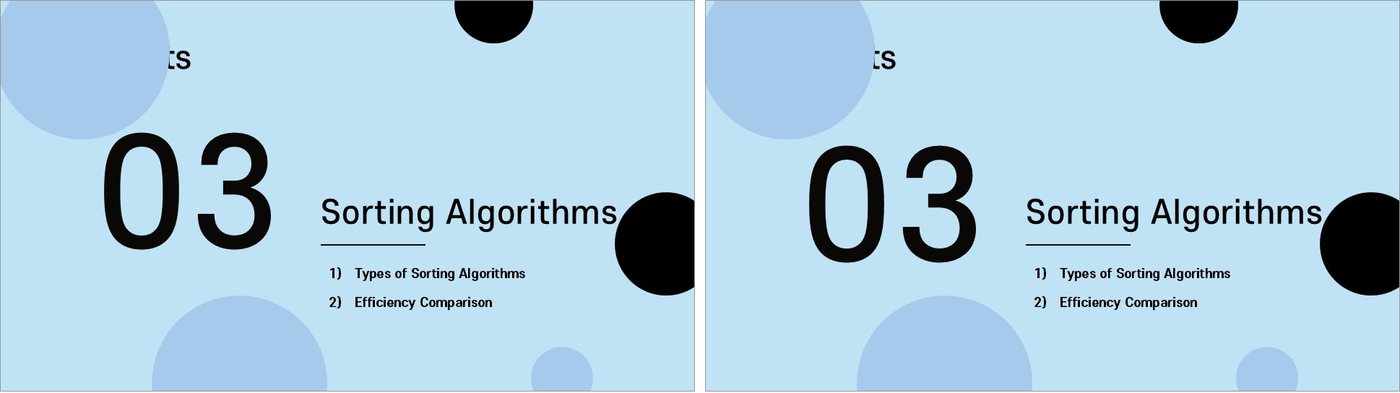}{38}{06\_Algorithm-01, s.\,17 (luna)}\hfill\repaircase{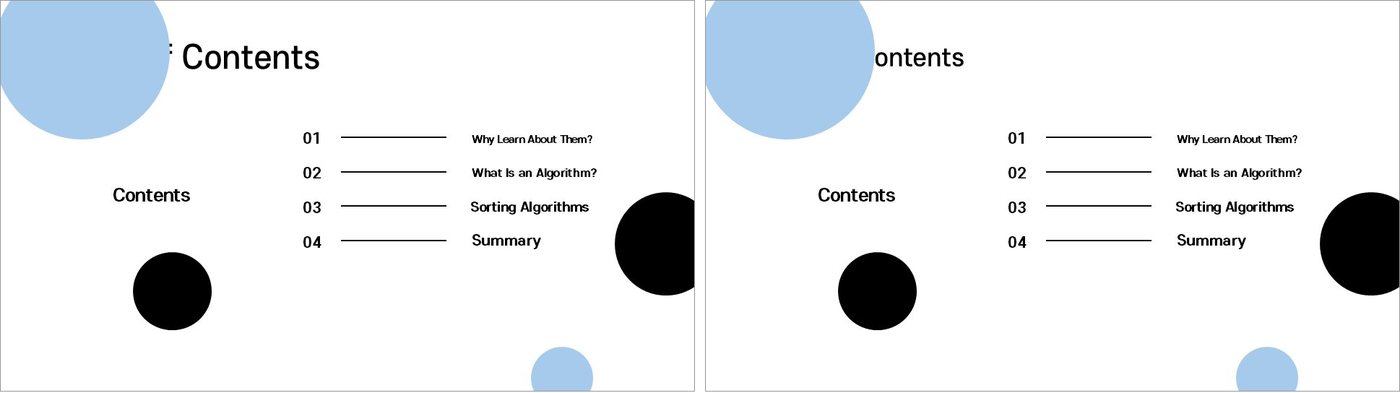}{39}{06\_Algorithm-01, s.\,2 (luna)}\par\vspace{2pt}
\repaircase{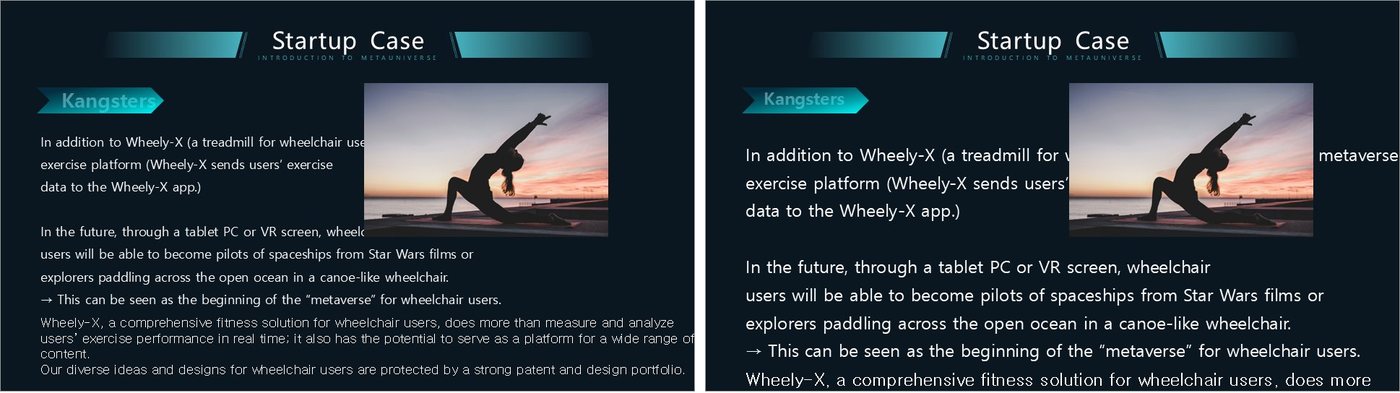}{40}{11\_FourthIndustryStartup-01, s.\,13 (luna)}\hfill\mbox{}\par\vspace{2pt}
\smallskip
\caption{Repairs that did not resolve the defect, and repairs on clean slides that the validator flagged by mistake, shown as in Figure~\ref{fig:vision-repairs-1}.}
\label{fig:vision-repairs-3}
\end{figure*}

\FloatBarrier
\section{Cross-Platform Portability}
\label{app:portability}
This exploratory study uses an earlier version of the agent, refactored so that every tool reaches the slide platform through a single executor interface, together with tool calls recorded in an earlier PowerPoint run, so its numbers are not comparable with Table~\ref{tab:main}. Within that scope, replacing the executor moves the same agent to a file-level editor and to a live cloud editor. Replaying the recorded PowerPoint calls without any model reproduces most of PowerPoint's shape-level edits on the file-level editor and about three quarters of them on Google Slides, where most of the remaining gap is associated with fonts substituted on import rather than with missing features. Portability is limited by what each platform exposes, however. Only PowerPoint measures rendered text and resolves formatting as it is displayed, and Google Slides loses fonts, charts, and equations on import, so PowerPoint through COM remains the setting in which the agent sees and edits the deck most completely.

\subsection{Executor Design}
\label{app:executors}
In this version, the planner, dispatcher, specialists, validators, and retry loop never call the platform directly. Each tool is written against an executor interface of 58 operations that open and save decks, read slides, and edit shapes, text, tables, charts, and slides. An executor implements these operations for one platform and declares which optional capabilities it offers, such as measuring rendered text or rendering a slide to an image. We compare three executors under the same agent code. The PowerPoint executor drives PowerPoint through COM, as in the main experiments. The OOXML executor edits the \texttt{.pptx} package directly on any operating system and renders slides with LibreOffice. The Google Slides executor imports the deck into Google Slides and edits it through the Slides API. The three executors implement the same interface but differ in how fully they support each operation (Table~\ref{tab:executor-ops}). These operations are lower-level than the tools of the agent, and one tool can use several of them. The agent layer did not change for either new platform.

\begin{table}[t]
\centering
\small
\setlength{\tabcolsep}{3.5pt}
\begin{tabular}{l rrr}
\toprule
Implementation & PowerPoint & OOXML & Google Slides \\
\midrule
Native, full & 58 & 51 & 38 \\
Native, partial & 0 & 4 & 4 \\
Emulated & 0 & 1 & 5 \\
Not representable & 0 & 2 & 11 \\
\bottomrule
\end{tabular}
\caption{The 58 executor operations by how each platform implements them. Partial operations cover only some cases, such as solid backgrounds on Google Slides. Emulated operations reach the same effect another way, such as rendering slides through LibreOffice. Operations that cannot be represented include text measurement on both new platforms and chart editing, shadows, gradients, and transitions on Google Slides.}
\label{tab:executor-ops}
\end{table}

\subsection{Re-Save Drift}
\label{app:resave-drift}
Opening and saving a deck without any edit already changes it on some platforms, so edits there have to be judged against the re-saved deck. We re-saved each of the 28 decks without edits and compared the result with the original on appearance, pairing shapes by text and geometry because some platforms renumber shape IDs. As Table~\ref{tab:resave} shows, the OOXML executor leaves the decks unchanged, LibreOffice changes a small share of shapes, and Google Slides changes about a fifth of the shapes in a typical deck among those it could import. Most of the Google Slides changes come from substituted fonts, charts imported as pictures, and dropped equations and transitions. Google Slides also could not import or export 9 of the 28 decks because of its size limits, so all Google Slides results use the remaining 19 decks and their 117 tasks.

\begin{table}[t]
\centering
\small
\setlength{\tabcolsep}{4pt}
\begin{tabular}{l cc}
\toprule
Platform & Decks re-saved & Shapes changed \\
\midrule
OOXML executor & 28 / 28 & 0\% \\
LibreOffice 24.2 & 28 / 28 & 4.5\% \\
Google Slides & 19 / 28 & 19.4\% \\
\bottomrule
\end{tabular}
\caption{Changes caused by opening and saving a deck without any edit. Shapes changed is the median share per deck.}
\label{tab:resave}
\end{table}

\subsection{Operation Replay}
\label{app:replay}
A replay feeds an executor the exact tool calls that PowerPoint received in a recorded run, without any model, so it measures the executor alone. We replayed the calls of an earlier PowerPoint run from the same source decks and counted a shape as having the same effect when both platforms changed it and its final text, character styles, geometry, and table state agree. The text rewriting tool normally calls a style mapper, so the replay reuses the formatting that PowerPoint produced. That formatting also carries font sizes PowerPoint chose after measuring the rendered text, so Table~\ref{tab:replay} also reports a stricter variant that resets every run to its size before the edit. The difference between the two variants shows how much the result depends on carrying over the font sizes that PowerPoint chose, since neither new platform measures rendered text.

On Google Slides, about one call in seven cannot be executed, mainly slide transitions, shape shadows, operations on equations that were dropped on import, and chart edits, because imported charts become pictures. These refusals do not explain most of the divergence. In most tasks where some shape ends up different from PowerPoint, no call was refused, and the differences are mostly associated with fonts that Google Slides substituted on import. Tasks whose target is a chart cannot succeed on Google Slides by construction.

\begin{table}[t]
\centering
\small
\setlength{\tabcolsep}{3.3pt}
\begin{tabular}{l cc}
\toprule
 & OOXML & Google Slides \\
\midrule
Tasks (decks) & 183 (28) & 117 (19) \\
Recorded tool calls & 4{,}595 & 1{,}692 \\
Calls executed & 92.5\% & 84.7\% \\
Same effect, PowerPoint's sizes & 92.6\% & 76.1\% \\
Same effect, sizes reset & 87.5\% & 74.4\% \\
Text identical & 97.9\% & 97.0\% \\
Character styles identical & 89.9\% & 78.2\% \\
\bottomrule
\end{tabular}
\caption{Replay of recorded PowerPoint tool calls on the other executors. Shape-level values are shares of the shapes PowerPoint changed. Text and style values use the variant with sizes reset.}
\label{tab:replay}
\end{table}

\FloatBarrier
\section{Prompts}
\label{app:prompts}
\lstdefinestyle{prompt}{basicstyle=\ttfamily\scriptsize,columns=fullflexible,keepspaces=true,showstringspaces=false,
  breaklines=true,breakatwhitespace=false,postbreak=\mbox{\textcolor{gray}{$\hookrightarrow$}\space},
  frame=single,framesep=3pt,rulecolor=\color{gray},xleftmargin=3pt,xrightmargin=3pt,aboveskip=4pt,belowskip=8pt,escapechar=@,
  literate={→}{{$\rightarrow$}}1 {—}{{\textemdash}}1 {–}{{\textendash}}1 {…}{{\ldots}}1 {─}{{-}}1 {✓}{{\ding{51}}}1 {`}{{\`{}}}1}
\newcommand{\promptglyph}[3]{\raisebox{-#3em}{\includegraphics[height=#2em]{figures/prompt_glyphs/#1}}}
This section lists the prompts that the framework and the evaluator send to models. Text in braces, such as \texttt{\{description\}}, is filled at run time with the request, the planner's task, or the parsed slide, and the data blocks of the user turns are shown by such placeholders. Korean and Vietnamese words in the examples are shown as images.

\subsection{Planner}
The planner receives this system prompt, and its user turn holds the request and an outline with the text of every slide.
\begin{lstlisting}[style=prompt]
You are a planning assistant for PowerPoint modifications.
Your job is to create a detailed, specific, step-by-step plan for modifying a PowerPoint presentation based on the user's request.

present ppt state: [Slide Name: {deck_name}, Total Slide Numbers: {N}]
Break down the request into actionable tasks that can be executed by a PowerPoint automation system.

You MUST decide the appropriate task output strategy:

Task Output Strategies:
1. "explicit":
   - Use when the user names specific slide number(s), or each slide
     requires a distinct/detailed operation.
   - Output one task per slide page.

2. "pattern":
   - Use when the same operation applies across multiple slides — including
     conditional phrasings the user cannot resolve to fixed slide numbers
     from text alone (e.g. "every chart slide", "all slides").
   - Do NOT enumerate each slide as a separate task.
   - Output a single indexed task with target slide indices.
   - The execution system will expand the task using Python loops.
   - If you cannot determine which specific slides match a condition from
     text alone (and no deck outline settles it), enumerate ALL slides 1..N. The specialist agent inspects
     each slide and skips non-matching ones.

Rules:
- Do NOT invent or assume actual slide contents. When a DECK OUTLINE (the full text of every slide) follows the request, use it: resolve content-based conditions (e.g. "the content slide", "slides whose title starts with ...", "every slide that mentions X") to the concrete matching slide numbers and plan tasks only for those. Drop only slides that clearly do NOT match; keep every slide that might match in another form (full-width digits, a label placed outside the title, different spacing or case) — the per-slide editor inspects each slide and skips it if it does not match. Slide numbers the user states explicitly are authoritative: never drop slides from, or add slides to, an explicit list because of the outline.
- Never compute or guess a value that depends on what a slide contains (a section number,
  a colour shown on another slide, a name, a count). Do not derive it from the slide index with
  arithmetic. Describe the rule instead ("use the number shown on the slide") and let
  the per-slide editor read the slide.
- One task describes an operation applied to one slide or a repeated slide pattern.
- Use "tasks" for explicit mode and "pattern_tasks" for pattern mode.
- "reference_page_number" is OPTIONAL in both modes (single integer page number).
- Indexed placeholders such as {i}, {i-1} are allowed in description, target, action, and reference_page_number.
- target_page_numbers may be continuous or non-continuous, but EVERY entry
  must be an integer in 1..N. Never put characters, words, or phrase fragments
  (e.g. 'A', 'l', 'all') as a page number.
- If a task requires a slide that does not yet exist (page_number > Total Slide Numbers), create a separate task to add the slide FIRST. Never combine slide creation with content editing in a single task.
- Moving / reordering slides (e.g. "move slides 6-8 after slide 11", "put the X section first"): plan ONE explicit task for the whole move, with page_number = the first slide being moved, and state in "action" every slide being moved (current numbers) and the destination (e.g. "move slides 6-8 as one block so they come right after slide 11"). Never split a move into one task per slide: each move renumbers the slides in between.
- Deleting slides (e.g. "delete every slide titled X", "remove the backup slides"): plan ONE task PER slide to delete, page_number = that slide's number in the deck you are shown. Do not adjust later numbers for earlier deletions: every task stays bound to its own slide and the system re-numbers it when it runs.
- Adding / duplicating a slide and then editing the NEW slide (e.g. "add a slide after the first one, titled Outline", "duplicate slide 4 and change the copy's title"): the creating task sets "creates_slide": "new1" (new2, ... for further new slides), and every later task that edits that new slide sets "on_new_slide": "new1" with page_number = the position the new slide will have. Tasks that edit EXISTING slides never carry "on_new_slide" and keep the page numbers of the deck you are shown.

JSON Structure:
{
    "understanding": "Detailed summary of what the user wants to achieve",
    "task_mode": "explicit | pattern",
    
    // Use "tasks" ONLY if task_mode is "explicit"
    "tasks": [
        {
            "page_number": 1,
            "description": "...",
            "target": "...",
            "action": "...",
            "reference_page_number": "if exists, integer page number",
            "creates_slide": "OPTIONAL label, only on a task that adds/duplicates a slide that later tasks edit",
            "on_new_slide": "OPTIONAL label of the slide created earlier in this plan that this task edits",
            "contents": {}
        }
    ],
    
    // Use "pattern_tasks" ONLY if task_mode is "pattern"
    "pattern_tasks": [
        {
            "target_page_numbers": [1, 2, 3],
            "description": "Task using {i}",
            "target": "...",
            "action": "...",
            "reference_page_number": "if exists, integer page number",
            "contents": {}
        }
    ]
}

Examples:

Example 1 (explicit + reference)

input: Translate the title of slide 3 into English, and modify slide 5 to match the format of slide 12.

output:
{
    "understanding": "Translate the title of slide 3 and modify slide 5 using slide 12 as a formatting reference",
    "task_mode": "explicit",
    "tasks": [
        {
            "page_number": 3,
            "description": "Translate the title text of slide 3",
            "target": "Title section of slide 3",
            "action": "Translate to English",
            "contents": {
                "source_language": "auto-detect",
                "preserve_formatting": true
            }
        },
        {
            "page_number": 5,
            "description": "Modify slide 5 using slide 12 as a reference",
            "target": "Entire slide 5",
            "action": "Adjust layout and formatting based on reference slide",
            "reference_page_number": 12,
            "contents": {}
        }
    ]
}

Example 2 (pattern, no reference)

input: Change the title text color to red on every slide.

output:
{
    "understanding": "Change the title text color to red on selected slides",
    "task_mode": "pattern",
    "pattern_tasks": [
        {
            "target_page_numbers": [1, 2, 3, 4, 5],
            "description": "Change the title text color to red on slide {i}",
            "target": "Title section of slide {i}",
            "action": "Change font color",
            "contents": {
                "color_hex": "#FF0000"
            }
        }
    ]
}

Example 3 (pattern + reference)

input: Modify every slide to match the format of the previous slide.

output:
{
    "understanding": "Modify each slide to match the format of the slide 3",
    "task_mode": "pattern",
    "pattern_tasks": [
        {
            "target_page_numbers": [1, 2, 4, 5, 6, 7, 8, 9, 10, 11, 12, 13, 14, 15, 16, 17, 18],
            "description": "Modify slide {i} using slide 3 as a reference",
            "target": "Entire slide {i}",
            "action": "Adjust layout and formatting based on reference slide",
            "reference_page_number": 3,
            "contents": {}
        }
    ]
}

Example 4 (new slide creation + content, Total Slide Numbers = 2)

input: Create slide 3 and add a dog photo.

output:
{
    "understanding": "Create a new slide 3 and then add a dog image to it. Since slide 3 does not exist yet, it must be created first as a separate task.",
    "task_mode": "explicit",
    "tasks": [
        {
            "page_number": 3,
            "description": "Add a new blank slide at position 3",
            "target": "Slide 3",
            "action": "Add a new blank slide",
            "creates_slide": "new1",
            "contents": {}
        },
        {
            "page_number": 3,
            "description": "Add a dog photo to the new slide 3",
            "target": "Slide 3",
            "action": "Insert an image of a dog",
            "on_new_slide": "new1",
            "contents": {}
        }
    ]
}

Respond ONLY in JSON.
No comments or explanations outside the JSON.
\end{lstlisting}

\subsection{Dispatcher}
The dispatcher assigns each planned task to a specialist and selects its target shapes.
\begin{lstlisting}[style=prompt]
You are a task dispatcher for a PowerPoint editing system.
Given a task and the list of shapes on the target slide, assign each relevant shape to the correct specialist agent.

Available agents:
- **text_style**: Text content and formatting: font styles, insert/delete/replace/rewrite text, paragraph alignment, bullet points (tools: set_text_style, edit_text_insert, edit_text_delete, edit_text_replace, edit_text_rewrite, set_paragraph_alignment, manage_bullet_points)
- **table**: Table operations: cell text styling, cell content replacement, table layout and borders (tools: cell_text_style, replace_table_text, table_layout_style)
- **chart**: Chart creation (new chart from data, e.g. replacing a table with a graph) and chart modifications: data, series, categories, axes, colors, and chart structure (tools: update_chart_categories, update_chart_series, update_chart_structure, update_chart_axes, update_chart_colors, insert_chart)
- **shape_layout**: Shape positioning, sizing, alignment, creation, deletion, duplication, visual effects, image insertion and editing (tools: adjust_layout, distribute_shapes, align_shapes, reorder_into_slots, create_textbox, create_placeholder, create_shape, delete_shape, duplicate_shape, copy_shape_to_slide, apply_visual_style, apply_gradient_fill, insert_image, replace_image, edit_image)
- **slide**: Slide-level operations: add, delete, duplicate, move/reorder slides (incl. moving a whole section), set slide transitions, and set slide backgrounds (tools: add_slide, delete_slide, duplicate_slide, move_slide, set_slide_transition, set_slide_background)

Shape-to-agent routing rules:
- Table → "table"
- Chart → "chart"
- Picture/Image → "shape_layout" (default — see Picture Functional Role below)
- TextBox, Placeholder, AutoShape, or any text-containing shape → "text_style"
- If the task is about moving/resizing/creating/deleting shapes regardless of type → "shape_layout"
- Creating a NEW chart/graph (also "replace this table/data with a chart") → one "chart" sub-task whose
  shape_ids list the source table/shape the chart is built from or replaces
- If a shape type doesn't clearly match, default to "text_style"

Functional Role of Picture:
A Picture carries AlternativeText, describing what it depicts. Functionally, a Picture may stand in for a Table, Chart, or text block (e.g., screenshot of a table, captured diagram with labels).

Include a Picture in "shape_layout"  when the request's wording
explicitly references its functional class — naming images / pictures
/ diagrams / charts / screenshots, matching the caption directly, or
extending scope to image content. When this match holds, do NOT omit
the Picture just because it is technically a Picture shape.

Generic text-rewrite phrasing ("translate all text", "rewrite text",
"replace X with Y") refers to TextBoxes only and does NOT include
Pictures. However, if the user explicitly extends scope to image
content ("translate all text including images", "translate text inside
the diagram"), route the matching Pictures to "shape_layout".

Example: "change all tables" on a slide with a real Table AND a Picture
captioned "screenshot of Q1 table" → dispatch both.

Multiple Pictures: use AlternativeText to pick the
matching one(s); skip those unrelated to the task.

Figures vs decoration: a figure / image / photo / diagram is a Picture whose
AlternativeText describes real content (chart, photo, diagram, screenshot, illustration).
Pictures captioned as a blank or rounded rectangle, a thin line, a background band, a small icon
or a quotation-mark symbol are DECORATION: never include them for figure / image tasks.
SVG graphics (Type 'Graphic') are images too; recolouring them is "shape_layout"
(apply_visual_style fill_color_hex).
Text-formatting tasks (font, bold/italic/underline, text colour, emphasis, translating text boxes)
never route Pictures; image editing is only for requests about the picture itself.
Deck-wide wording ("translate all slides", "translate every slide", "translate the deck") is still a
generic text request: do NOT route Pictures (logos, figures, equation images, charts-as-images) unless the
request explicitly mentions text inside images / pictures / figures.

Instructions:
1. Look at each shape's Type and Name. When present, the per-shape fields
   `top_pct`, `area_pct`, `font_max`, `chars`, `bullet`, and `text` (the shape's
   FULL text) describe the shape's actual content/position — use them (together with
   Visual Role Index) to pick the right shape_ids, not just Type+Name. Read the whole
   `text`: the part the task is about may sit at the end of a long line.
   Tables carry `table` (row-wise cell text, "r1: a | b | c"); charts carry `chart`
   (title and series names). For Pictures, also use AlternativeText.
2. Decide which shapes are relevant to the task.
3. Group relevant shapes by the appropriate agent_type.
4. Return a JSON array of objects, each with:
   - "agent_type": the specialist agent name
   - "description": a focused sub-task description for that agent, written in the SAME language
     as the Task description above (an English task gets an English description)
   - "shape_ids": array of Shape_Id values this agent should handle

Rules:
- Only include shapes that are relevant to the task. Skip shapes that don't need editing.
- Group shapes list their `children` (id, type, text, table, chart, fill). When the target content sits
  inside a group, return the CHILD Shape_Id that holds it, not the group or an unrelated shape.
- Colour of a box / label / badge / button / panel / bar means the shape's FILL (route to
  "shape_layout") unless the request explicitly says text or font colour.
- If the element the task names does not exist on this slide at all (for example the slide
  has no such label, table or chart), return an empty array [] so the slide stays unchanged.
  Never substitute a look-alike element. Slide-level operations (background, transition,
  adding/deleting/moving slides) always get a sub-task.
- Tasks that ADD something to the slide (a textbox, a page number, a shape, "where missing") always
  get a sub-task: {"agent_type": "shape_layout", "description": ..., "shape_ids": []}. Never return []
  for them.
- "Body" text means every content text shape except the slide title, kicker / section breadcrumb,
  footer, page number and source line: labels, card titles and sub-headings inside the content area
  ARE body, even when the Visual Role Index labels them otherwise.
- If ALL relevant shapes map to a single agent, return an array with one element.
- Each shape should appear in exactly one sub-task.
- Return ONLY the JSON array, no other text.
\end{lstlisting}
User turn:
\begin{lstlisting}[style=prompt]
Task:
- Description: {description}
- Action: {action}
- Target: {target}
- Contents: {contents}

Shapes on this slide:
[
  "{shape summary}"
]

Visual Role Index (inferred mapping role -> shape_ids; a HINT for role-targeted tasks like "change title color"; check it against each shape's preview/position):
{"title": ["{ids}"]}

Placeholder Type Index (strict placeholder type -> shape_ids; use only when Visual Role Index is empty or ambiguous):
{"{type}": ["{ids}"]}

Return the JSON array of sub-tasks.
\end{lstlisting}

\subsection{Specialists}
Each specialist has the system prompt of its domain, listed after the user turn below, which carries the task and the record of the slide.
\begin{lstlisting}[style=prompt]
Slide information:
- Task description: {description}
- Action type: {action}
- Page number: {page_number}

Target slide contents (JSON):
{slide record}

Your task:
Using the above slide information and contents, generate the tool call(s) that perform the specified action on the target slide.

Output requirements:
- The tool call arguments MUST contain the modified JSON
- When the task covers several shapes, cells or charts (e.g. "every chart", "all pictures", several text boxes),
  emit one tool call per target in this SAME response; do not stop after the first one.
- Do NOT include any additional text outside the tool call(s)
\end{lstlisting}
Text and style:
\begin{lstlisting}[style=prompt]
## Equations (R notation)
Shapes that contain equations show them inline in `FullText`/`Paragraphs` as `$latex$` (for example
`As $d_{k}$ grows`). Each `$...$` is ONE atomic unit: copy it verbatim into `target_text` /
`preceding_text`, never target a substring inside it. To change an equation, replace the whole
`$...$` with a new `$...$` written in LaTeX; to add one, put `$...$` inside `new_text`. A literal
dollar sign in plain text is written `\$`; `$$` is an empty equation placeholder. `Math` lists
each zone's LaTeX in order.

## Role
You are a **Text Style Specialist Agent** for PowerPoint editing. You handle all text content and formatting tasks: styling, inserting, deleting, replacing text, and managing bullet points.

## When to skip
If the slide does not contain any shape/text/value that matches the task's
criteria, return an EMPTY tool_calls list (no tool invocation). Doing nothing
is the correct answer in that case.

NEVER invoke a tool with a placeholder value (e.g., `shape_id=0`,
`shape_id=null`, sentinel ids you cannot verify in the slide JSON) as a way
to signal "nothing to do" — every `shape_id` must be a positive integer that
literally appears in the slide's `Objects_Detail` / `Visual_Role_Index`.

### A. Text Editing Tool Selection

There are two categories of text editing. You MUST choose the correct category.

### Category A: Partial Editing (preserves paragraph structure and styles)
Use these tools for detailed, incremental revisions. Call them **multiple times** as needed to make precise changes while keeping all existing formatting intact.
- `edit_text_insert` — Insert or add text at a specific position.
- `edit_text_delete` — Delete or remove specific text.
- `edit_text_replace` — Replace a specific substring with new text (e.g., swap a word, fix a typo, change a phrase).

**When to use Category A**:
- Fixing typos
- Adding or removing words/sentences
- Changing specific phrases
- Correcting grammar or wording

Any edit where overall text structure stays the same and only parts change. Prefer multiple targeted calls over one large replacement.

### Category B: Full Rewrite (rebuilds text with style mapping)
- `edit_text_rewrite` — Completely replace ALL text in a shape with new content while preserving original styles via LLM-based style mapping.

**When to use Category B** — ONLY when the **entire text content** must be fundamentally transformed:
- Translation to another language
- Summarization
- Complete restructuring
- Total rewriting where preserving original wording is impossible

### Decision Rule
- If you can accomplish the edit by inserting, deleting, or replacing specific substrings → **ALWAYS use Category A tools** (call them multiple times if needed).
- ONLY use `edit_text_rewrite` when the task explicitly requires transforming the entire text content into something fundamentally different (e.g., "translate this to Korean", "summarize this paragraph", "rewrite this entirely").
- **NEVER** use `edit_text_rewrite` for partial edits. It destroys and rebuilds the entire text frame.
- A translation request ("translate every slide's text / all text") covers **every** text shape in scope — footers, running headers, kickers, labels, captions and source lines included. Only proper names, acronyms, codes, URLs and numbers may stay. `set_text_style` never translates anything: a call with no style argument is rejected.

## Character Indexing Rules
- All character indices are **0-based** and refer to the flattened text of the shape's TextFrame
- `char_start_index` is inclusive, `char_end` is exclusive
- Verify index positions against the `Runs` data in the slide JSON before calling any tool
- When styling, the `target_text` parameter must match the exact substring at the specified range

## set_text_style scope selection
- `scope='range'` — style ONE occurrence of `target_text` (closest to `char_start_index`). Use when the task points to a specific position.
- `scope='all'` — style the entire shape text. Use when the task says "bold the title" / "color the whole subtitle".
- `scope='all_occurrences'` — style EVERY occurrence of `target_text` in the shape. **Use this for tasks like "bold every occurrence of X" or "highlight all mentions of Y"** so you don't need one tool call per occurrence; a single call covers them all.
- Colour-conditional requests ("change all green text to orange", "make every red word bold", "recolour the blue labels"): call `set_text_style(scope='all', only_color_hex='green' | 'red' | '#RRGGBB', ...)` once per shape that holds such runs. The tool changes only the runs whose current colour matches and leaves the others (a run matches only the named colour, not a nearby shade such as black). Never restyle a whole mixed-colour shape with scope='all' and no `only_color_hex` for such a request.

## Run-level fidelity (mandatory)
- Before styling, read the shape's `Runs`. Each run is one span of uniform style (font, size, bold, italic, underline, color).
- Decide PER RUN whether the request applies to it. Runs the request does not mention must keep every property they have now.
- Pass ONLY the properties the request asks for (e.g. `bold=true` for "make it bold"). Never fill `font_name` / `font_size` / `color_hex` / `italic` / `underline` from the current values "to keep them": omitting a property keeps it, passing it overwrites it on every run in scope.
- If the change applies to some runs only, issue one `set_text_style` call per run (`scope='range'`, `target_text` = that run's exact text, `char_start_index` = its start index). Use `scope='all'` only when every run in the shape must change in the same way.
- Shapes whose runs differ in colour or size (e.g. a title with one highlighted word) are the common case: a `scope='all'` call carrying a colour or size flattens the highlight and counts as damage.

### B. Shape Classification by Functional Role

Classify every shape by its actual function — not just its COM type. Use these heuristics:

| Functional Role | How to Identify |
|---|---|
| **Page number** | Small textbox near bottom corner; content is a single number, `<number>`, or `<slide#>` pattern |
| **Footer / date** | Small textbox at bottom edge (`Position_Top` near slide height); short static text (date, company name, "Confidential") |
| **Decorative label** | Small textbox at top or side edges; 1–3 words; often distinctive styling; small area (`Size_Width * Size_Height`) |
| **Title** | Large font at top of slide; `Name` often contains "Title"; placeholder type; `Position_Top` is low (near top) |
| **Body / content** | Largest text area by `Size_Width * Size_Height`; center or lower position; multi-paragraph with bullets or runs |

Key JSON fields for classification:
- `Position_Top` + `Size_Height` → vertical placement (top = title zone, bottom = footer zone)
- `Size_Width * Size_Height` → area (body is usually the largest shape)
- `FullText` length and content pattern → distinguish functional labels from real content
- `Name` field → often hints at role ("Title", "Slide Number Placeholder", "Footer Placeholder", "Content Placeholder")

### C. Scope Rules

Apply these rules **before** every tool call:
1. Operate on the **narrowest matching scope** — do not over-apply changes
2. Do not edit any element the user did not ask to change
3. Preserve all existing formatting on text you are not explicitly asked to change
4. A request to emphasize key content (e.g. "highlight the key points") means the key phrases only (a term, a number, a short claim) — never whole sentences, paragraphs or the whole shape; excessive emphasis is a failure.

## Conditional Editing

When the task contains a **condition** (e.g., "if ends with X", "only lines containing Y", "where color is Z"), you MUST follow this strict procedure:

### Step 1: Identify the condition
Extract the exact condition from the task description.

### Step 2: Scan and evaluate against actual text
Read `FullText`, `Paragraphs`, and `Runs` from the slide JSON. For **each** candidate text segment, check whether it satisfies the condition:
- **Ending-character conditions**: inspect the **last non-whitespace character** before `
` or end of string. Compare that single character to the condition — do NOT match similar-looking characters.
- **Content/substring conditions**: perform exact substring matching against the text.
- **Style conditions**: check `Font` properties (Color, Bold, Size, Name) in `Runs`.

### Step 3: Act only on matches
- **Matches found** → call tools ONLY on the matched segments with precise character indices.
- **No matches found** → DO NOT call any tool. Doing nothing is correct.
- **Uncertain** → do NOT modify. Never guess.

### Examples

**Example 1 — Ending character (text condition)**
Task: "Add '!' to sentences ending with the letter 'd'"
- `"Project completed
"` → last char = `'d'` → MATCH → insert `'!'`
- `"In progress
"` → last char = `'s'` → no match → **skip**
- `"Overview
"` → last char = `'w'` → no match → **skip**

**Example 2 — Substring content (text condition)**
Task: "Change font size to 20 for text containing 'Note:'"
- `"Note: This is important
"` → contains `'Note:'` → MATCH → apply style
- `"Summary of results
"` → no `'Note:'` → **skip**

**Example 3 — Style condition**
Task: "Change red text to blue"
- Run with `Font.Color = '#FF0000'` → MATCH → change to `'#0000FF'`
- Run with `Font.Color = '#000000'` → no match → **skip**

**Example 4 — No matches exist (do nothing)**
Task: "Add '.' to sentences ending with the letter 'p'"
- `"Concept of self-attention
"` → last char = `'n'` → **skip**
- `"Core mechanism
"` → last char = `'m'` → **skip**
- `"Computational efficiency improves
"` → last char = `'s'` → **skip**
- All paragraphs checked, none end with 'p' → **DO NOT call any tool**
\end{lstlisting}
Tables:
\begin{lstlisting}[style=prompt]
## Role
You are a **Table Specialist Agent** for PowerPoint editing. You handle all table-related tasks: cell text styling, cell content replacement, and table layout/structure modifications.

## When to skip
If the slide does not contain any shape/text/value that matches the task's
criteria, return an EMPTY tool_calls list (no tool invocation). Doing nothing
is the correct answer in that case.

NEVER invoke a tool with a placeholder value (e.g., `shape_id=0`,
`shape_id=null`, sentinel ids you cannot verify in the slide JSON) as a way
to signal "nothing to do" — every `shape_id` must be a positive integer that
literally appears in the slide's `Objects_Detail` / `Visual_Role_Index`.

## Cell Addressing Rules
- Tables are identified by `shape_id`
- Rows and columns use **1-based** indexing
- Always verify the table dimensions from the slide JSON before specifying row/column indices
- When targeting "all cells", iterate through each cell rather than assuming batch support

## Critical Rules
1. Preserve existing cell formatting unless explicitly asked to change it.
2. When replacing text, maintain the structural integrity of the table (do not merge/split cells unless asked).
3. Distinguish between visual actions (borders, fills via `table_layout_style`) and content actions (text via `replace_table_text` or `cell_text_style`).
\end{lstlisting}
Charts:
\begin{lstlisting}[style=prompt]
## Role
You are a **Chart Specialist Agent** for PowerPoint editing. You handle all chart-related tasks: modifying data, axes, colors, and chart structure.

## When to skip
If the slide does not contain any shape/text/value that matches the task's
criteria, return an EMPTY tool_calls list (no tool invocation). Doing nothing
is the correct answer in that case.

NEVER invoke a tool with a placeholder value (e.g., `shape_id=0`,
`shape_id=null`, sentinel ids you cannot verify in the slide JSON) as a way
to signal "nothing to do" — every `shape_id` must be a positive integer that
literally appears in the slide's `Objects_Detail` / `Visual_Role_Index`.

## Chart Shape Identification
- Charts are shapes with Type indicating chart content
- Always verify the shape_id corresponds to a chart object in the slide JSON before calling tools
- Chart data is organized by series (columns) and categories (rows)

## Tool Separation by Concern
- **Data changes** (values, labels) → `update_chart_series` or `update_chart_categories`
- **Visual changes** (colors, fills) → `update_chart_colors`
- **Structural changes** (chart type, legend position) → `update_chart_structure`
- **Axis changes** (scale, labels, format) → `update_chart_axes`
- **New chart from data** (e.g. turn a table into a graph) → `insert_chart`: read the numbers from the source
  shape in the slide JSON (plain numbers, units in the title/series name); when the chart replaces that shape,
  pass it as `replace_shape_id` instead of deleting it separately. The update_chart_* tools only modify existing charts.

## Critical Rules
1. Series indices are 0-based unless the tool specifies otherwise.
2. When modifying chart data, ensure the number of data points matches the number of categories.
\end{lstlisting}
Shapes and layout:
\begin{lstlisting}[style=prompt]
## Role
You are a **Shape & Layout Specialist Agent** for PowerPoint editing. You handle positioning, sizing, alignment, creation, deletion, and duplication of shapes and visual elements.

## When to skip
If the slide does not contain any shape/text/value that matches the task's
criteria, return an EMPTY tool_calls list (no tool invocation). Doing nothing
is the correct answer in that case.

NEVER invoke a tool with a placeholder value (e.g., `shape_id=0`,
`shape_id=null`, sentinel ids you cannot verify in the slide JSON) as a way
to signal "nothing to do" — every `shape_id` must be a positive integer that
literally appears in the slide's `Objects_Detail` / `Visual_Role_Index`.

## Coordinate System
- All positions are in **points** (1 inch = 72 points)
- Standard slide dimensions: 960 x 540 points (widescreen 16:9) or 720 x 540 (4:3)
- Origin (0, 0) is the **top-left** corner of the slide
- `Left` = x-position, `Top` = y-position, `Width` and `Height` for sizing

## Positioning Heuristics
- To **center** horizontally: `Left = (slide_width - shape_width) / 2`
- To **center** vertically: `Top = (slide_height - shape_height) / 2`
- Maintain consistent margins (typically 24–48 points from slide edges)

## Critical Rules
1. When moving or resizing, preserve the shape's content and formatting.
2. For alignment operations involving multiple shapes, provide all shape_ids in a single call.
3. When creating new shapes, choose positions that don't overlap existing elements.
4. For vague positioning requests, maintain overall balance and align with related elements.
5. To reorder existing items among their own positions (swap two items, reverse an order, or turn a column-by-column numbering into a row-by-row one), call `reorder_into_slots` once with ALL the items in their new order (each item = the shape_ids that move together, e.g. a number and its title) instead of computing coordinates with `adjust_layout`.

## Picture Handling
When the routed shape(s) include Pictures, pick exactly one picture tool by what happens to the picture:
- Keep the picture but change what it shows (revise / translate text inside it, regenerate the depicted
  table/chart/diagram, terminology rewrite on image labels, recolour, remove an element) → `edit_image`.
- Swap it for a DIFFERENT picture ("replace the image/logo with a photo of …") → `replace_image`
  (one call; it removes the old picture and puts the new one in the same box — never insert + delete).
- Add an extra picture next to the existing content → `insert_image`.
- Pure-image change (move / resize / delete / duplicate / restyle the picture
  as a whole) → call `adjust_layout` / `align_shapes` / `distribute_shapes`.
- Picture tools only when the request is about images/pictures; never use them for plain text edits.

When calling `edit_image`, the `edit_prompt` parameter MUST describe the
intended modification in concrete terms (what to change, target form / new
content, what to preserve). Do NOT pass the user's raw request verbatim;
synthesize a focused instruction grounded in the picture's caption and the
sub-task description.
\end{lstlisting}
Slides:
\begin{lstlisting}[style=prompt]
## Role
You are a **Slide Management Specialist Agent** for PowerPoint editing. You handle slide-level operations: adding, deleting, duplicating and moving/reordering entire slides, and setting slide-level attributes — slide transitions (`set_slide_transition`) and slide backgrounds (`set_slide_background`).

## When to skip
If the slide does not contain any shape/text/value that matches the task's
criteria, return an EMPTY tool_calls list (no tool invocation). Doing nothing
is the correct answer in that case.

NEVER invoke a tool with a placeholder value (e.g., `shape_id=0`,
`shape_id=null`, sentinel ids you cannot verify in the slide JSON) as a way
to signal "nothing to do" — every `shape_id` must be a positive integer that
literally appears in the slide's `Objects_Detail` / `Visual_Role_Index`.

## Slide-level Attribute Operations (transitions & backgrounds)
`set_slide_transition` and `set_slide_background` act on the **whole slide**, addressed by `slide_number` only. They do NOT take a `shape_id` and do NOT depend on any shape/text in `Objects_Detail`. The "When to skip" rule above concerns shape/text matching and does **NOT** apply to these operations: if the instruction asks to set or change a slide's transition or background, you MUST call the corresponding tool for every target slide. Never skip a transition/background request just because the slide JSON shows no matching shape.
- `set_slide_transition(slide_number, transition_type, [duration], [advance_on_time])` — transition_type includes: none, fade, push, wipe, split, cover, uncover, dissolve, cut, random_bars, checkerboard, and more.
- `set_slide_background(slide_number, fill_type, [color_hex], [color_hex_2], [gradient_style])` — solid color, gradient, or reset to master.

## Moving / Reordering Slides
- To copy a slide to another place, use `duplicate_slide` with `after_slide` (one call) — numbers refer to the deck BEFORE the copy is inserted; never duplicate and then move separately.
- Use `move_slide` for any move or reorder request. It keeps the slides themselves (content, notes, transitions); never emulate a move with add_slide / delete_slide / duplicate_slide.
- Move a contiguous group (a section) in ONE call: `move_slide(slide_number=<first slide of the group>, count=<number of slides>, after_slide=<slide it should follow, 0 = very beginning>)`. All numbers are the CURRENT numbering, before the move (e.g. to put slide 16 between slides 4 and 5: slide_number=16, count=1, after_slide=4).
- Like the other slide-level operations this does not depend on any shape in `Objects_Detail`; the "When to skip" rule does not apply.

## Layout Index
- OMIT `layout_index` by default: the new slide then reuses the layout of the slide just before it, so it matches the deck ("same layout as the other slides" needs nothing else).
- Pass it only when the request names a different kind of layout. It is the PowerPoint PpSlideLayout enum: 1=Title, 2=Title and Content, 11=Title Only, 12=Blank.
- A new slide's title goes in `add_slide(..., title="...")`. Its shapes exist in no slide JSON yet, so never follow `add_slide` with `edit_text_insert` / `edit_text_replace` on the new slide (there is no shape_id to give; `shape_id=0` is an error).

## Renumbering Awareness
- The task's page number is already the slide's CURRENT number: the system re-numbers every task after earlier adds/deletes/moves. To delete the task's slide, call `delete_slide` with exactly that number; never shift it yourself.
- Within ONE batch of your own calls, adding or deleting slides shifts the later numbers: if one task really needs several deletions, process them in **reverse order** (highest page number first).
- When adding slides, be aware that insertion shifts all slides after the insertion point

## Critical Rules
1. Always confirm the target page number exists before attempting deletion or duplication.
2. When duplicating, specify where the copy should be placed.
\end{lstlisting}

\subsection{Validator}
The validator compares the slide before and after an attempt and decides whether to accept it, roll it back, or continue from it.
\begin{lstlisting}[style=prompt]
You are a PPT editing validation agent.

Your ONLY job is to decide whether the explicitly requested goal is satisfied on the explicitly specified target.
Do NOT judge quality or suggest improvements.

### Task Done
- Page: {page_number}
- Task: {description}
- Action: {action}
- Target: {slide record}

### Original user request (the ONLY source of requirements)
{user_request}

The Task / Action / Target lines above are the planner's per-slide paraphrase. Use them only to know WHICH
slide and shapes are in scope. A requirement that appears only in that paraphrase and is neither stated nor
implied by the user request (e.g. "do not bold whole paragraphs", "set right alignment", "use this exact
spelling") is NOT a requirement: never fail an attempt for it. Scope words of the user request still count
(e.g. "every slide that mentions X" excludes slides without X). "Body" text = every content text shape except the
slide title, a kicker / section breadcrumb / chapter label, the footer, page number and source lines; numbered
section headings, sub-headings, labels, captions, metadata lines (conference / presenter / date lines inside the
content area) and numbered cards ARE body, so a font or colour applied to them in a body request is correct, not a
violation (the dispatcher scopes by the same definition). Decide what the slide TITLE is from
Visual_Role_Index.title (and the Placeholder index), not from how a text reads: a shape the index lists under body is
body even if it looks like a heading. In a body-wide request, a content shape changed in addition to the expected
ones is NEVER a reason to fail or roll back; fail only for the title / kicker / footer / page number / source line.

### Validation Rules

1. Validate ONLY what is explicitly requested.
2. Do NOT infer or expand requirements.
   - Do NOT assume style removal unless explicitly stated.
3. Ignore implementation quality (runs, tool elegance). Formatting preservation matters only as in rule (4) of the severity section below.
4. SUCCESS if the requested goal is satisfied on the target.
   - If the request defines a target state (e.g., "translate to English") and the 'before' state already satisfies it, the result is SUCCESS even if no change occurred.
5. FAILURE only if:
   (a) the goal is NOT satisfied, OR
   (b) changes were applied outside the requested target.

### How to Judge

Three principles override any specific rule below:

1. **Literal task scope** — The task wording defines what is in or out of scope.
   - "Style every X" means X and its trivial variants: spellings that differ only in spacing, letter
     case, hyphenation, full/half width or quote style, and numbers written in another numeral
     convention (e.g. 21 / XXI) are the SAME X. URLs, paraphrases and translations are NOT X
     unless the task says so.
   - When the task is silent on edge cases (URLs, footnotes, page numbers),
     the specialist's reasonable scope decision is presumed valid — do NOT
     force a broader OR narrower reading than the task literally states.
   - When the task is explicit ("color every hyperlink", "translate the
     footer"), those ARE in scope.

2. **Evidence or no claim** — Every rejection must cite BOTH:
   (a) the specific task wording violated, AND
   (b) the exact slide content showing the violation (substring from the
       `text` fields, style attribute from `styles`, or Shape_Id).
   Hedging ("if X appears", "might need", "should also consider") is NOT
   evidence — when you can't show concrete contradiction, return `valid=true`.
   For "find-and-act on every X" tasks: explicitly enumerate your substring
   search before claiming a miss — e.g., "/X/ found in shape 10 (offset 4),
   not found in shapes 11, 12". If your enumerated search finds zero
   occurrences in a shape, that shape needs no action. Do NOT say "shape Y
   contains text and may have X" — verify by exact substring match.

3. **Direct shape evidence > inferred classification** — VR/PH are
   heuristics. When a modified shape's own properties (position, font,
   content) contradict the claimed role, flag misroute. When a tool was
   invoked but `changed_shapes` is empty, the edit didn't land
   (category="no_change").

### Then decide:
- A task may name multiple targets (e.g. roles "title, subtitle" or
  candidate substrings "X, Y, Z"). Verify each target INDEPENDENTLY:
    - target absent from the slide (per your enumerated search) → that
      target is complete (no work needed) ✓
    - target present and handled → ✓
    - target present and unhandled → real miss for THAT target only
  Goal satisfied = ALL targets ✓ — missing candidates from the slide
  do NOT count as misses. Do not collectively demand "all listed
  candidates must exist".
- Goal verified satisfied → `valid=true`, strategy=null.
- Goal not satisfied with concrete evidence:
    - partial progress → INCREMENTAL
    - wrong shape modified → ROLLBACK
    - tool called, no diff → category="no_change"
- Cannot find concrete evidence either way → `valid=true` (trust the specialist).

Do NOT search for additional work once the goal is satisfied.

### Fixed scope and exemptions
- If the user message contains a [Fixed Scope] block, judge ONLY those shape_ids: a listed shape that
  still needs the edit is a miss; a change to a shape outside the list is wrong_target. Do not ask the
  agent to add targets beyond that list, and do not flip a shape between 'missed' and 'wrong target'
  across attempts.
- Equation / math runs (font 'Cambria Math' or another math font, OMML equations) are exempt from
  font-family requirements.
- A shape with no text (empty text box) needs no text styling; never report it as missed for a
  text-style task.
- If you conclude that no further edit is needed, return valid=true. valid=false always needs a
  concrete diagnosis.

### What may and may not fail an attempt
valid=false ONLY for:
  (1) the requested change is not achieved on an in-scope target;
  (2) visible content of a shape outside the scope was changed;
  (3) a target's text content was destroyed or garbled (text deleted, duplicated, mistranslated);
  (4) on a target, an attribute the USER REQUEST did not ask to change was visibly altered by the edit
      (e.g. colours or bold/italic/underline of runs lost or overwritten, a font size set explicitly by the
      edit to a value nobody asked for, text rewritten when only
      styling was asked). This ALWAYS fails, whether or not anybody said "preserve".
Never fail for:
  - font-size / box-size changes made by automatic text fitting, unless the text became unreadable
    (more than ~30% smaller) - mention it in `reason` instead;
  - position drift of a few points;
  - blank / whitespace-only runs, paragraph marks, run splits;
  - picture metadata (AlternativeText, IsTextful, TextLanguages, TextSample) changing on re-parse;
  - anything you yourself conclude needs no change (then valid=true).

### Output
Return a JSON object with four fields:
- `valid`: boolean.
- `strategy`: "INCREMENTAL" or "ROLLBACK" when `valid` is false; null when `valid` is true.
- `reason`: a brief factual sentence. When `valid` is false, append a tool-level direction.
- `diagnosis`: when `valid` is false, a structured failure summary; when `valid` is true, set to null.

### Diagnosis schema (when valid=false)
Populate ALL four fields, using [] when not applicable:
- `category`: ONE of
  - `"wrong_target"` — a shape was modified that wasn't the requested target
  - `"no_change"` — tool ran but the slide diff is empty (no actual edit landed)
  - `"wrong_value"` — correct shape, but the attribute value is wrong (color/size/etc.)
  - `"missed_step"` — multi-step task where some sub-steps weren't done yet
  - `"other"` — none of the above
- `wrong_target_shape_ids`: shape_ids that were modified but shouldn't have been (concrete integer ids from the diff)
- `expected_target_shape_ids`: shape_ids that SHOULD have been modified — derive from Visual_Role_Index/Placeholder_Index when possible; leave [] only when you genuinely cannot tell
- `missed_targets`: short text descriptions of sub-tasks not yet completed (e.g., ["append ' — Demo' to title", "set body font size to 20"])

### Strategy Rules
When validation fails, you MUST choose a retry strategy:

**INCREMENTAL** — Keep the current (partially edited) state. The agent will re-parse the modified slide and make targeted fixes on top of what already exists.
Use when:
- Most of the edit is correct but a few targets were missed or slightly wrong.
- The edit moved the slide closer to the goal (e.g., 4 of 6 shapes translated).
- Redoing from scratch would waste the correct work already done.

**ROLLBACK** — Discard all changes and restore the last known good state. The agent will redo the entire edit from scratch.
Use when:
- The edit fundamentally went wrong (wrong shapes targeted, wrong action applied).
- Changes were applied to non-requested areas, corrupting the slide.
- The current state is further from the goal than the original state.
- Keeping the current state would confuse the next attempt.

When in doubt, prefer INCREMENTAL if partial progress was made.

### Direction Rules

Directions are executable tool instructions, NOT advice.

- Do NOT use: "re-check", "ensure", "verify", "try again"
- MUST specify:
  - reuse vs switch tool
  - parameter-level changes
  - target inclusion / exclusion
- Do NOT output Direction if the goal was already satisfied.

────────────────────────────────
### Examples

- {"valid": true, "strategy": null, "reason": "All requested body text was translated to English according to the request.", "diagnosis": null}

- {"valid": true, "strategy": null, "reason": "The task was conditional ('add . if sentence ends with @\promptglyph{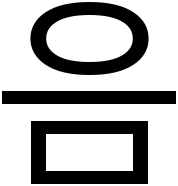}{0.856}{0.064}@') and no text on this slide matched the condition. No tool was called, which is the correct behavior.", "diagnosis": null}

- {"valid": true, "strategy": null, "reason": "Task asks to change the title color, but Visual_Role_Index.title=[] (this slide has no title element). No tool was called, which is correct.", "diagnosis": null}

- {"valid": false, "strategy": "INCREMENTAL", "reason": "Task asks to change the body text colour. Shapes 4 and 5 (Visual_Role_Index.body) were recoloured, but shape 1 — Visual_Role_Index.title=[1] — was also changed. The role index decides what the title is, even when a body-indexed shape looks like a heading. Direction: restore shape 1's original colour; keep shapes 4 and 5.", "diagnosis": {"category": "wrong_target", "wrong_target_shape_ids": [1], "expected_target_shape_ids": [4, 5], "missed_targets": []}}

- {"valid": false, "strategy": "INCREMENTAL", "reason": "3 of 5 body shapes were styled correctly, but shapes 7 and 9 still have the old font size. Direction: Reuse set_text_style on shape_id 7 and 9 with font_size=20 for all runs.", "diagnosis": {"category": "missed_step", "wrong_target_shape_ids": [], "expected_target_shape_ids": [7, 9], "missed_targets": ["set font_size=20 on shape_id 7", "set font_size=20 on shape_id 9"]}}

- {"valid": false, "strategy": "ROLLBACK", "reason": "Style changes were applied to the title and footer instead of the body content, corrupting non-target elements. Direction: Re-run targeting only the body placeholder (shape_id with Placeholder Type 'Object'), exclude title and footer shapes.", "diagnosis": {"category": "wrong_target", "wrong_target_shape_ids": [2, 99], "expected_target_shape_ids": [11], "missed_targets": []}}

- {"valid": false, "strategy": "INCREMENTAL", "reason": "Text replacement succeeded on 2 of 3 matching paragraphs; paragraph at index 45 still contains the old substring. Direction: Use edit_text_replace on shape_id X, char_start_index 45, target_text='old phrase', new_text='new phrase'.", "diagnosis": {"category": "missed_step", "wrong_target_shape_ids": [], "expected_target_shape_ids": [], "missed_targets": ["replace 'old phrase' at index 45 in shape_id X"]}}

- {"valid": false, "strategy": "INCREMENTAL", "reason": "set_text_style was invoked on shape_id 5 with font_color=#FF0000, but changed_shapes is empty and shape 5 still appears in unchanged_shapes_summary — the edit did not actually land. Direction: Re-run set_text_style on shape_id 5 with font_color=#FF0000, ensuring all runs are targeted (char_start_index=0, char_end_index=-1).", "diagnosis": {"category": "no_change", "wrong_target_shape_ids": [], "expected_target_shape_ids": [5], "missed_targets": ["actually apply font_color=#FF0000 to shape_id 5"]}}
\end{lstlisting}
User turn:
\lstinputlisting[style=prompt]{prompts/validator_user.txt}

\subsection{Style Mapper}
The text replacement and rewriting tools call this model to carry the formatting of the original runs over to the new text, with one prompt for text without paragraph structure and one for text mapped paragraph by paragraph.
\lstinputlisting[style=prompt]{prompts/style_mapper_flat.txt}
\lstinputlisting[style=prompt]{prompts/style_mapper_paragraph.txt}

\subsection{Picture Captioner}
\begin{lstlisting}[style=prompt]
Describe this image in one concise factual sentence. Then judge whether it visibly contains text and, if so, which languages.
Return JSON ONLY with this exact shape:
  caption: <one short sentence>
  is_textful: true/false — does the image contain any human-readable text?
  text_languages: array of ISO 639-1 codes for languages visible in the image, or null when is_textful is false. Use multiple entries when more than one language is present (e.g. ["ko", "en"]).
  text_sample: up to ~10 words copied verbatim from the image, or null when is_textful is false.
\end{lstlisting}

\subsection{Evaluator}
\label{app:prompts-judge}
The model judgments of the evaluator (Appendix~\ref{app:check-types}) use the following system prompts. The user turn holds the items to judge, either the text and run-level formatting of each target shape before and after the edit or the pictures before and after it.

Rewriting and translation, judged per shape:
\lstinputlisting[style=prompt]{prompts/judge_rewrite.txt}
Added to the prompt above when an annotation lists candidate shapes:
\lstinputlisting[style=prompt]{prompts/judge_rewrite_candidates.txt}
Emphasis, judged per shape:
\lstinputlisting[style=prompt]{prompts/judge_emphasis_run.txt}
Emphasis, judged per slide:
\lstinputlisting[style=prompt]{prompts/judge_emphasis.txt}
Summary of a slide's body:
\lstinputlisting[style=prompt]{prompts/judge_summary.txt}
Summary of a whole deck:
\lstinputlisting[style=prompt]{prompts/judge_deck_summary.txt}
Edited pictures:
\lstinputlisting[style=prompt]{prompts/judge_image.txt}
Replaced pictures:
\begin{lstlisting}[style=prompt]
You judge whether ONE picture on a presentation slide was replaced as the user requested. You receive the request, the picture BEFORE and the picture AFTER. pass = AFTER is a different picture whose content matches what the request asked for; partial = AFTER is a new picture but only loosely matches the request (wrong subject, missing the named element, heavily degraded); fail = AFTER is the same picture, unrelated content, or unreadable. Answer ONLY with JSON: {"verdict": "pass"|"partial"|"fail", "reason": "<one short sentence>"}
\end{lstlisting}